\documentclass[lettersize,journal]{IEEEtran}
\usepackage{amsmath,amsfonts}
\usepackage{algorithmic}
\usepackage{algorithm}
\usepackage{array}
\usepackage[font=normalsize,labelfont=sf,textfont=sf]{subcaption}
\usepackage{textcomp}
\usepackage{stfloats}
\usepackage{url}
\usepackage{verbatim}
\usepackage{graphicx}
\usepackage{tabularx}
\usepackage{hyperref}
\usepackage{multirow}
\usepackage{cite}
\usepackage{booktabs}
\usepackage{tabularx}
\usepackage{amssymb}
\usepackage{amsmath}
\usepackage{amssymb}
\usepackage{mathtools}
\usepackage{amsthm}

\usepackage{array}

\theoremstyle{plain}

\theoremstyle{definition}

\theoremstyle{remark}

\begin{document}

\title{A Drift-Stable Quantum Federated Learning for Intelligent Services}

\author{Shanika Iroshi Nanayakkara and Shiva Raj Pokhrel,~\IEEEmembership{~Senior IEEE Member}
\thanks{Shanika Iroshi Nanayakkara is with School of IT,  Deakin University, VIC 3125, Burwood, Australia, (e-mail: s222112938@deakin.edu.au).
}
\thanks{Shiva Raj Pokhrel is with School of IT,  Deakin University, VIC 3125, Burwood, Australia, (e-mail: shiva.pokhrel@deakin.edu.au).
}
}

\markboth{Journal of \LaTeX\ Class Files,~Vol.~14, No.~8, June~2025}%
{Shell \MakeLowercase{\textit{et al.}}: A Sample Article Using IEEEtran.cls for IEEE Journals}


\maketitle

\begin{abstract}
Quantum federated learning enables distributed clients to train quantum neural networks without sharing local data, making it promising for privacy-aware intelligent services. Intelligent services in this context refer to privacy-sensitive distributed decision systems, such as fraud detection and genomic classification, where reliable and fair client-level learning is as important as the accuracy of the aggregate model. However, heterogeneous client data and noisy quantum optimization often cause unstable local updates, client drift, and unfair performance between clients. This paper proposes DUQFL-Prox, a drift-stable quantum federated learning framework based on deep-unfolded local optimization. Instead of using a fixed local optimizer, each client performs adaptive unfolded SPSA updates, while a proximal term keeps the local model close to the global model. A lightweight controller learns step-specific optimization parameters to improve post-aggregation performance. Experiments on financial fraud and genomic classification tasks show that DUQFL-Prox improves stability, generalization, and client fairness compared with standard QFL baselines. The results suggest that deep-unfolded quantum federated learning can support more reliable and fair intelligent services in heterogeneous distributed environments.
\end{abstract}


\section{Introduction}


\IEEEPARstart{F}{ederated} learning (FL) has emerged as a promising framework for collaborative model training, where multiple clients optimize a shared model without centralizing raw local data. By allowing clients to train locally and communicate only model updates to a coordinating server, FL provides an attractive paradigm for privacy-aware and communication-efficient intelligence~\cite{li2020fedProx}. As quantum machine learning continues to develop, this distributed setting becomes increasingly relevant in quantum contexts as well, giving rise to quantum federated learning (QFL), in which multiple clients collaboratively train quantum or quantum-enhanced models while keeping local data decentralized~\cite{yu2022quantum, huang2022quantum, nguyen2025quantum}. In this setting, however, optimization becomes substantially more challenging than in conventional classical FL, because local training of quantum neural networks (QNNs) is often noisy, nonconvex, shot-sensitive, and highly dependent on optimizer configuration \cite{jerbi2024shadows,yu2022quantum}.

In this work, the phrase ``intelligent services'' refers to distributed AI-enabled service environments in which data-driven decisions must be learned from decentralized, privacy-sensitive, and heterogeneous client data. Such services include financial fraud detection, genomic classification, healthcare analytics, cyber-physical monitoring~\cite{azadeh2}, and edge intelligence, where data are naturally distributed across institutions, devices, or service providers and cannot be freely centralized due to privacy, regulatory, or ownership constraints~\cite{azadeh1}. These settings require not only high aggregate accuracy, but also stable optimization, fair client-level performance, and reliable generalization across heterogeneous participants~\cite{azadeh3}. 

\begin{figure}[t]
     \centering
\includegraphics[width=0.68\linewidth]{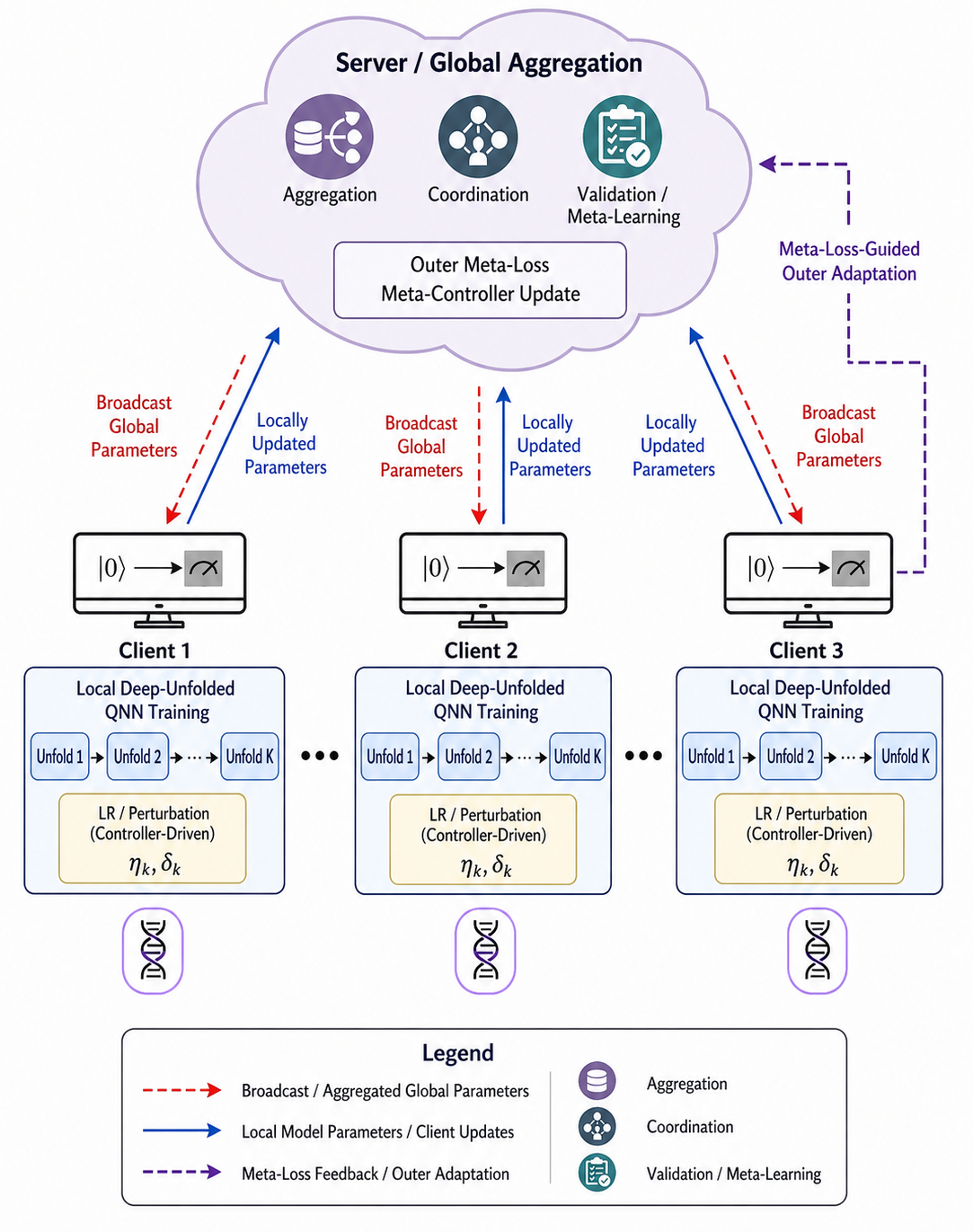}
  \caption{High-level DUQFL setup with adaptive local QNN training, selected-model aggregation, and controller-guided global optimization.}
\label{fig:fl_Archi} 
\end{figure}
In heterogeneous federated environments, one common strategy for improving
training efficiency is to exclude, down-weight, delay, or selectively sample
slow clients, often referred to as stragglers. Previous FL studies have shown that
system heterogeneity, including differences in computation and communication
capabilities, can substantially slow synchronous federated training
\cite{li2020fedProx,kairouz2021advances,reisizadeh2022straggler}. 
Client-selection methods such as Oort therefore improve time-to-accuracy by
prioritizing clients with favorable statistical and system utility
\cite{lai2021oort}. However, selection or straggler-mitigation strategies can
also create a representativeness issue when slow, resource-limited, or
statistically distinctive clients are repeatedly underrepresented.
Recent work
on biased client selection and participation imbalance shows that non-uniform
client participation can bias the learned global model and affect fairness or
client-level performance \cite{cho2022towards,selialia2024mitigating}. 
In practical QFL settings, this issue is undesirable because heterogeneous
clients may contain rare, domain-specific, or clinically important data
distributions. Rather than excluding such clients, DUQFL-Prox aims to stabilize
their local QNN optimization trajectories so that their updates remain useful
for global aggregation.

A central limitation in existing QFL pipelines is that local optimization is often treated as a fixed procedure. In a typical setting, each client receives the broadcast global model, performs a predetermined number of local optimization steps using a fixed optimizer schedule, and returns the resulting parameters to the server. This design implicitly assumes that the same local training rule is suitable for all clients, data distributions, and communication rounds. Such an assumption is restrictive in QFL, where QNN optimization is sensitive to learning rate, perturbation scale, measurement noise, and the non-convex geometry of variational quantum loss landscapes. In non-IID data, fixed local optimization can therefore produce uneven client updates, excessive client drift, and unstable post-aggregation behavior.

SPSA is attractive for QNN training because it estimates an update direction using only two objective-function evaluations, independent of the number of trainable parameters \cite{spall2002multivariatespsa}. This property makes SPSA practical for variational quantum models, particularly when analytic gradients are costly, unavailable, or noisy \cite{qiskit_spsa}. However, in federated QNN training, a fixed SPSA schedule can be too rigid. Conservative settings may slow local improvement, whereas aggressive settings may amplify client drift and reduce the compatibility of local updates during aggregation. Therefore, the central challenge is not merely to use SPSA, but to make local SPSA-based QNN optimization adaptive while keeping client updates aligned with the global federated objective.

To address this challenge, we propose \emph{Deep-Unfolded Quantum Federated Learning with Proximal Regularization (DUQFL-Prox)}, a drift-stable QFL framework for heterogeneous intelligent services. DUQFL-Prox decomposes local SPSA-based QNN training into \(K\) unfolded optimization blocks. At each unfold step, a lightweight shared controller generates step-specific SPSA learning rates and perturbation scales from optimization-state features, including unfold progress, recent loss behavior, parameter displacement, and client context. This enables local QNN optimization to adapt across clients, unfold steps, and communication rounds, rather than relying on a fixed optimizer schedule.

However, adaptive local optimization alone can amplify client drift under non-IID federated data. DUQFL-Prox therefore introduces proximal regularization into the client objective to penalize excessive deviation from the broadcast global model. This encourages each client to improve their local QNN objective while maintaining aggregation-compatible updates. In addition, each client uploads the validation-preferred unfolded checkpoint rather than necessarily returning the final unfolded state, preventing over-aggressive later updates from degrading global aggregation.

DUQFL-Prox follows a bilevel learning structure. At the inner level, each client performs deep-unfolded proximal SPSA optimization of its local QNN parameters. At the outer level, the shared controller is periodically refined using a meta-objective defined on post-aggregation validation behaviour. Thus, DUQFL-Prox learns not only the QNN parameters but also how local QNN optimization should evolve so that client updates become more stable, fair, and useful after aggregation.

The main contributions of this work are summarized as follows:
\begin{itemize}
    \item We propose DUQFL-Prox, a drift-stable deep-unfolded QFL framework for heterogeneous intelligent services, where local QNN training is modeled as an adaptive multi-step optimization process rather than a fixed optimizer routine.
    \item We design a controller-driven SPSA mechanism that generates unfold-specific learning rates and perturbation scales from local optimization-state features, enabling adaptive QNN training across clients, communication rounds, and unfolded steps.
    \item We incorporate proximal regularization and validation-based best-unfold selection to reduce client drift, prevent over-specialized local updates, and improve aggregation compatibility.
\end{itemize}

Importantly, we introduce an outer meta-loss-guided controller adaptation mechanism that aligns local optimization behavior with post-aggregation global performance.
We evaluate DUQFL-Prox across financial fraud detection and genomic classification tasks using global accuracy, client-level generalization, train--test gap, fairness gap, and imbalance-aware classification metrics.

The remainder of this paper is organized as follows. Section~\ref{sec:related_work} reviews related work on QFL, QNN optimization, federated heterogeneity, and deep unfolding. Section~\ref{sec:problem_methodology} presents the problem formulation and the DUQFL-Prox methodology. Section~\ref{sec:experimental_setup} describes the experimental setup and reproducibility protocol. Section~\ref{sec:experimental_results} presents the empirical results and ablation analysis. Section~\ref{sec:discussion} discusses implications, limitations, and future research directions, and Section~\ref{sec:conclusion} concludes the paper.

\section{Related Work}
\label{sec:related_work}

\begin{table*}[htbp]
\centering
\scriptsize
\caption{Research gap. 
$\checkmark$ indicates \textit{explicit support}, $\triangle$ indicates \textit{partial support}, and $\times$ indicates \textit{limited or no support}.}
\label{tab:related_work_positioning}
\resizebox{\textwidth}{!}{%
\begin{tabular}{p{3cm} p{1.5cm} c c c c c c p{6.6cm}}
\toprule
\textbf{Research stream} 
& \textbf{References} 
& \textbf{QFL} 
& \textbf{QNN} 
& \textbf{Adaptive} 
& \textbf{Client drift } 
& \textbf{L2L} 
& \textbf{Generalization} 
& \textbf{Observed Gap} \\
\midrule

Secure / privacy-preserving QFL 
& \cite{li2021quantumblind,chu2023cryptoqfl,yun2022quantum, hanna2023real} 
& $\checkmark$ 
& $\triangle$ 
& $\times$ 
& $\times$ 
& $\times$ 
& $\times$ 
& Focuses mainly on privacy/security rather
than optimiser stability or heterogeneous
client drift. \\

\midrule

QFL for financial fraud detection
& \cite{innan2025qfnn}
& $\checkmark$
& $\checkmark$
& $\triangle$
& $\triangle$
& $\times$
& $\triangle$
& Applies quantum federated neural networks to financial fraud detection, but does not explicitly formulate local QNN optimization as a deep-unfolded SPSA trajectory with proximal client-drift control or outer meta-loss-guided controller adaptation. \\
\midrule
Quantum-secured fraud detection / blockchain-assisted quantum applications
& \cite{11264016quanfraud}
& $\times$
& $\triangle$
& $\times$
& $\times$
& $\times$
& $\triangle$
& Studies quantum-state verification, blockchain-assisted verification, and quantum-enhanced fraud detection, but does not address federated QNN optimization, client drift, or learning-to-learn. \\

\midrule

Post-quantum data integrity and secure outsourcing
& \cite{dopiv}
& $\times$
& $\times$
& $\times$
& $\times$
& $\times$
& $\times$
& Addresses post-quantum secure data outsourcing and public integrity verification for cloud storage, but does not focus on QFL, QNN training, or heterogeneous federated optimization. \\

\midrule

QFL for autonomous / distributed systems 
& \cite{yamany2021oqfl,narottama2023federated} 
& $\checkmark$ 
& $\checkmark$ 
& $\triangle$ 
& $\triangle$ 
& $\times$ 
& $\triangle$ 
& Studies distributed or application-specific QFL, but local optimiser trajectories are not meta-controlled. \\

\midrule

QFL for healthcare and biomedical learning 
& \cite{huang2022quantum,pokhrel2024quantum,zhao2023nonQFLinf} 
& $\checkmark$ 
& $\checkmark$ 
& $\triangle$ 
& $\triangle$ 
& $\times$ 
& $\triangle$ 
& Demonstrates application feasibility, but
Often relies on global accuracy without
detailed client-level analysis.\\

\midrule
Quantum natural-gradient / optimization-based QFL 
& \cite{qi2023optimizing,chehimi2023foundations,xia2021quantumfed} 
& $\checkmark$ 
& $\checkmark$ 
& $\triangle$ 
& $\triangle$ 
& $\times$ 
& $\triangle$ 
& Improves local quantum optimization but
is usually not designed for federated non-
IID settings. \\

\midrule

Classical privacy-preserving FL with adaptive pruning
& \cite{AdpFL}
& $\times$
& $\times$
& $\checkmark$
& $\triangle$
& $\times$
& $\triangle$
& Addresses privacy-preserving FL, adaptive model pruning, communication cost, and non-IID data in classical federated learning, but does not consider QNN training, quantum optimization, deep unfolding, or quantum client drift. \\
\midrule
Classical deep unfolding / learning-to-optimise 
& \cite{nanayakkara2025new, nakai2024deep}
& $\times$ 
& $\times$ 
& $\checkmark$ 
& $\triangle$ 
& $\checkmark$ 
& $\triangle$ 
& Provides learnable optimisation principles, but is not designed for quantum neural networks or federated QNN training. \\

\midrule

\textbf{Proposed DUQFL-Prox} 
& \textbf{This work} 
& $\checkmark$ 
& $\checkmark$ 
& $\checkmark$ 
& $\checkmark$ 
& $\checkmark$ 
& $\checkmark$ 
& Deep-unfolded SPSA-based local QNN optimization with proximal client-drift regularisation, outer/meta validation guidance, and multi-domain evaluation using global accuracy, client accuracy, fairness, and generalization metrics. \\

\bottomrule
\end{tabular}%
}
\end{table*}

Table~\ref{tab:related_work_positioning} demonstrates that prior QFL studies have mainly focused on secure quantum communication, privacy-preserving learning, application-specific QFL implementations, and server-side aggregation mechanisms~\cite{li2021quantumblind,chu2023cryptoqfl,yun2022quantum,hanna2023real,
huang2022quantum,pokhrel2024quantum,zhao2023nonQFLinf,qi2023optimizing,
chehimi2023foundations,xia2021quantumfed}. While these studies establish the
feasibility of QFL. However, local QNN optimization is typically treated as a fixed inner
routine. This assumption becomes restrictive when client data are non-independent and identically distributed (non-IID) and quantum measurements are stochastic.
In such cases, local updates may become unstable, diverge from the global model, and reduce client-level generalization.

QNN training commonly relies on optimizers such as gradient descent, Adam,
COBYLA, and SPSA. SPSA is particularly suitable for variational quantum models
because it estimates a stochastic update direction using only two objective
function evaluations, independent of the number of trainable
parameters~\cite{qiskit_spsa_doc,spall1998overview}. However, in federated QNN
training, a fixed SPSA schedule may be inadequate: conservative settings can
slow local improvement, whereas aggressive settings can amplify client drift and
reduce aggregation compatibility. Therefore, the key challenge is not merely to
use SPSA, but to adapt SPSA-based local QNN optimization while preserving global
federated consistency.

Deep unfolding provides a principled way to convert iterative optimization into
a structured and learnable update process~\cite{shlezinger2025deep}. In
parallel, classical FL methods such as FedProx and SCAFFOLD show that client
drift is a central obstacle under heterogeneous data and that proximal or
correction-based mechanisms can improve stability~\cite{li2020fedProx,
karimireddy2020scaffold}. Although these methods are not quantum-specific, they
motivate the development of adaptive and drift-aware local optimization strategies for QFL.

Motivated by these gaps, we propose \emph{Deep-Unfolded Quantum Federated
Learning with Proximal Regularization (DUQFL-Prox)}. DUQFL-Prox unfolds local
SPSA-based QNN training into multiple controller-guided optimization blocks. The
controller generates step-specific learning rates and perturbation scales from
local optimization-state features, while a proximal term penalizes excessive
deviation from the broadcast global model. The controller is further refined
using a post-aggregation meta-objective, linking local optimization behaviour to
global validation performance.

The proposed framework differs from prior QFL work in three aspects. First,
local QNN optimization is modeled as a controller-driven unfolded process rather
than a fixed client routine. Second, proximal regularization is incorporated to
improve aggregation compatibility under client drift. Third, the controller is
updated using post-aggregation validation behaviour, rather than relying only on
local training signals. To the best of our knowledge, this combination of
unfolded SPSA-based QNN optimization, proximal client-drift control, and
outer meta-loss-guided controller adaptation has not been explicitly developed
in existing QFL literature.

Unlike client-filtering or straggler-removal strategies, DUQFL-Prox does not
discard heterogeneous clients. Instead, it stabilizes their local optimization
trajectories so that their updates remain useful for aggregation. This is
important in biomedical, genomic, fraud-detection, and remote-sensing settings,
where difficult clients may contain rare but important patterns. Accordingly,
our evaluation reports global accuracy together with mean client test accuracy,
train--test gap, client fairness gap, and imbalance-aware metrics such as
F1-score, precision, recall, ROC-AUC, PR-AUC, and MCC.

\section{Problem Formulation and Methodology}
\label{sec:problem_methodology}

\subsection{Problem Statement}
\label{subsec:problem_statement}

The objective of this work is to improve the stability, generalization, and
client-level reliability of QFL under heterogeneous client participation. In
conventional QFL, each client receives the broadcast global QNN parameters,
performs local optimization, and returns the updated parameters for server
aggregation. However, under non-IID data and stochastic quantum measurements,
fixed local optimization can produce unstable client trajectories and excessive
drift from the global model. Consequently, the aggregated model may achieve
reasonable global accuracy while still exhibiting poor client-level
generalization or imbalanced performance across clients.

Although adaptive optimizers such as Adam can adjust local parameter updates,
they are not designed to explicitly account for federated client heterogeneity,
quantum measurement noise, post-aggregation behaviour, or drift from the
broadcast global model~\cite{kingma2014adam}. Similarly, straggler removal or
client down-weighting can improve training efficiency, but may reduce
representativeness by underutilizing clients with limited resources or
statistically distinctive data~\cite{li2020fedProx,kairouz2021advances,
reisizadeh2022straggler}. This is undesirable in QFL applications such as
genomics, biomedical analysis, fraud detection, and remote sensing, where
difficult clients may contain rare but important data patterns.

DUQFL-Prox addresses this problem by regulating local QNN optimization rather
than excluding heterogeneous clients. The proposed method seeks a QFL training
procedure in which:
\begin{enumerate}
    \item local SPSA-based QNN updates adapt across unfolded optimization steps;
    \item local trajectories remain sufficiently close to the broadcast global
    model to remain aggregation-compatible; and
    \item the hyperparameter-generation policy is refined using
    post-aggregation validation performance.
\end{enumerate}

Thus, the central problem is to jointly control client-side unfolded QNN
optimization and server-side controller adaptation so that the global model
achieves stable post-aggregation performance, reduced client drift, and improved
client-level generalization under heterogeneous federated data.
\subsection{QFL Setting and Notation}
\label{subsec:qfl_setting_notation}

We consider a QFL system consisting of $N$ distributed clients and one
coordinating server. The clients collaboratively train a shared QNN model without
exchanging raw local data. Let the client set be
\begin{equation}
    \mathcal{C}=\{1,2,\ldots,N\}.
\end{equation}
Training proceeds over $T$ communication rounds indexed by
\begin{equation}
    t \in \{0,1,\ldots,T-1\}.
\end{equation}

At communication round $t$, the server maintains a global QNN parameter vector
\begin{equation}
    \boldsymbol{\theta}^{(t)} \in \mathbb{R}^{P},
\end{equation}
where $P$ denotes the number of trainable parameters in the variational quantum
model. These parameters define a parameterized quantum circuit, denoted by
\begin{equation}
    U(\mathbf{x};\boldsymbol{\theta}^{(t)}),
\end{equation}
where $\mathbf{x}$ is the classical input encoded into the quantum circuit and
$\boldsymbol{\theta}^{(t)}$ parameterizes the trainable ansatz layers. The QNN
output is obtained by measuring the resulting quantum state and applying
classical post-processing to obtain prediction probabilities.

The server broadcasts $\boldsymbol{\theta}^{(t)}$ to the participating clients,
and each client initializes its local QNN training from this global parameter
vector.

Each client $i \in \mathcal{C}$ holds a private local dataset
\begin{equation}
    \mathcal{D}_i
    =
    \left\{
    \left(\mathbf{x}_{i,j},y_{i,j}\right)
    \right\}_{j=1}^{n_i},
    \label{eq:client_dataset}
\end{equation}
where $n_i = |\mathcal{D}_i|$ is the number of local training samples. The client
datasets may be statistically heterogeneous and non-identically distributed. Let
\begin{equation}
    \mathcal{S}^{(t)} \subseteq \mathcal{C}
\end{equation}
denote the subset of clients participating in round $t$. Unless otherwise stated,
all subsequent expressions are written for participating clients
$i \in \mathcal{S}^{(t)}$.

\begin{figure*}[t]
    \centering
    \includegraphics[width=0.785\linewidth]{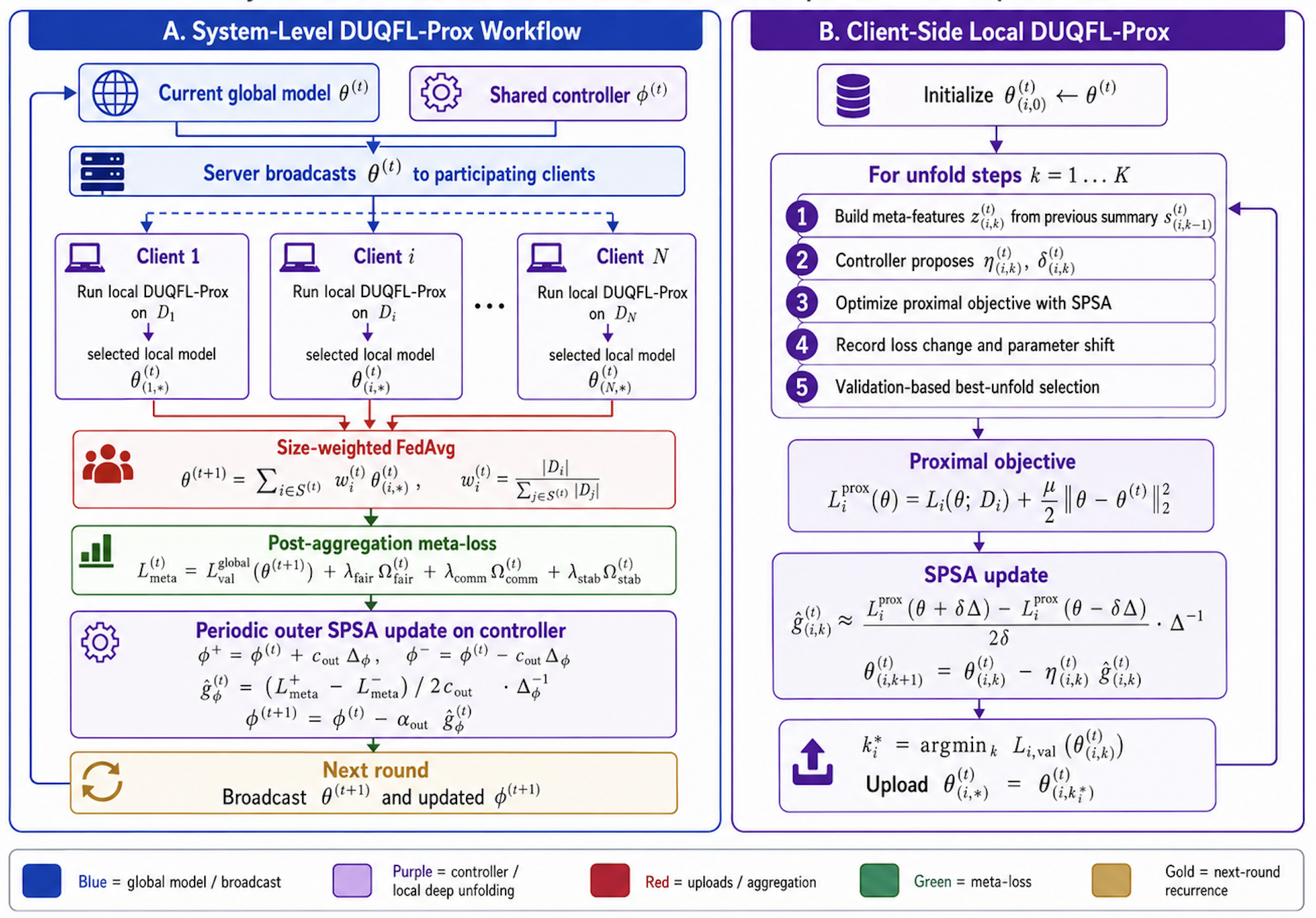}
    \caption{Overview of DUQFL-Prox. The server broadcasts global QNN parameters to clients, which perform deep-unfolded SPSA optimization with controller-generated hyperparameters, proximal drift control, and best-unfold selection. The selected local models are aggregated at the server, while a meta-objective periodically updates the shared controller using post-aggregation validation performance.}
    \label{fig:duqfl_prox_overview}
\end{figure*}

\subsection{Local Quantum Model}
\label{subsec:local_quantum_model}

Each participating client trains a parameterized quantum model represented as a
QNN with trainable parameter vector $\boldsymbol{\theta}$. Given an input
$\mathbf{x}$, the QNN applies a parameterized quantum circuit, followed by
measurement and classical post-processing, to produce either a class-probability
distribution or a predicted label. We denote the resulting client-side predictor
by
\begin{equation}
    f(\mathbf{x};\boldsymbol{\theta}).
\end{equation}

The empirical local objective at client $i$ is defined as
\begin{equation}
\mathcal{L}_i(\boldsymbol{\theta})
=
\frac{1}{n_i}
\sum_{(\mathbf{x},y)\in \mathcal{D}_i}
\ell
\left(
f(\mathbf{x};\boldsymbol{\theta}),y
\right),
\label{eq:local_loss}
\end{equation}
where $\ell(\cdot,\cdot)$ denotes the task-specific loss function, such as
cross-entropy loss for classification.

In standard QFL, client $i$ minimizes
$\mathcal{L}_i(\boldsymbol{\theta})$ using a fixed local optimizer initialized
from the broadcast global model $\boldsymbol{\theta}^{(t)}$. In contrast,
DUQFL-Prox treats local QNN training as a structured unfolded optimization
process. The local optimizer is decomposed into a finite sequence of adaptive
SPSA-based update blocks, allowing the learning rate and perturbation scale to
vary across clients, unfold steps, and communication rounds.

The proposed method is decomposed into three functional components.
Algorithm~\ref{alg:local_duqfl_prox} describes the client-side unfolded
proximal QNN training procedure. Algorithm~\ref{alg:federated_round_duqfl_prox}
describes one server-side communication round, including broadcast, local
training, aggregation, and meta-loss evaluation. Algorithm~\ref{alg:outer_controller_update}
describes the outer SPSA-based refinement of the shared controller.
This
decomposition makes the bilevel structure explicit: the inner level optimizes
local QNN parameters, 
while the outer level adapts the controller to improve
post-aggregation global behaviour.

\subsection{Deep-Unfolded Local Optimization}
\label{subsec:deep_unfolded_local_optimization}

\begin{algorithm}[t]
\caption{Client-Side Deep-Unfolded Proximal QNN Training}
\label{alg:local_duqfl_prox}
\begin{algorithmic}[1]
\REQUIRE Broadcast global parameters $\boldsymbol{\theta}^{(t)}$, client dataset $\mathcal{D}_i$, client validation set $\mathcal{V}_i$, controller $\boldsymbol{\phi}^{(t)}$, number of unfolds $K$, proximal coefficient $\mu$
\ENSURE Selected local model $\boldsymbol{\theta}_{i,\star}^{(t)}$, optimization summaries $\{\mathbf{s}_{i,k}^{(t)}\}_{k=1}^{K}$

\STATE Initialize local parameters:
$\boldsymbol{\theta}_{i,0}^{(t)} \leftarrow \boldsymbol{\theta}^{(t)}$
\STATE Initialize optimization summary $\mathbf{s}_{i,0}^{(t)}$
\STATE Initialize $L_{i,\star}^{(t)} \leftarrow \infty$ and $k_i^\star \leftarrow 0$

\FOR{$k=0,1,\dots,K-1$}
    \STATE Construct optimization-state feature vector:
    $
    \mathbf{z}_{i,k}^{(t)}
    \leftarrow
    \Psi
    \left(
    k,t,\mathbf{s}_{i,k}^{(t)},|\mathcal{D}_i|
    \right)
    $

    \STATE Generate unfold-specific SPSA hyperparameters:
    $
    \eta_{i,k}^{(t)},\delta_{i,k}^{(t)}
    \leftarrow
    \Gamma
    \left(
    \mathbf{z}_{i,k}^{(t)};\boldsymbol{\phi}^{(t)}
    \right)
    $

    \STATE Define the proximal local objective:
    $
    \mathcal{L}_{i}^{\mathrm{prox}}(\boldsymbol{\theta})
    =
    \mathcal{L}_{i}(\boldsymbol{\theta})
    +
    \frac{\mu}{2}
    \left\|
    \boldsymbol{\theta}-\boldsymbol{\theta}^{(t)}
    \right\|_2^2
    $

    \STATE Compute pre-update proximal loss:
    $
    L_{i,k}^{\mathrm{before}}
    \leftarrow
    \mathcal{L}_{i}^{\mathrm{prox}}
    \left(
    \boldsymbol{\theta}_{i,k}^{(t)}
    \right)
    $

    \STATE Apply one SPSA-based proximal QNN update:
    $
    \boldsymbol{\theta}_{i,k+1}^{(t)}
    \leftarrow
    \mathcal{U}_{\mathrm{SPSA}}
    \left(
    \boldsymbol{\theta}_{i,k}^{(t)};
    \eta_{i,k}^{(t)},
    \delta_{i,k}^{(t)},
    \mathcal{L}_{i}^{\mathrm{prox}}
    \right)
    $

    \STATE Compute post-update proximal loss:
    $
    L_{i,k}^{\mathrm{after}}
    \leftarrow
    \mathcal{L}_{i}^{\mathrm{prox}}
    \left(
    \boldsymbol{\theta}_{i,k+1}^{(t)}
    \right)
    $

    \STATE Compute unfold-step displacement:
    $
    \Delta\boldsymbol{\theta}_{i,k}^{(t)}
    \leftarrow
    \boldsymbol{\theta}_{i,k+1}^{(t)}
    -
    \boldsymbol{\theta}_{i,k}^{(t)}
    $

    \STATE Evaluate validation loss:
    $
    L_{i,k}^{\mathrm{val}}
    \leftarrow
    \mathcal{L}_{i,\mathrm{val}}
    \left(
    \boldsymbol{\theta}_{i,k+1}^{(t)};\mathcal{V}_i
    \right)
    $

    \IF{$L_{i,k}^{\mathrm{val}} < L_{i,\star}^{(t)}$}
        \STATE Update best unfolded state:
        $
        L_{i,\star}^{(t)} \leftarrow L_{i,k}^{\mathrm{val}},
        \quad
        k_i^\star \leftarrow k+1
        $
    \ENDIF

    \STATE Record optimization summary:
    $
    \mathbf{s}_{i,k+1}^{(t)}
    \leftarrow
    \mathrm{Summarize}
    \left(
    L_{i,k}^{\mathrm{before}},
    L_{i,k}^{\mathrm{after}},
    L_{i,k}^{\mathrm{val}},
    \Delta\boldsymbol{\theta}_{i,k}^{(t)}
    \right)
    $
\ENDFOR

\STATE Select best unfolded local model:
$
\boldsymbol{\theta}_{i,\star}^{(t)}
\leftarrow
\boldsymbol{\theta}_{i,k_i^\star}^{(t)}
$

\RETURN $\boldsymbol{\theta}_{i,\star}^{(t)}$, $\{\mathbf{s}_{i,k}^{(t)}\}_{k=1}^{K}$
\end{algorithmic}
\end{algorithm}

For each participating client $i \in \mathcal{S}^{(t)}$, the broadcast global
model initializes the local unfolded trajectory,
\begin{equation}
    \boldsymbol{\theta}_{i,0}^{(t)}
    \leftarrow
    \boldsymbol{\theta}^{(t)} .
    \label{eq:local_initialization}
\end{equation}
Local training is then unfolded into $K$ SPSA-based update blocks indexed by
$k \in \{0,1,\ldots,K-1\}$. Each block updates the local QNN parameters according to
\begin{equation}
    \boldsymbol{\theta}_{i,k+1}^{(t)}
    =
    \mathcal{U}_{\mathrm{SPSA}}
    \left(
    \boldsymbol{\theta}_{i,k}^{(t)};
    \eta_{i,k}^{(t)},
    \delta_{i,k}^{(t)},
    \mathcal{L}_{i}^{\mathrm{prox}}
    \right),
    \label{eq:unfolded_local_update}
\end{equation}
where $\eta_{i,k}^{(t)}$ and $\delta_{i,k}^{(t)}$ are the unfold-specific SPSA
learning rate and perturbation scale generated by the controller.

Rather than always returning the last unfolded state, the client selects the
validation-preferred checkpoint:
\begin{equation}
    k_i^\star
    =
    \arg\min_{k \in \{1,\ldots,K\}}
    \mathcal{L}_{i,\mathrm{val}}
    \left(
    \boldsymbol{\theta}_{i,k}^{(t)}
    \right),
    \label{eq:best_unfold_index}
\end{equation}
and uploads
\begin{equation}
    \boldsymbol{\theta}_{i,\star}^{(t)}
    =
    \boldsymbol{\theta}_{i,k_i^\star}^{(t)} .
    \label{eq:selected_unfold_model}
\end{equation}
The intermediate unfolded states remain local to the client and are used only
for trajectory construction, validation-based checkpoint selection, and
optimization-summary generation.

\subsection{Controller-Driven Hyperparameter Generation}
\label{subsec:controller_hyperparameter_generation}

The unfolded local optimizer requires step-specific SPSA hyperparameters for
each client and unfold step. Instead of using a fixed learning rate and
perturbation scale throughout local training, DUQFL-Prox uses a shared
meta-controller to generate these quantities adaptively. The controller is
parameterized by $\boldsymbol{\phi}^{(t)}$ and is shared across participating
clients at communication round $t$.

At unfold step $k$, client $i$ constructs an optimization-state feature vector
\begin{equation}
    \mathbf{z}_{i,k}^{(t)} \in \mathbb{R}^{d},
    \label{eq:controller_feature_vector}
\end{equation}
which summarizes the current local optimization context. In our implementation,
$\mathbf{z}_{i,k}^{(t)}$ includes normalized unfold progress, normalized
communication-round progress, previous loss information, previous parameter
displacement, client data fraction, and a client heterogeneity indicator.

 The controller maps this feature vector to the SPSA learning rate and
perturbation scale:
\begin{equation}
    \left(
    \eta_{i,k}^{(t)},
    \delta_{i,k}^{(t)}
    \right)
    =
    \Gamma
    \left(
    \mathbf{z}_{i,k}^{(t)};
    \boldsymbol{\phi}^{(t)}
    \right),
    \label{eq:controller_mapping}
\end{equation}
where $\Gamma(\cdot)$ denotes the controller mapping.

In this work, $\Gamma(\cdot)$ is implemented as a clipped log-linear controller.
Two parameter vectors, $\boldsymbol{\phi}_{\eta}^{(t)}$ and
$\boldsymbol{\phi}_{\delta}^{(t)}$, generate the logarithmic learning rate and
perturbation scale:
\begin{align}
    \log \eta_{i,k}^{(t)}
    &=
    \left(\boldsymbol{\phi}_{\eta}^{(t)}\right)^{\top}
    \mathbf{z}_{i,k}^{(t)}, \\
    \log \delta_{i,k}^{(t)}
    &=
    \left(\boldsymbol{\phi}_{\delta}^{(t)}\right)^{\top}
    \mathbf{z}_{i,k}^{(t)}.
\end{align}
The resulting values are exponentiated and clipped to predefined feasible
intervals:
\begin{align}
    \eta_{i,k}^{(t)}
    &=
    \mathrm{clip}
    \left(
    \exp
    \left(
    \left(\boldsymbol{\phi}_{\eta}^{(t)}\right)^{\top}
    \mathbf{z}_{i,k}^{(t)}
    \right),
    \eta_{\min},
    \eta_{\max}
    \right), \\
    \delta_{i,k}^{(t)}
    &=
    \mathrm{clip}
    \left(
    \exp
    \left(
    \left(\boldsymbol{\phi}_{\delta}^{(t)}\right)^{\top}
    \mathbf{z}_{i,k}^{(t)}
    \right),
    \delta_{\min},
    \delta_{\max}
    \right).
\end{align}

This controller provides the adaptive component of DUQFL-Prox by allowing the
local SPSA behaviour to vary across clients, unfold steps, and communication
rounds. The clipping bounds prevent numerically unstable hyperparameter values,
while the proximal objective in Section~\ref{subsec:proximal_drift_control}
provides the stabilizing mechanism that restricts excessive client drift.

\subsection{Proximal Drift Control}
\label{subsec:proximal_drift_control}

Client drift is a central challenge in federated learning, particularly under
non-IID data distributions. In QFL, this issue is further amplified by
stochastic SPSA updates, finite-shot measurement effects, and the non-convex
loss landscape of variational quantum circuits. A client may reduce its local
training loss while producing a parameter update that becomes overly specialized
to its own local distribution and less compatible with global aggregation.

DUQFL-Prox addresses this issue by introducing a proximal penalty into the local
client objective. Let $\boldsymbol{\theta}^{(t)}$ denote the global model
broadcast by the server at communication round $t$. For client $i$, the
proximal local objective is defined as
\begin{equation}
\mathcal{L}_i^{\mathrm{prox}}(\boldsymbol{\theta})
=
\mathcal{L}_i(\boldsymbol{\theta})
+
\frac{\mu}{2}
\left\|
\boldsymbol{\theta}
-
\boldsymbol{\theta}^{(t)}
\right\|_2^2,
\label{eq:prox_local_objective}
\end{equation}
where $\mathcal{L}_i(\boldsymbol{\theta})$ is the empirical local QNN loss and
$\mu \geq 0$ controls the strength of proximal regularization.

Within each unfolded local optimization step, SPSA minimizes
$\mathcal{L}_i^{\mathrm{prox}}$ rather than the unregularized local objective.
Thus, the proximal unfolded update is written as
\begin{equation}
\boldsymbol{\theta}_{i,k+1}^{(t)}
=
\mathcal{U}_{\mathrm{SPSA}}
\left(
\boldsymbol{\theta}_{i,k}^{(t)};
\eta_{i,k}^{(t)},
\delta_{i,k}^{(t)},
\mathcal{L}_i^{\mathrm{prox}}
\right).
\label{eq:prox_spsa_update}
\end{equation}

The proximal term encourages local models to improve their client-specific
objective while remaining close to the current global reference point. This is
important in the unfolded setting because multiple adaptive local update blocks
can improve flexibility, but may also increase the risk of excessive client
movement. DUQFL-Prox therefore combines an adaptive component, provided by the
controller-generated SPSA hyperparameters, with a stabilizing component,
provided by proximal regularization.

To monitor local movement, we record the unfold-step displacement
\begin{equation}
\Delta \boldsymbol{\theta}_{i,k}^{(t)}
=
\boldsymbol{\theta}_{i,k+1}^{(t)}
-
\boldsymbol{\theta}_{i,k}^{(t)},
\label{eq:local_update_vector}
\end{equation}
and its norm
\begin{equation}
\left\|
\Delta \boldsymbol{\theta}_{i,k}^{(t)}
\right\|_2 .
\label{eq:local_update_norm}
\end{equation}
This quantity is used as a diagnostic indicator of local client movement and can
also be included in the controller state features for subsequent unfold steps.
Overall, the proximal formulation helps produce adaptive yet
aggregation-compatible local QNN updates under heterogeneous federated data.

\subsection{Best-Unfold Model Selection}
\label{subsec:best_unfold_selection}

The last unfolded state is not necessarily the best local model to upload. Later
unfold steps may continue to reduce local training loss while degrading
validation behaviour or increasing local specialization. Therefore, DUQFL-Prox
uses validation-based checkpoint selection over the local unfolded trajectory.

After generating the candidate states
\begin{equation}
    \left\{
    \boldsymbol{\theta}_{i,k}^{(t)}
    \right\}_{k=1}^{K},
\end{equation}
client $i$ selects the unfold index with the lowest validation loss:
\begin{equation}
    k_i^\star
    =
    \arg\min_{k\in\{1,\ldots,K\}}
    \mathcal{L}_{i,\mathrm{val}}
    \left(
    \boldsymbol{\theta}_{i,k}^{(t)}
    \right).
    \label{eq:best_unfold_selection}
\end{equation}
The selected local model is then
\begin{equation}
    \boldsymbol{\theta}_{i,\star}^{(t)}
    =
    \boldsymbol{\theta}_{i,k_i^\star}^{(t)}.
    \label{eq:selected_unfolded_model}
\end{equation}

Only $\boldsymbol{\theta}_{i,\star}^{(t)}$ is uploaded to the server. The
remaining unfolded states are retained locally and are used only for trajectory
construction, checkpoint selection, and optimization-summary generation. This
selection mechanism prevents over-aggressive later unfold steps from dominating
the uploaded client update.

\begin{algorithm}[t]
\caption{One Federated Round of DUQFL-Prox}
\label{alg:federated_round_duqfl_prox}
\begin{algorithmic}[1]
\REQUIRE Client set $\mathcal{C}$, current global model $\boldsymbol{\theta}^{(t)}$, current controller $\boldsymbol{\phi}^{(t)}$, participating clients $\mathcal{S}^{(t)}$, number of unfolds $K$, proximal coefficient $\mu$
\ENSURE Updated global model $\boldsymbol{\theta}^{(t+1)}$, outer meta-loss $\mathcal{L}_{\mathrm{meta}}^{(t)}$

\STATE Server broadcasts $\boldsymbol{\theta}^{(t)}$ to all clients in $\mathcal{S}^{(t)}$

\FORALL{clients $i \in \mathcal{S}^{(t)}$ \textbf{in parallel}}
    \STATE Execute Algorithm~\ref{alg:local_duqfl_prox}:
    $
    \left(
    \boldsymbol{\theta}_{i,\star}^{(t)},
    \{\mathbf{s}_{i,k}^{(t)}\}_{k=1}^{K}
    \right)
    \leftarrow
    \mathrm{LocalDUQFLProx}
    \left(
    \boldsymbol{\theta}^{(t)},
    \mathcal{D}_i,
    \mathcal{V}_i,
    \boldsymbol{\phi}^{(t)},
    K,
    \mu
    \right)
    $
\ENDFOR

\STATE Compute sample-size aggregation weights:
$
w_i^{(t)}
=
\frac{|\mathcal{D}_i|}
{\sum_{j\in\mathcal{S}^{(t)}}|\mathcal{D}_j|}
$

\STATE Aggregate selected unfolded local models:
$
\boldsymbol{\theta}^{(t+1)}
\leftarrow
\sum_{i\in\mathcal{S}^{(t)}}
w_i^{(t)}
\boldsymbol{\theta}_{i,\star}^{(t)}
$

\STATE Evaluate post-aggregation meta-loss:
$
\mathcal{L}_{\mathrm{meta}}^{(t)}
\leftarrow
\mathcal{L}_{\mathrm{val}}^{\mathrm{global}}
\left(
\boldsymbol{\theta}^{(t+1)}
\right)
+
\lambda_{\mathrm{fair}}
\Omega_{\mathrm{fair}}^{(t)}
+
\lambda_{\mathrm{comm}}
\Omega_{\mathrm{comm}}^{(t)}
+
\lambda_{\mathrm{stab}}
\Omega_{\mathrm{stab}}^{(t)}
$
\RETURN $\boldsymbol{\theta}^{(t+1)}$, $\mathcal{L}_{\mathrm{meta}}^{(t)}$
\end{algorithmic}
\end{algorithm}

\begin{algorithm}[t]
\caption{Outer SPSA-Based Controller Update}
\label{alg:outer_controller_update}
\begin{algorithmic}[1]
\REQUIRE Current controller $\boldsymbol{\phi}^{(t)}$, current global model $\boldsymbol{\theta}^{(t)}$, participating clients $\mathcal{S}^{(t)}$, outer learning rate $\alpha_{\mathrm{out}}$, outer perturbation radius $c_{\mathrm{out}}$
\ENSURE Updated controller $\boldsymbol{\phi}^{(t+1)}$

\STATE Sample Rademacher perturbation vector $\boldsymbol{\Delta}_{\phi}$

\STATE Construct perturbed controllers:
$
\boldsymbol{\phi}^{+}
=
\boldsymbol{\phi}^{(t)}
+
c_{\mathrm{out}}\boldsymbol{\Delta}_{\phi},
\qquad
\boldsymbol{\phi}^{-}
=
\boldsymbol{\phi}^{(t)}
-
c_{\mathrm{out}}\boldsymbol{\Delta}_{\phi}
$

\STATE Evaluate one federated round under $\boldsymbol{\phi}^{+}$ using Algorithm~\ref{alg:federated_round_duqfl_prox} to obtain:
$
\mathcal{L}_{\mathrm{meta}}^{+}
$

\STATE Evaluate one federated round under $\boldsymbol{\phi}^{-}$ using Algorithm~\ref{alg:federated_round_duqfl_prox} to obtain:
$
\mathcal{L}_{\mathrm{meta}}^{-}
$

\STATE Estimate the outer SPSA gradient:
$
\widehat{\mathbf{g}}_{\phi}^{(t)}
\leftarrow
\frac{
\mathcal{L}_{\mathrm{meta}}^{+}
-
\mathcal{L}_{\mathrm{meta}}^{-}
}{
2c_{\mathrm{out}}
}
\boldsymbol{\Delta}_{\phi}^{-1}
$

\STATE Update the controller:
$
\boldsymbol{\phi}^{(t+1)}
\leftarrow
\boldsymbol{\phi}^{(t)}
-
\alpha_{\mathrm{out}}
\widehat{\mathbf{g}}_{\phi}^{(t)}
$

\RETURN $\boldsymbol{\phi}^{(t+1)}$
\end{algorithmic}
\end{algorithm}

\subsection{Server Aggregation}
\label{subsec:server_aggregation}

After receiving the selected unfolded local models from participating clients,
the server constructs the next global model using sample-size-weighted FedAvg:
\begin{equation}
    \boldsymbol{\theta}^{(t+1)}
    =
    \sum_{i \in \mathcal{S}^{(t)}}
    w_i^{(t)}
    \boldsymbol{\theta}_{i,\star}^{(t)},
    \label{eq:fedavg_selected_aggregation}
\end{equation}
where $w_i^{(t)}$ denotes the aggregation weight of client $i$. The weights
satisfy
\begin{equation}
    w_i^{(t)} \geq 0,
    \qquad
    \sum_{i \in \mathcal{S}^{(t)}} w_i^{(t)} = 1 .
\end{equation}
In this work, the sample-size-weighted coefficient is
\begin{equation}
    w_i^{(t)}
    =
    \frac{n_i}
    {\sum_{j \in \mathcal{S}^{(t)}} n_j},
    \label{eq:sample_size_weight}
\end{equation}
where $n_i = |\mathcal{D}_i|$ is the number of local training samples available
to client $i$ in the current round.

Thus, the server does not aggregate every unfolded state. It aggregates only the
validation-selected local checkpoint $\boldsymbol{\theta}_{i,\star}^{(t)}$ from
each participating client.

\subsection{Outer Meta-Objective and Controller Update}
\label{subsec:outer_meta_objective}

The controller is not treated as a fixed hyperparameter generator. Instead, it is
periodically updated using an outer meta-objective evaluated after server
aggregation. Let $\mathcal{D}_{\mathrm{val}}$ denote the validation data used to
measure post-aggregation global performance. At communication round $t$, the
outer meta-loss is defined as
\begin{equation}
    \mathcal{L}_{\mathrm{meta}}^{(t)}
    =
    \mathcal{L}_{\mathrm{val}}^{\mathrm{global}}
    \left(
    \boldsymbol{\theta}^{(t+1)}
    \right)
    +
    \lambda_{\mathrm{fair}}
    \Omega_{\mathrm{fair}}^{(t)}
    +
    \lambda_{\mathrm{comm}}
    \Omega_{\mathrm{comm}}^{(t)}
    +
    \lambda_{\mathrm{stab}}
    \Omega_{\mathrm{stab}}^{(t)} .
    \label{eq:meta_loss}
\end{equation}
Here, $\mathcal{L}_{\mathrm{val}}^{\mathrm{global}}$ is the validation loss of
the aggregated global model, $\Omega_{\mathrm{fair}}^{(t)}$ measures
client-level performance imbalance, $\Omega_{\mathrm{comm}}^{(t)}$ accounts for
communication cost, and $\Omega_{\mathrm{stab}}^{(t)}$ penalizes unstable
optimization trajectories. The coefficients
$\lambda_{\mathrm{fair}}$, $\lambda_{\mathrm{comm}}$, and
$\lambda_{\mathrm{stab}}$ control the relative contribution of these terms.

To update the controller, DUQFL-Prox applies an outer SPSA step to
$\boldsymbol{\phi}^{(t)}$. A Rademacher perturbation vector
$\boldsymbol{\Delta}_{\phi}$ is sampled, and two perturbed controllers are
formed:
\begin{equation}
    \boldsymbol{\phi}^{+}
    =
    \boldsymbol{\phi}^{(t)}
    +
    c_{\mathrm{out}}
    \boldsymbol{\Delta}_{\phi},
    \qquad
    \boldsymbol{\phi}^{-}
    =
    \boldsymbol{\phi}^{(t)}
    -
    c_{\mathrm{out}}
    \boldsymbol{\Delta}_{\phi}.
    \label{eq:outer_controller_perturbation}
\end{equation}
The corresponding perturbed meta-losses,
$\mathcal{L}_{\mathrm{meta}}^{+}$ and
$\mathcal{L}_{\mathrm{meta}}^{-}$, are obtained by evaluating virtual
federated rounds under $\boldsymbol{\phi}^{+}$ and
$\boldsymbol{\phi}^{-}$, respectively. The outer SPSA gradient estimate is then
\begin{equation}
    \widehat{\mathbf{g}}_{\phi}^{(t)}
    =
    \frac{
    \mathcal{L}_{\mathrm{meta}}^{+}
    -
    \mathcal{L}_{\mathrm{meta}}^{-}
    }
    {2c_{\mathrm{out}}}
    \boldsymbol{\Delta}_{\phi}^{-1}.
    \label{eq:outer_spsa_gradient}
\end{equation}
Since each perturbation entry satisfies
$\Delta_{\phi,j}\in\{-1,+1\}$, we have
$\Delta_{\phi,j}^{-1}=\Delta_{\phi,j}$.

The controller is updated as
\begin{equation}
    \boldsymbol{\phi}^{(t+1)}
    =
    \boldsymbol{\phi}^{(t)}
    -
    \alpha_{\mathrm{out}}
    \widehat{\mathbf{g}}_{\phi}^{(t)},
    \label{eq:controller_update}
\end{equation}
where $\alpha_{\mathrm{out}}$ is the outer learning rate.

This establishes a bilevel learning structure. The inner level performs
client-side deep-unfolded proximal SPSA optimization of QNN parameters, while
the outer level adapts the shared controller so that future local optimization
trajectories become more compatible with post-aggregation global performance.
The outer update is applied periodically rather than necessarily at every round,
which reduces computational overhead while still allowing the controller to
improve across training.

\section{Experimental Setup}
\label{sec:experimental_setup}
This section describes the implementation environment, quantum model configuration,
datasets, federated partitioning strategy, baseline methods, evaluation metrics,
and reproducibility protocol used to evaluate DUQFL-Prox. The objective of the
experiments is to assess not only final global accuracy, but also client-level
generalization, train-test gap, client fairness, and stability under
heterogeneous federated data.

\subsection{Implementation Details}
\label{subsec:implementation_details}

All experiments were implemented in Python using Visual Studio Code as the main
development environment. Quantum models were implemented using Qiskit and
Qiskit Machine Learning, with Qiskit Aer used for simulator-based experiments.
The main Python libraries used were NumPy, pandas, scikit-learn, matplotlib,
seaborn, Qiskit, Qiskit Aer, and Qiskit Machine Learning. Fixed random seeds
were used for dataset splitting, client partitioning, QNN parameter
initialization, and optimizer-related stochasticity wherever supported.

The main experimental configuration is summarized in
Table~\ref{tab:experimental_config}. The reported values correspond to the
default configuration used in the main experiments; dataset-specific adjustments
were made when necessary due to dataset capacity or computational constraints.

\begin{table}[htbp]
\centering
\caption{Main experimental configuration used in DUQFL-Prox experiments.}
\label{tab:experimental_config}
\begin{tabular}{ll}
\hline
\textbf{Setting} & \textbf{Value} \\
\hline
Programming language & Python 3.11.14 \\
Development environment & Visual Studio Code \\
Quantum framework & Qiskit, Qiskit Machine Learning \\
Simulator & Qiskit Aer simulator \\
Feature map & ZZFeatureMap \\
Ansatz & RealAmplitudes \\
Optimizer & SPSA / DUQFL-Prox SPSA \\
Number of clients & 5 or 10 \\
Communication rounds & 10--50 depending on dataset capacity \\
Unfold steps & $K=5$ \\
SPSA iterations per unfold & 5--10 \\
Initial learning rate & 0.13 \\
Initial perturbation & 0.13 \\
Proximal coefficient & $\mu=10^{-2}$ \\
Shots & 1024 \\
\hline
\end{tabular}
\end{table}

\subsection{Quantum Neural Network Architecture}
\label{subsec:qnn_architecture}

Each client model was implemented as a parameterized QNN. The classical input
features were first preprocessed into a low-dimensional representation and then
encoded into a quantum circuit using a ZZFeatureMap. A RealAmplitudes ansatz was
used as the trainable variational circuit. The number of qubits was determined
by the number of retained input features after preprocessing. In the main
experiments, we used either two or four QNN input features depending on the
dataset and computational budget.

Let $\mathbf{x}\in\mathbb{R}^{d}$ denote the preprocessed input vector. The QNN
maps $\mathbf{x}$ to a quantum state through the feature map and then applies a
trainable ansatz parameterized by $\boldsymbol{\theta}$. Measurement outcomes
are classically post-processed to obtain class probabilities or predicted
labels. All compared methods used the same QNN architecture for a given dataset
to ensure a fair comparison.

\subsection{Datasets and Preprocessing}
\label{subsec:datasets_preprocessing}

The proposed DUQFL-Prox framework was evaluated on datasets from genomics and
financial fraud detection. Since current QNN models are constrained by the
number of available qubits and circuit depth, each dataset was transformed into
a low-dimensional QNN-compatible representation before federated training.
Table~\ref{tab:datasets} summarizes the dataset usage,
preprocessing pipeline, and final QNN input dimension.
\begin{table}[htbp]
\centering
\caption{Datasets used for evaluating DUQFL-Prox across multiple domains.}
\label{tab:datasets}
\begin{tabular}{lllll}
\hline
\textbf{Dataset} & \textbf{Domain} & \textbf{Task} & \textbf{Source} & \textbf{QNN features} \\
\hline
BAF & Finance & Bank fraud detection 
& \cite{jesus2022turningBAF} & 4 \\

Genome & Genomics & Genomic sequencing 
& \cite{grevsova2023genomic} & 2 or 4 \\
\hline
\end{tabular}
\end{table}
The BAF dataset is taken from the Bank Account Fraud Dataset Suite introduced by
Jesus et al.~\cite{jesus2022turningBAF}. The dataset suite was designed as a
privacy-preserving, large-scale tabular benchmark for bank account-opening fraud
detection, with realistic challenges including temporal dynamics, severe class
imbalance, and bias/fairness-related distributional shifts.
The Genome experiments use the DemoHumanOrWorm task from the Genomic Benchmarks
suite~\cite{grevsova2023genomic}. Genomic Benchmarks provides curated datasets
for genomic sequence classification and offers standardized access through
common machine-learning and deep-learning interfaces.

\subsubsection{BAF Dataset}

For the BAF, Bank Account fraud-detection experiment, the target variable was
\texttt{fraud\_bool}. After removing missing values, the dataset was optionally
subsampled using stratified sampling to preserve the class distribution under
the QNN computational budget. The data were split into training, validation, and
test sets before feature encoding and scaling to avoid data leakage. Categorical
variables were encoded using ordinal encoding with support for unseen
categories, while numerical variables were standardized. PCA was then applied to
obtain a low-dimensional representation compatible with the number of QNN
qubits. Finally, the PCA features were optionally scaled to the quantum angle
range $[0,\pi]$. In the reported BAF run, the resulting split contained 2,999
training samples, 750 validation samples, and 1,250 test samples, using four QNN
features/qubits.
\subsubsection{Genome Dataset}

For the Genome experiment, we used the DemoHumanOrWorm task from the Genomic
Benchmarks suite. Each DNA sequence was converted into a numerical vector using
a fixed word-size encoding strategy. Specifically, all unique words of length
\texttt{word\_size} were assigned integer identifiers, and each sequence was
represented by the corresponding sequence of integer word indices. The resulting
vectors were shuffled and scaled using MinMax scaling to obtain bounded
QNN-compatible input features. In the implementation used in this work, 10,000
processed records were used for training and 2,000 records were used for
testing.

The BAF and Genome datasets are used as representative intelligent-service tasks because they capture two unique privacy-sensitive distributed decision setting challenges: \textit{financial fraud detection and genomic classification}.
\section{Experimental Results and Analysis}
\label{sec:experimental_results}

\begin{figure*}[t]
\centering

\begin{subfigure}[t]{0.245\textwidth}
    \centering
    \includegraphics[width=\linewidth]{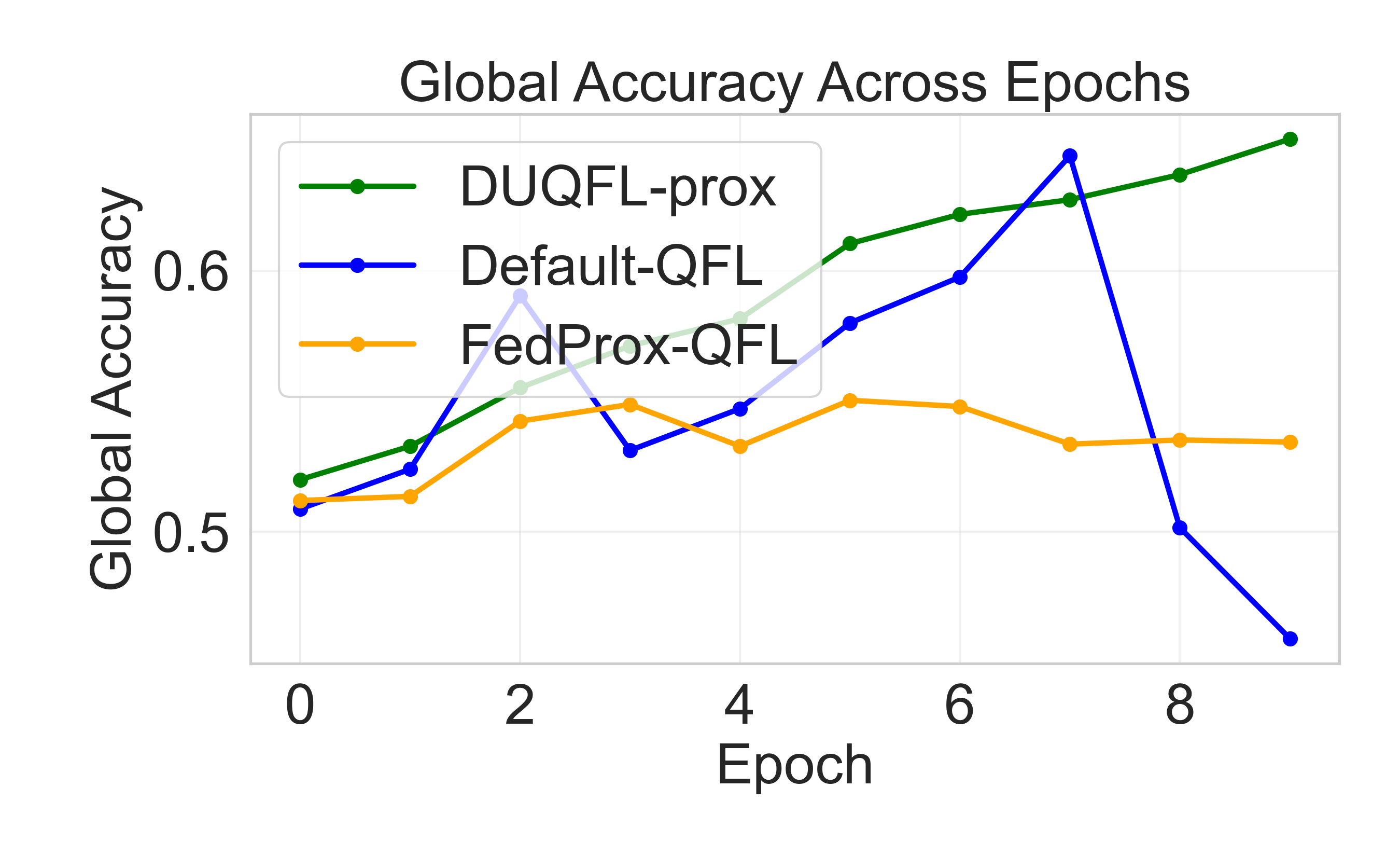}
    \caption{Global accuracy}
    \label{fig:baf_global_accuracy}
\end{subfigure}\hfill
\begin{subfigure}[t]{0.245\textwidth}
    \centering
    \includegraphics[width=\linewidth]{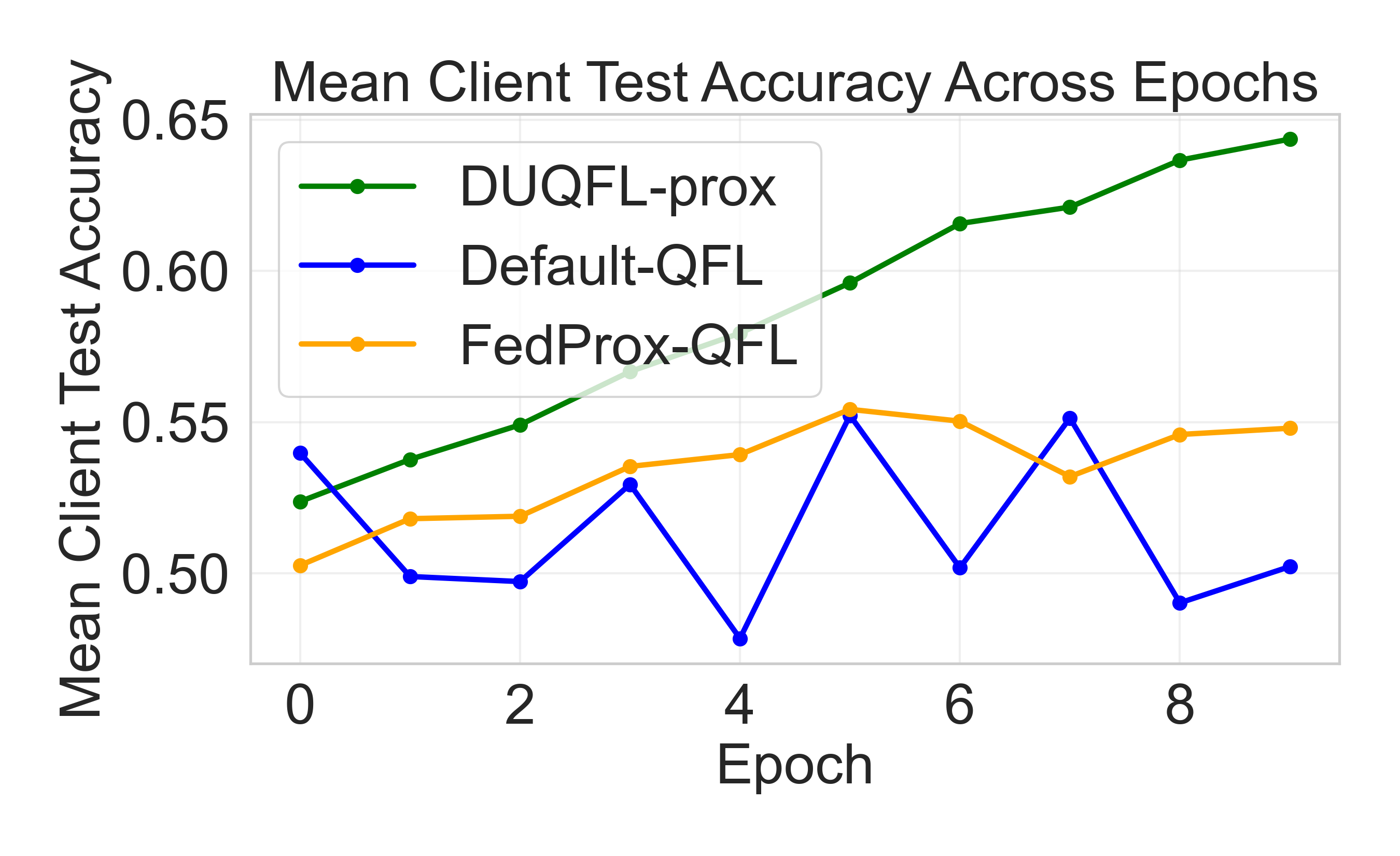}
    \caption{Mean client test accuracy}
    \label{fig:baf_mean_client_test}
\end{subfigure}\hfill
\begin{subfigure}[t]{0.245\textwidth}
    \centering
    \includegraphics[width=\linewidth]{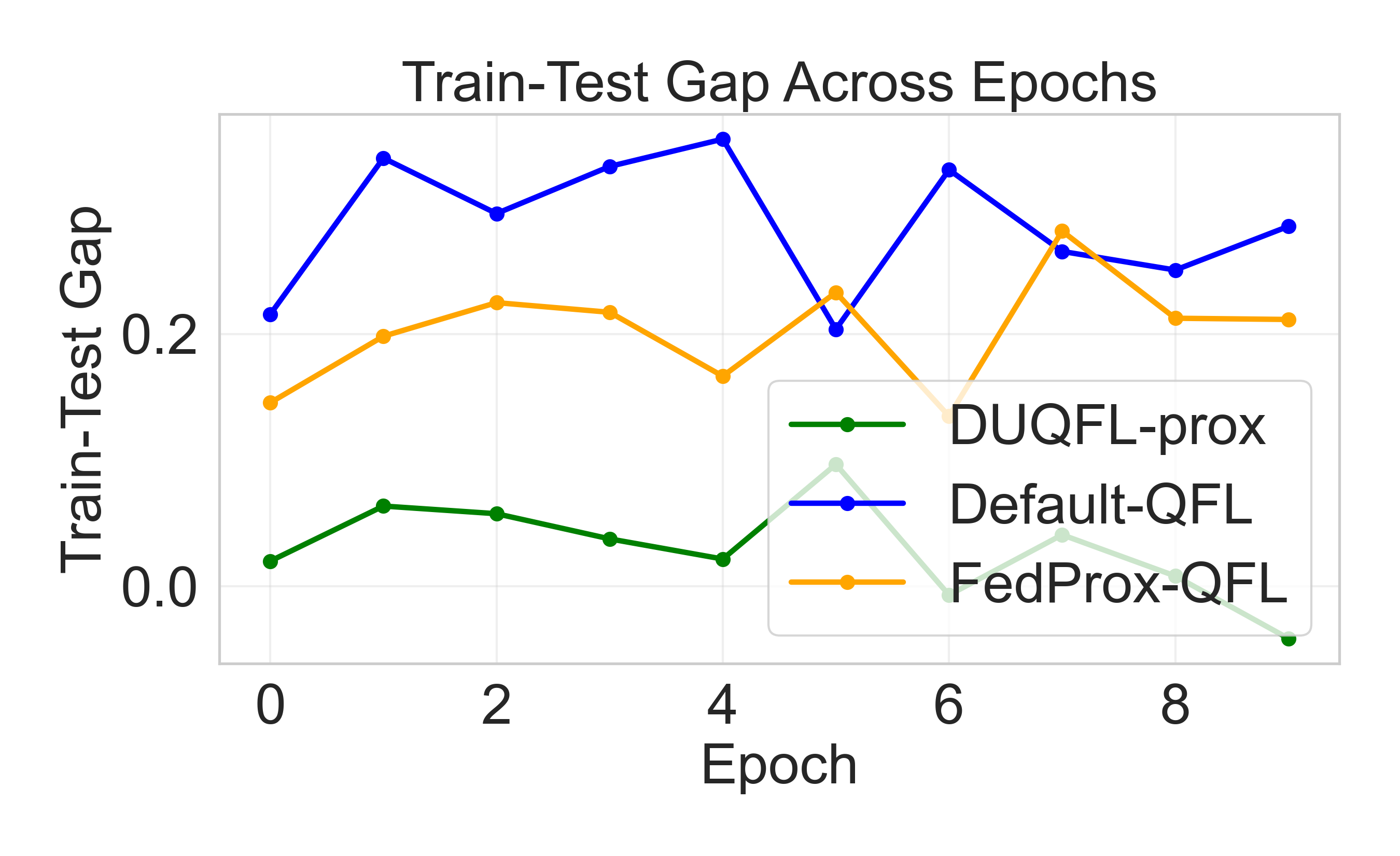}
    \caption{Train--test gap}
    \label{fig:baf_train_test_gap}
\end{subfigure}\hfill
\begin{subfigure}[t]{0.245\textwidth}
    \centering
    \includegraphics[width=\linewidth]{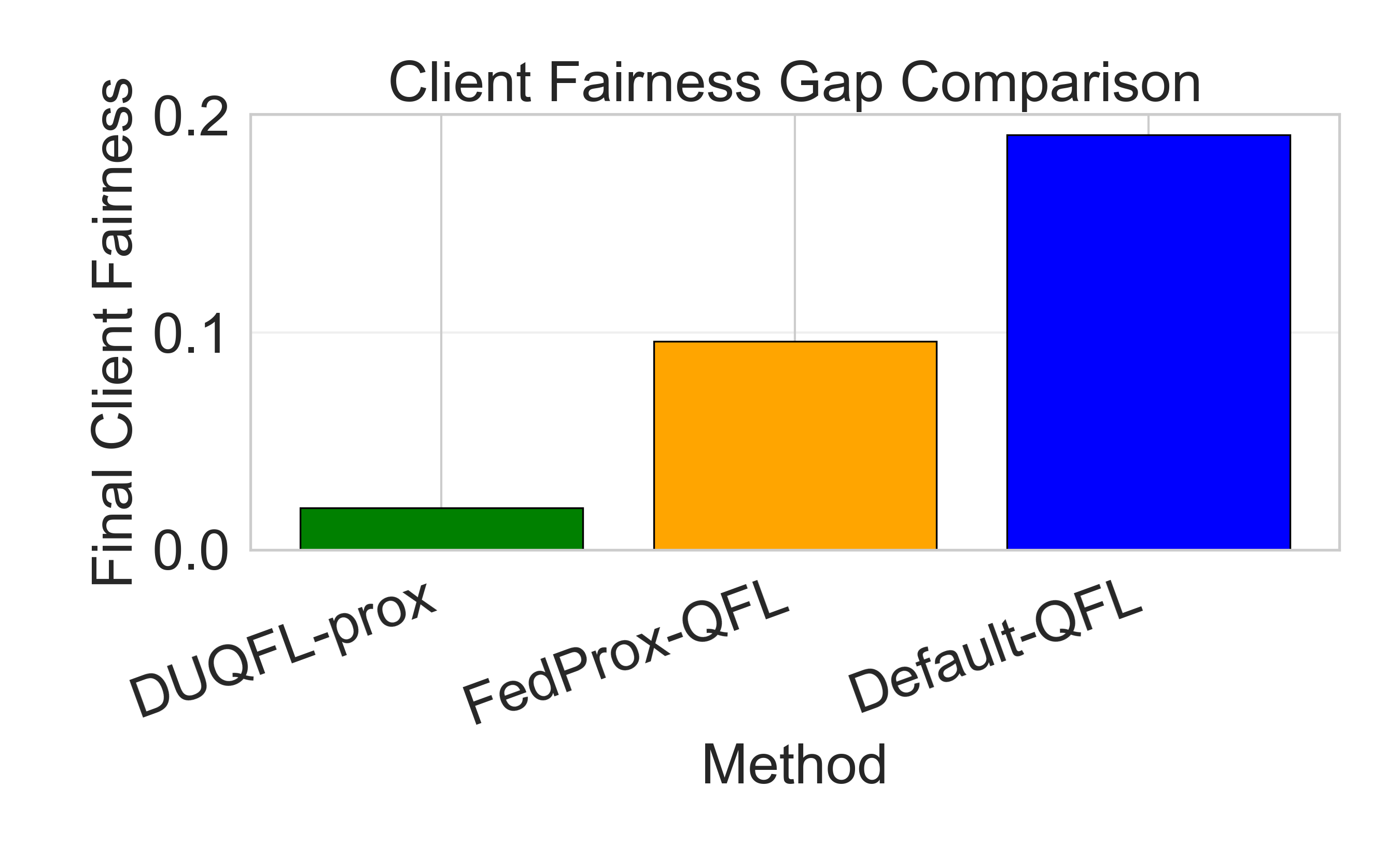}
    \caption{Client fairness gap}
    \label{fig:baf_client_fairness_gap}
\end{subfigure}

\caption{Federated performance comparison on the BAF dataset. DUQFL-Prox improves
final global accuracy and mean client test accuracy while substantially reducing
the train--test gap and client fairness gap. Lower values are preferable for
train--test gap and fairness gap.}
\label{fig:baf_global_client_metrics}
\end{figure*}

\begin{figure*}[h]
\centering

\begin{subfigure}[t]{0.245\textwidth}
    \centering
    \includegraphics[width=\linewidth]{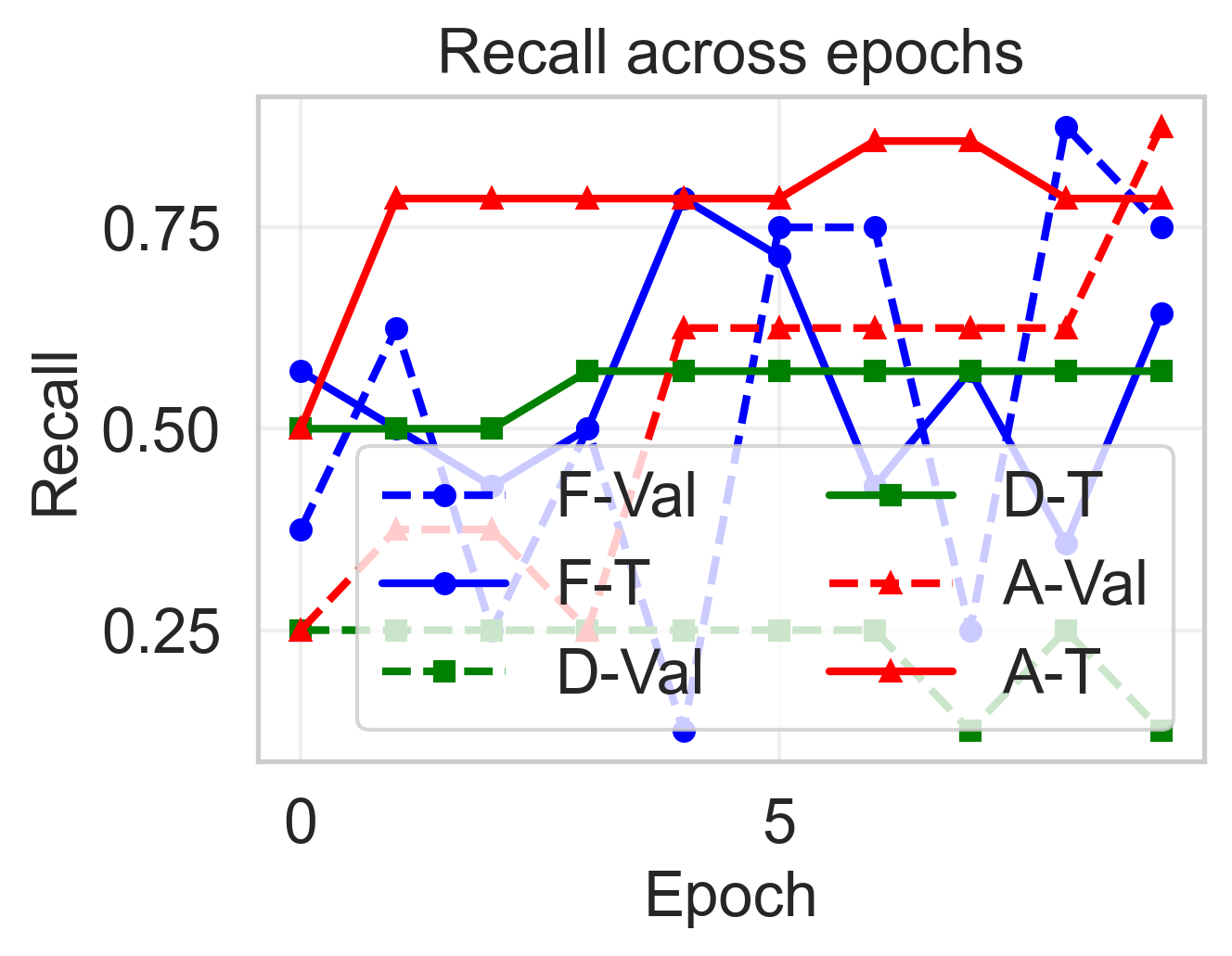}
    \caption{Recall}
    \label{fig:recall}
\end{subfigure}\hfill
\begin{subfigure}[t]{0.245\textwidth}
    \centering
    \includegraphics[width=\linewidth]{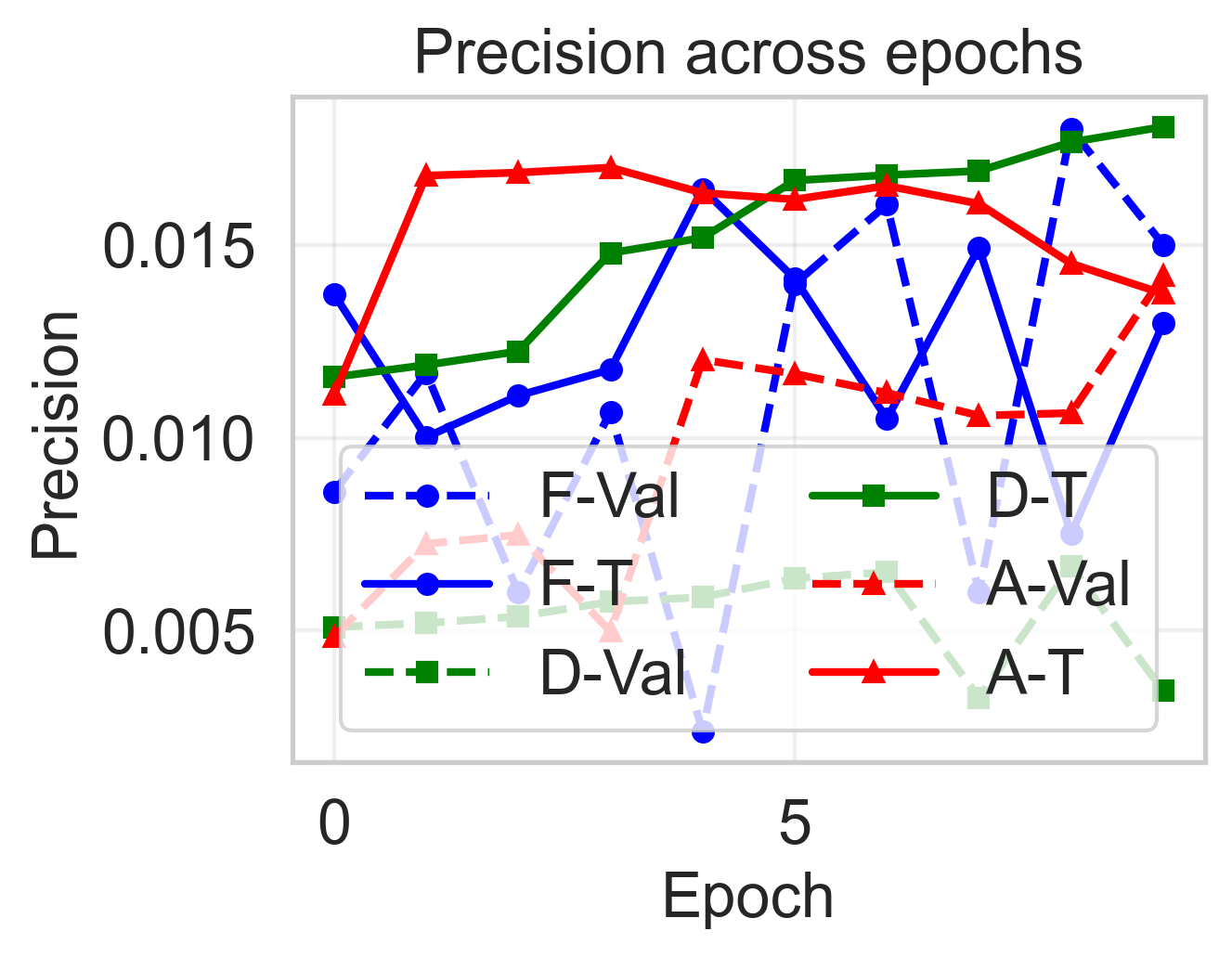}
    \caption{precision}
    \label{fig:precision_epochs}
\end{subfigure}\hfill
\begin{subfigure}[t]{0.245\textwidth}
    \centering
    \includegraphics[width=\linewidth]{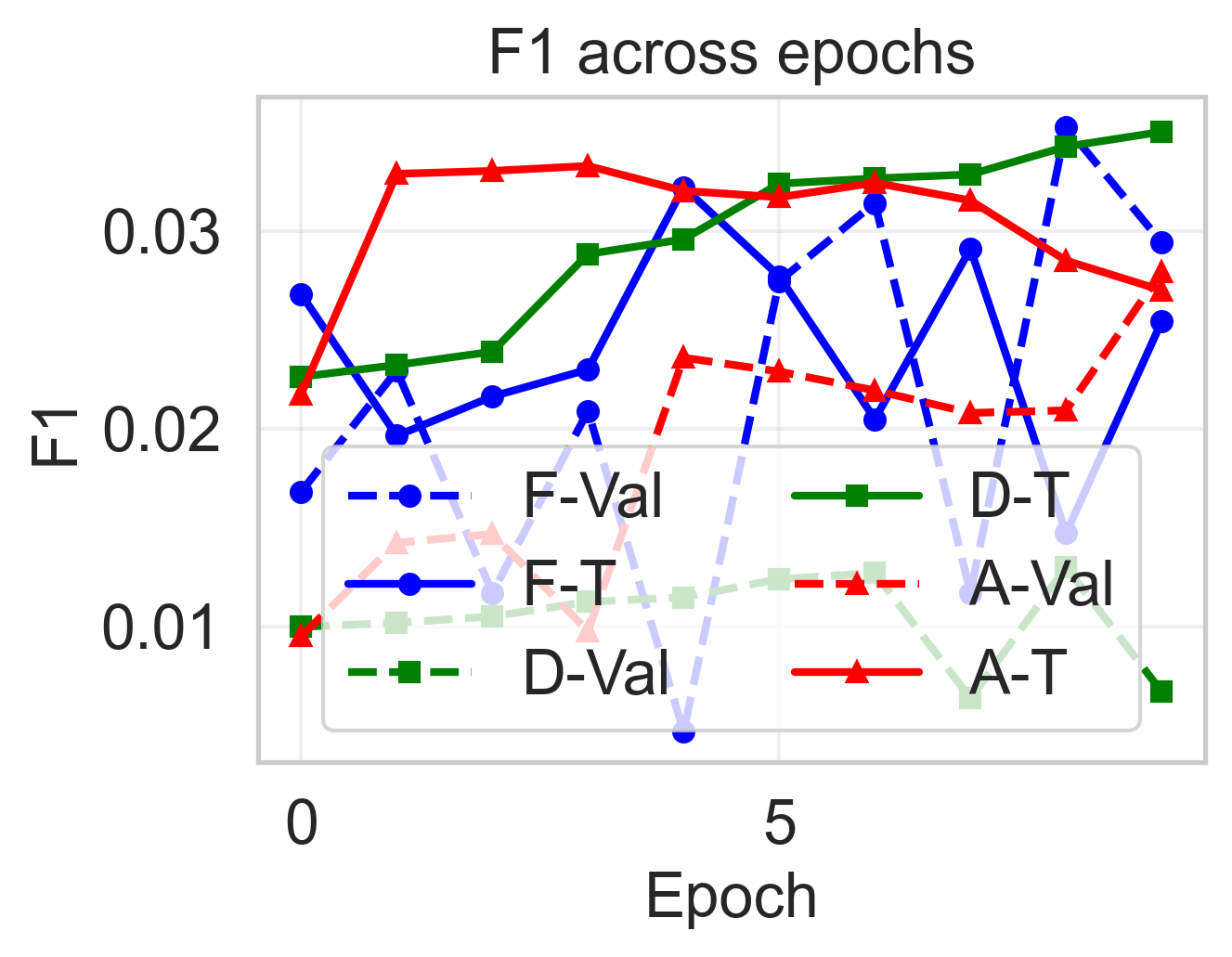}
    \caption{F1}
    \label{fig:f1_epochs}
\end{subfigure}\hfill
\begin{subfigure}[t]{0.245\textwidth}
    \centering
    \includegraphics[width=\linewidth]{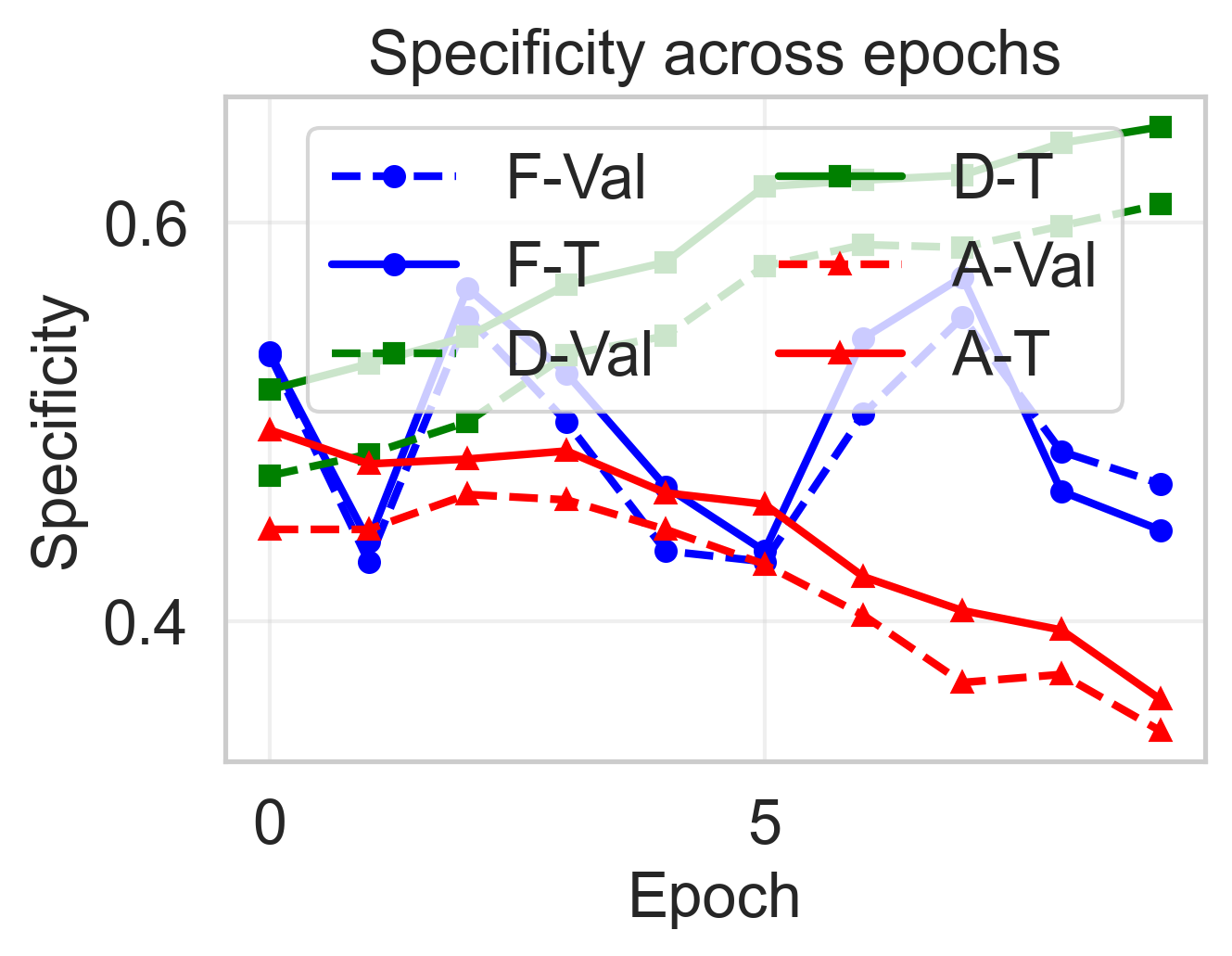}
    \caption{Specificity}
    \label{fig:specificity_epochs}
\end{subfigure}

\caption{Epoch-wise ROC-AUC, PR-AUC, MCC, and specificity trajectories on the BAF
dataset. DUQFL-Prox shows more stable late-epoch behaviour on test ROC-AUC,
PR-AUC, and MCC, while Adam-QFL exhibits a recall-heavy operating regime with
declining specificity.}
\label{fig:metrics_BAF_1}
\end{figure*}

\begin{figure*}[h]
\centering

\begin{subfigure}[t]{0.245\textwidth}
    \centering
    \includegraphics[width=\linewidth]{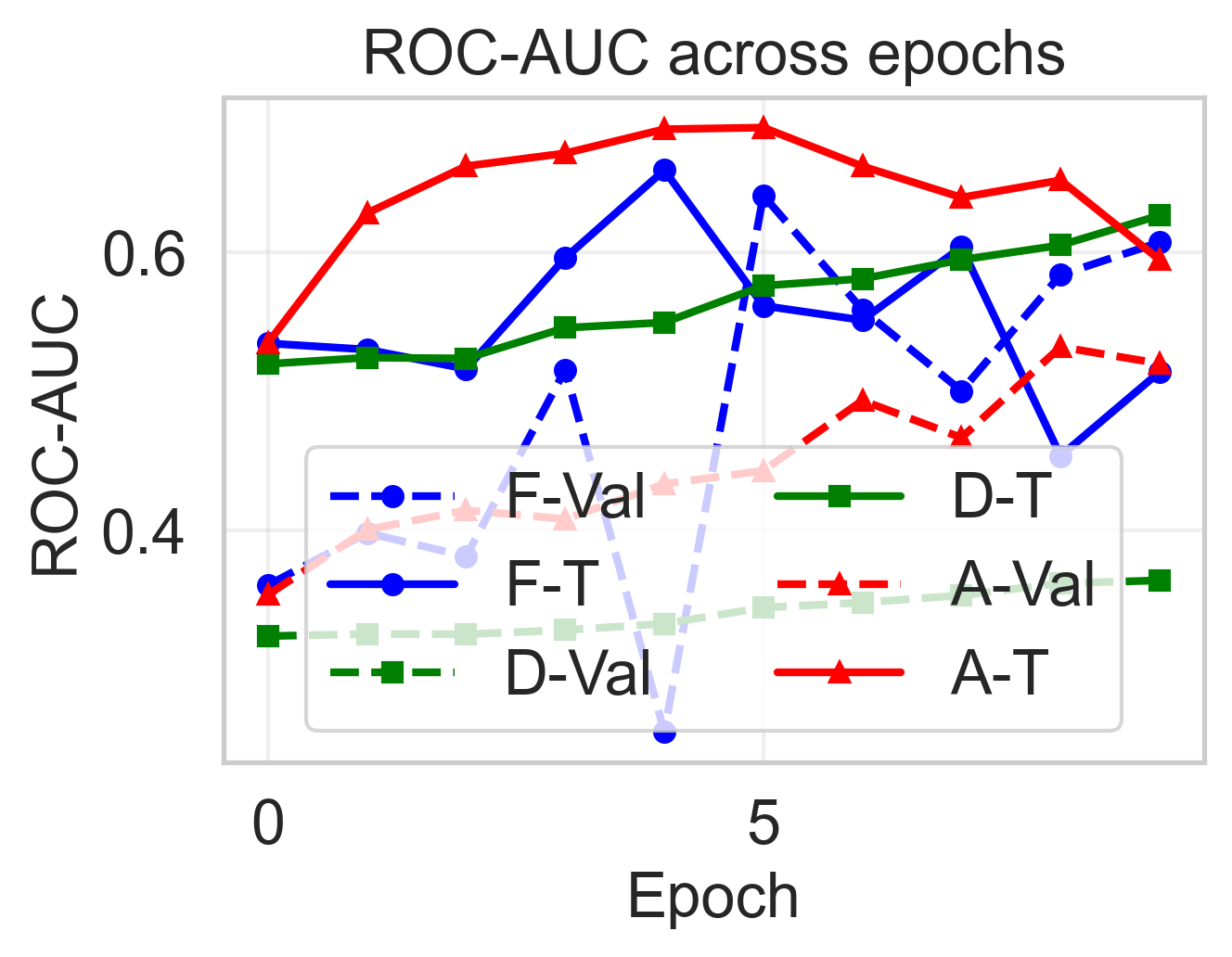}
    \caption{rocauc}
    \label{fig:roc_auc_epochs}
\end{subfigure}\hfill
\begin{subfigure}[t]{0.245\textwidth}
    \centering
    \includegraphics[width=\linewidth]{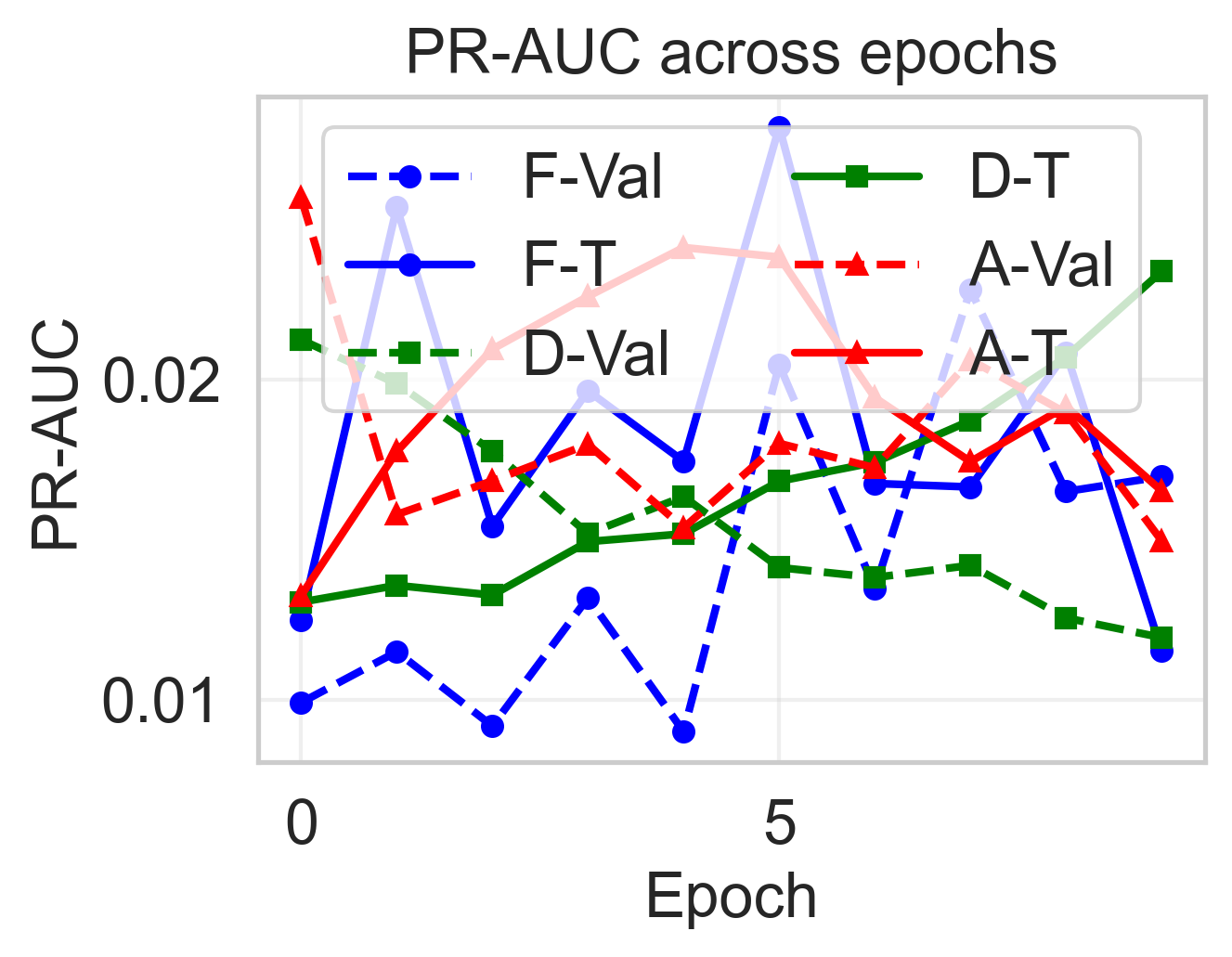}
    \caption{pr auc}
    \label{fig:pr_auc_epochs}
\end{subfigure}\hfill
\begin{subfigure}[t]{0.245\textwidth}
    \centering
    \includegraphics[width=\linewidth]{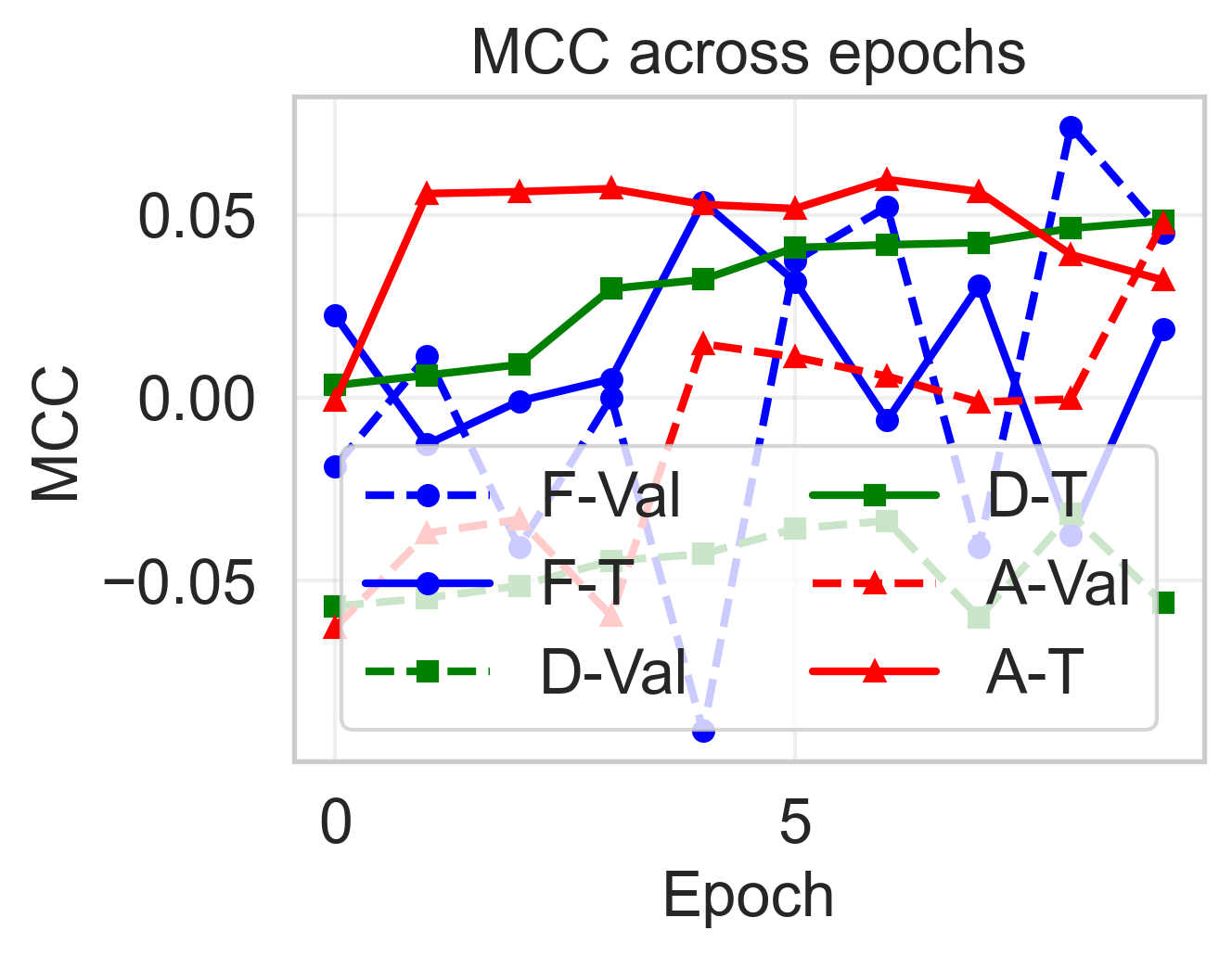}
    \caption{mcc}
    \label{fig:f1_epochs}
\end{subfigure}\hfill
\begin{subfigure}[t]{0.245\textwidth}
    \centering
    \includegraphics[width=\linewidth]{specificity_epochs.png}
    \caption{Client Fairness}
    \label{fig:roc_auc_epochs}
\end{subfigure}

\caption{BAF Data}
\label{fig:metrics_BAF_2}
\end{figure*}

We evaluate DUQFL-Prox against representative QFL baselines and
ablation variants. The analysis is organized around four questions: 
(i) whether DUQFL-Prox improves global and client-level performance, 
(ii) whether the proposed method reduces train--test generalization gap and
client fairness gap, 
(iii) how DUQFL-Prox behaves under severe class imbalance, and 
(iv) whether the learned global QNN checkpoints remain executable on real IBM
quantum hardware.

We report both global and client-level metrics because global accuracy alone is
insufficient to evaluate heterogeneous QFL. A method may obtain strong global
accuracy while still producing unstable or imbalanced performance across
clients. Therefore, in addition to global accuracy, we analyse mean client test
accuracy, train--test gap, client fairness gap, precision, recall, F1-score,
ROC-AUC, PR-AUC, MCC, and specificity.

\subsection{Results on the BAF Dataset}
\label{subsec:baf_results}

The BAF dataset represents a highly imbalanced financial fraud-detection task.
This setting is challenging for QFL because the minority class is rare and the
client partitions are non-IID. Consequently, high global accuracy alone does not
necessarily imply good fraud-detection behaviour. We therefore evaluate BAF using
both federated learning metrics and imbalance-aware classification metrics.

Figure~\ref{fig:baf_global_client_metrics} compares DUQFL-Prox with Default-QFL
and FedProx-QFL using global accuracy, mean client test accuracy, train--test
gap, and client fairness gap. DUQFL-Prox achieves the strongest final global
accuracy, reaching approximately $0.6504$, compared with approximately $0.5344$
for FedProx-QFL and $0.4592$ for Default-QFL. Although Default-QFL reaches a
temporary peak during intermediate rounds, its late-round performance decreases
substantially, indicating unstable post-aggregation behaviour under the non-IID
BAF setting.

The client-level results provide a clearer indication of the benefit of
DUQFL-Prox. The proposed method obtains the highest final mean client test
accuracy, approximately $0.6436$, while FedProx-QFL and Default-QFL obtain
approximately $0.5480$ and $0.5022$, respectively. This suggests that
DUQFL-Prox improves not only the aggregated global model, but also the
generalization behaviour observed across distributed clients.

The train--test gap further supports this conclusion. Default-QFL and
FedProx-QFL show relatively large final train--test gaps, approximately $0.2854$
and $0.2116$, respectively. In contrast, DUQFL-Prox obtains a near-zero final
gap. Since lower train--test gap is preferable, this indicates that DUQFL-Prox
reduces local over-specialization and improves generalization under
heterogeneous client data.

The client fairness gap, measured as the $P90-P10$ spread of client test
accuracies, is also lowest for DUQFL-Prox. The final fairness gap is
approximately $0.0193$ for DUQFL-Prox, compared with $0.0958$ for FedProx-QFL
and $0.1910$ for Default-QFL. This result is important because a high global
accuracy can hide poor performance on difficult or minority clients. The low
fairness gap shows that DUQFL-Prox produces more balanced client-level
performance.

\subsection{Imbalance-Aware Classification Behaviour on BAF}
\label{subsec:baf_imbalance_metrics}

Because the BAF dataset is severely class-imbalanced, we further analyse
precision, recall, F1-score, specificity, ROC-AUC, PR-AUC, and MCC on validation
and test splits. In Figures~\ref{fig:metrics_BAF_1} and
\ref{fig:metrics_BAF_2}, the prefixes F, D, and A denote
FedAvg-default, DUQFL-Prox, and FedAvg-tuned-Adam, respectively, while Val and T
denote validation and test splits.

The results reveal different operating behaviours. FedAvg-tuned-Adam achieves
high recall for much of training, indicating aggressive minority-class
detection. However, this behaviour is accompanied by a decline in specificity,
which suggests a larger number of false positives. Thus, Adam-QFL behaves as a
recall-oriented baseline but provides a less selective operating point.

DUQFL-Prox shows a more balanced late-epoch behaviour. It provides more
consistent improvements in test F1, MCC, ROC-AUC, precision, and specificity.
This suggests that DUQFL-Prox does not simply increase minority-class detection
at the expense of false positives; instead, it maintains a more balanced
recall--specificity trade-off. FedAvg-default exhibits stronger fluctuations
across epochs, particularly for MCC, F1-score, and ROC-AUC, indicating weaker
stability under severe class imbalance.

The validation curves are noisier than the test curves, especially for PR-AUC,
MCC, precision, and recall. This is expected because the validation split
contains very few positive fraud samples. Therefore, epoch-wise trajectories are
important in addition to checkpoint-based summaries, since they reveal the
underlying training dynamics and the recall--specificity trade-off among the
methods.

\subsection{Results on the Genome Dataset}
\label{subsec:genome_results}

\begin{figure*}[t]
\centering

\begin{subfigure}[t]{0.245\textwidth}
    \centering
    \includegraphics[width=\linewidth]{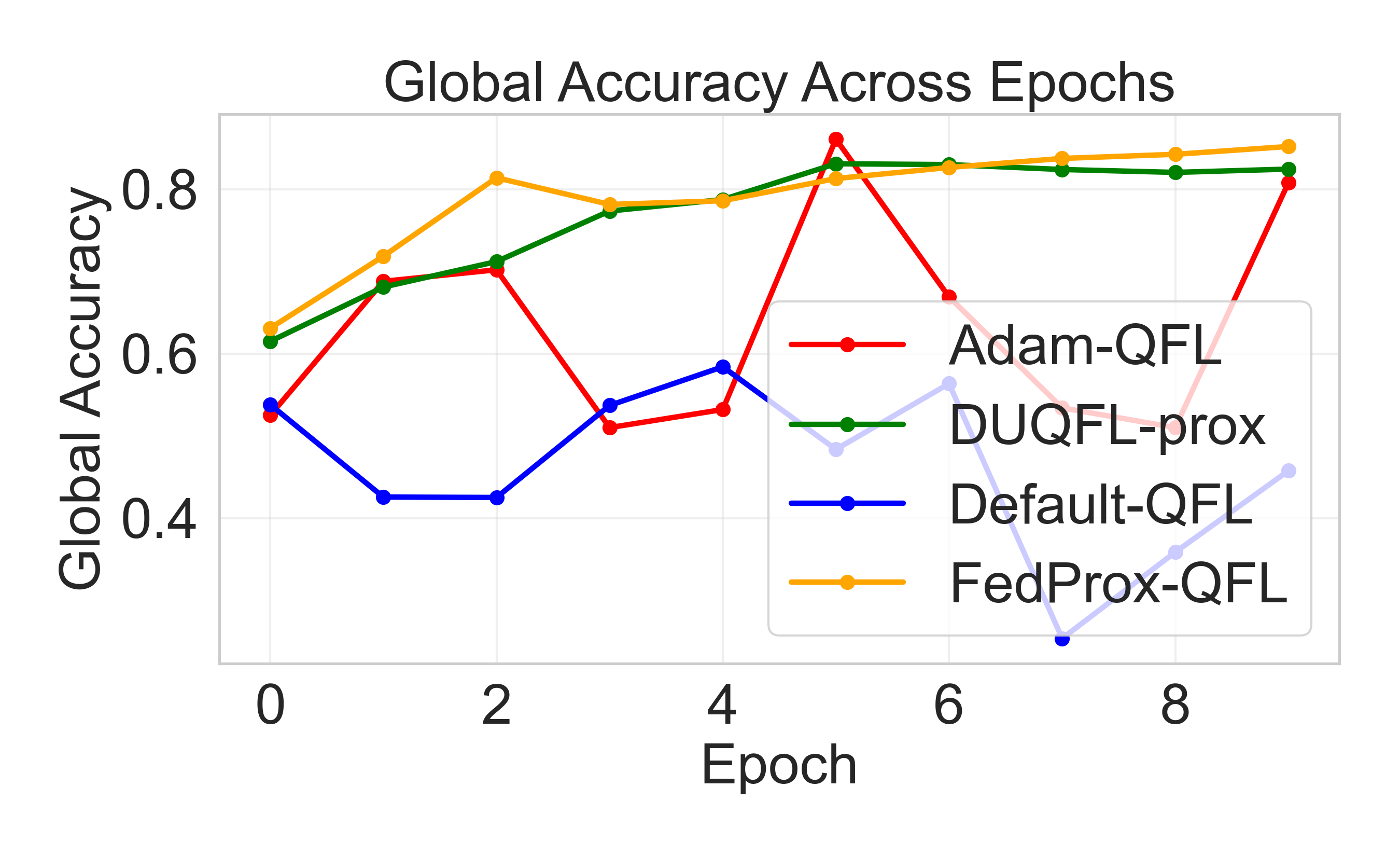}
    \caption{Global accuracy}
    \label{fig:genome_global_accuracy}
\end{subfigure}\hfill
\begin{subfigure}[t]{0.245\textwidth}
    \centering
    \includegraphics[width=\linewidth]{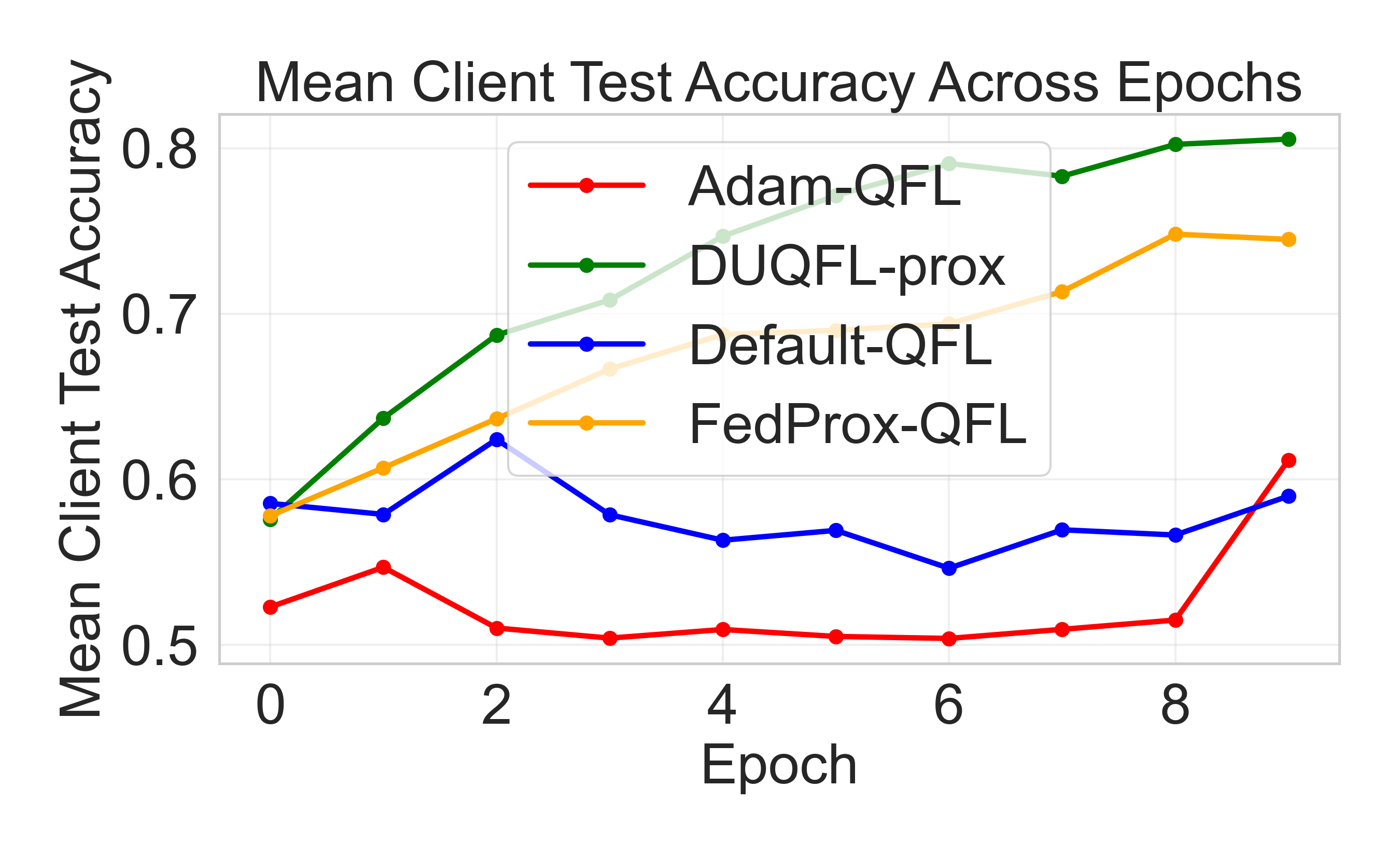}
    \caption{Mean client test accuracy}
    \label{fig:genome_mean_client_test}
\end{subfigure}\hfill
\begin{subfigure}[t]{0.245\textwidth}
    \centering
    \includegraphics[width=\linewidth]{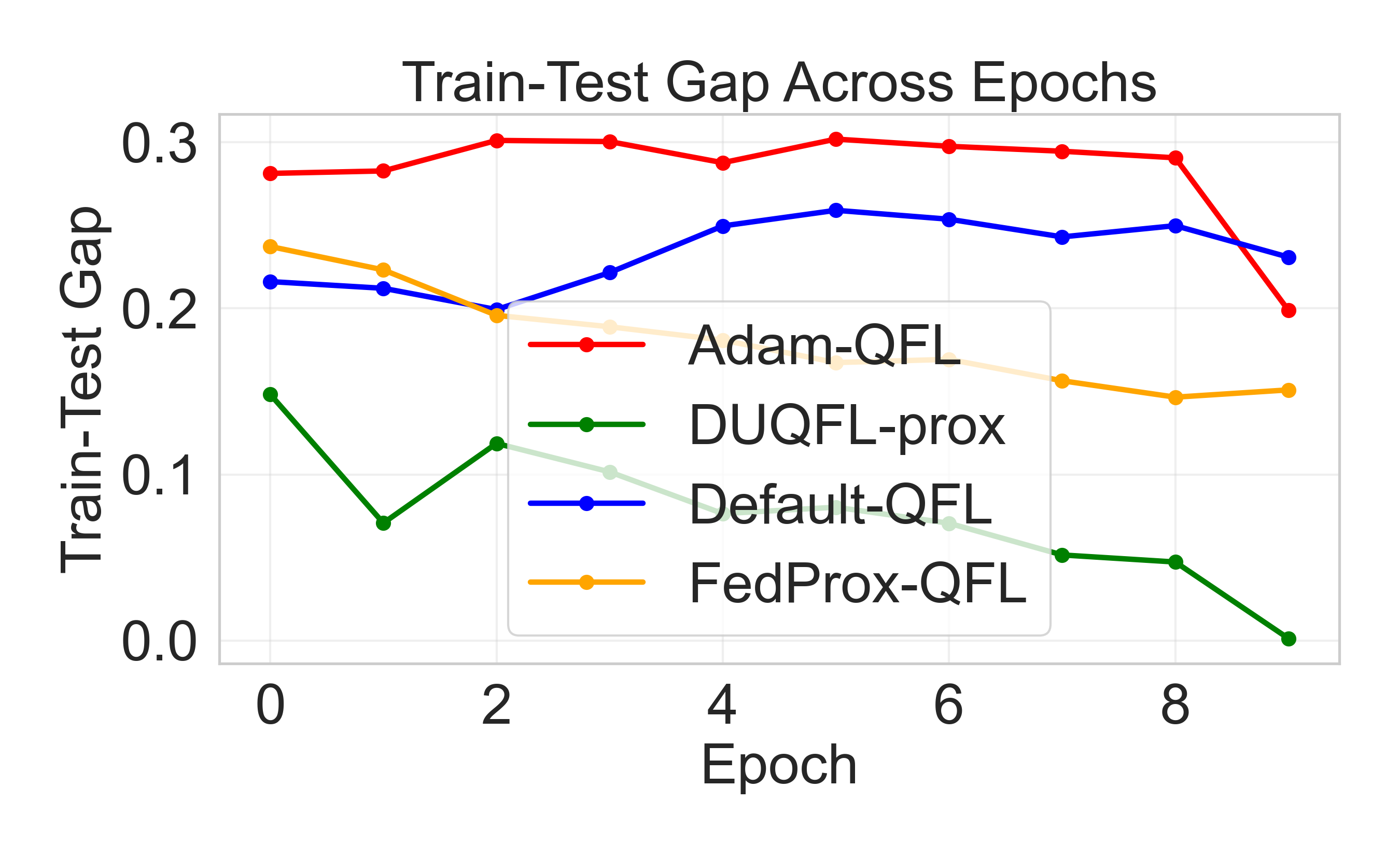}
    \caption{Train--test gap}
    \label{fig:genome_train_test_gap}
\end{subfigure}\hfill
\begin{subfigure}[t]{0.245\textwidth}
    \centering
    \includegraphics[width=\linewidth]{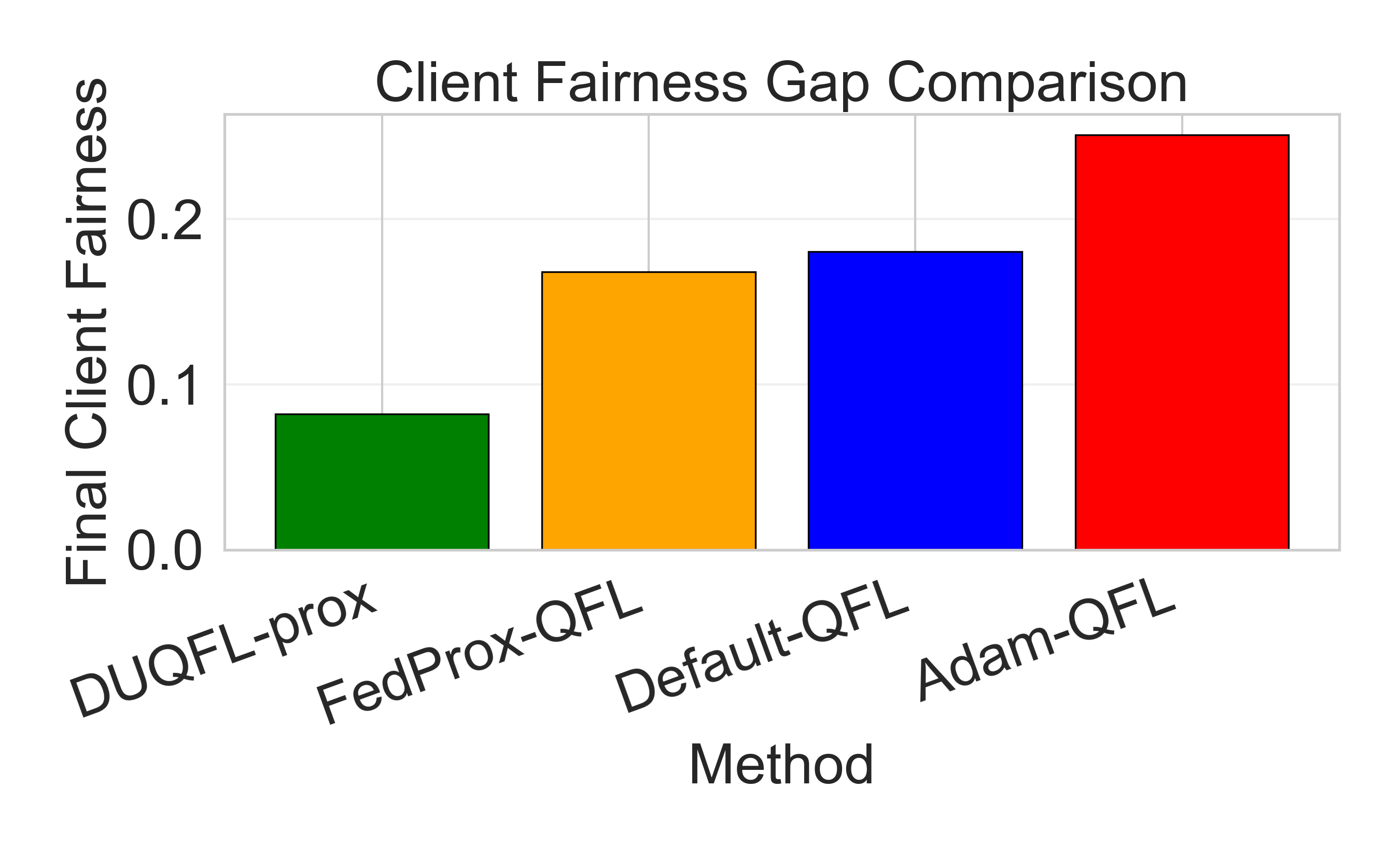}
    \caption{Client fairness gap}
    \label{fig:genome_client_fairness_gap}
\end{subfigure}

\caption{Federated performance comparison on the Genome dataset. FedProx-QFL
achieves the strongest final global accuracy, while DUQFL-Prox provides the best
client-level generalization, lowest train--test gap, and lowest fairness gap.}
\label{fig:genome_global_client_metrics}
\end{figure*}

The Genome dataset presents a more nuanced comparison. As shown in
Figure~\ref{fig:genome_global_client_metrics}, FedProx-QFL achieves the highest
final global accuracy, reaching approximately $0.85$. DUQFL-Prox remains
competitive, with final global accuracy around $0.82$--$0.83$, while
Default-QFL ends with substantially lower global accuracy. This result shows
that DUQFL-Prox is not always the single best method in terms of final global
accuracy.

However, the client-level metrics show a different and important trend.
DUQFL-Prox achieves the highest final mean client test accuracy, approximately
$0.80$, outperforming the other compared methods. This indicates that
DUQFL-Prox provides stronger generalization across distributed clients, even
when another baseline obtains slightly higher final global accuracy.

The train--test gap and client fairness gap further support this interpretation.
DUQFL-Prox obtains the lowest final train--test gap, close to zero, suggesting
reduced overfitting and improved client-level generalization. It also achieves
the lowest final client fairness gap, indicating that its performance is more
balanced across clients. DUQFL-best and DUQFL-drift also improve over
Default-QFL, but DUQFL-Prox provides the strongest overall stability and
fairness profile.

Therefore, the Genome experiment supports a more federated-learning-relevant
conclusion: DUQFL-Prox is not merely an accuracy-maximizing method, but a
stability-aware and generalization-aware QFL method. It provides the best
trade-off among global accuracy, mean client test accuracy, train--test gap, and
fairness under heterogeneous client data.

\begin{table*}[htbp]
\centering
\scriptsize
\caption{Qualitative comparison of QFL and DUQFL methods across Genome and BAF datasets. 
Here, $\checkmark\checkmark$ denotes the strongest performance, $\checkmark$ denotes competitive or moderate performance, and $-$ denotes weak, unstable, or less consistent performance. For train--test gap, client drift/regularization, and fairness/balance, lower values are preferable.}
\label{tab:qualitative_genome_baf}
\resizebox{\textwidth}{!}{%
\begin{tabular}{llcccccc}
\toprule
\textbf{Dataset} & \textbf{Method} 
& \textbf{Final global accuracy} 
& \textbf{Peak global accuracy} 
& \textbf{Mean client test accuracy} 
& \textbf{Low train--test gap} 
& \textbf{Low client drift / regularization} 
& \textbf{Client fairness / balance} \\
\midrule

\multirow{4}{*}{\textbf{Genome}}
& Adam-QFL 
& $-$ 
& $\checkmark\checkmark$ 
& $-$ 
& $-$ 
& $-$ 
& $-$ \\

& DUQFL-Prox 
& $\checkmark$ 
& $-$ 
& $\checkmark\checkmark$ 
& $\checkmark\checkmark$ 
& $\checkmark\checkmark$ 
& $\checkmark\checkmark$ \\

& Default-QFL 
& $-$ 
& $-$ 
& $-$ 
& $-$ 
& $-$ 
& $-$ \\

& FedProx-QFL 
& $\checkmark\checkmark$ 
& $\checkmark$ 
& $\checkmark$ 
& $\checkmark$ 
& $\checkmark$ 
& $\checkmark$ \\

\midrule

\multirow{4}{*}{\textbf{BAF}}
& Adam-QFL 
& $-$ 
& $-$ 
& $-$ 
& $-$ 
& $-$ 
& $-$ \\

& DUQFL-Prox 
& $\checkmark\checkmark$ 
& $\checkmark\checkmark$ 
& $\checkmark\checkmark$ 
& $\checkmark\checkmark$ 
& $\checkmark\checkmark$ 
& $\checkmark\checkmark$ \\

& Default-QFL 
& $-$ 
& $\checkmark$ 
& $-$ 
& $-$ 
& $-$ 
& $-$ \\

& FedProx-QFL 
& $\checkmark$ 
& $-$ 
& $\checkmark$ 
& $\checkmark$ 
& $\checkmark$ 
& $\checkmark$ \\

\bottomrule
\end{tabular}%
}
\end{table*}

\subsection{Cross-Dataset Comparison}
\label{subsec:cross_dataset_comparison}

Table~\ref{tab:qualitative_genome_baf} summarizes the qualitative behaviour of
the compared methods across the Genome and BAF datasets. The results show that
DUQFL-Prox provides the most consistent stability--generalization trade-off
across datasets.

On BAF, DUQFL-Prox achieves the strongest performance across final global
accuracy, mean client test accuracy, train--test gap, and client fairness gap.
This indicates that proximal deep-unfolded QNN optimization is particularly
effective under severe class imbalance and non-IID client partitions.

On Genome, FedProx-QFL achieves the highest final global accuracy, whereas
DUQFL-Prox achieves the strongest client-level generalization, lowest
train--test gap, and lowest fairness gap. This distinction is important because
a single global test metric does not fully characterize federated performance.
The Genome results show that DUQFL-Prox improves the reliability and balance of
client-level performance even when another method slightly improves the final
centralized global accuracy.

Overall, the cross-dataset results indicate that DUQFL-Prox should be interpreted
as a stability-aware and generalization-aware QFL method. Its advantage lies in
improving the quality and aggregation compatibility of local QNN updates through
deep-unfolded optimization, validation-based checkpoint selection, and proximal
drift control.

\begin{figure}[t]
\centering

\begin{subfigure}[t]{0.48\linewidth}
    \centering
    \includegraphics[width=\linewidth]{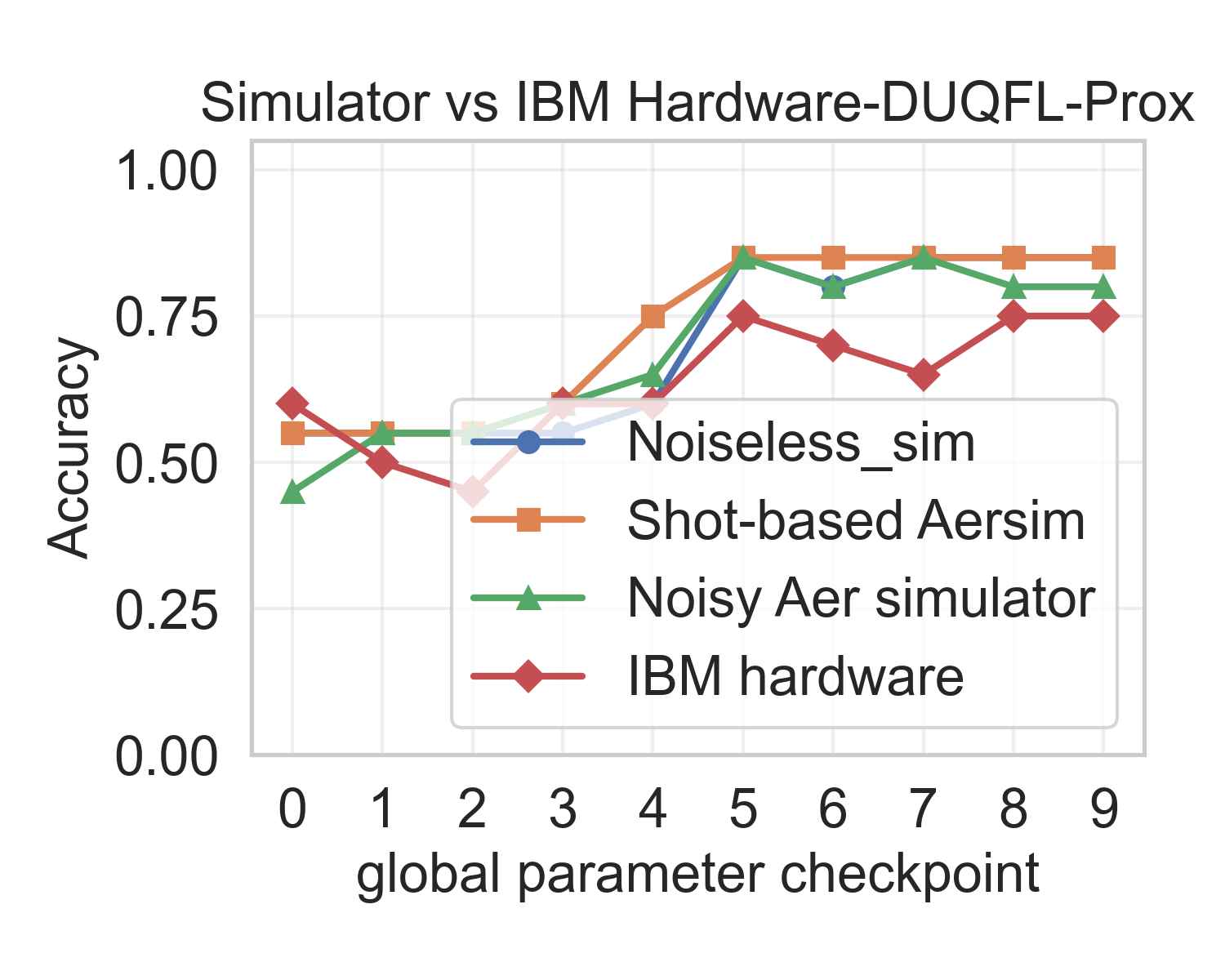}
    \caption{DUQFL-Prox}
    \label{fig:ibm_duqfl_prox}
\end{subfigure}\hfill
\begin{subfigure}[t]{0.48\linewidth}
    \centering
    \includegraphics[width=\linewidth]{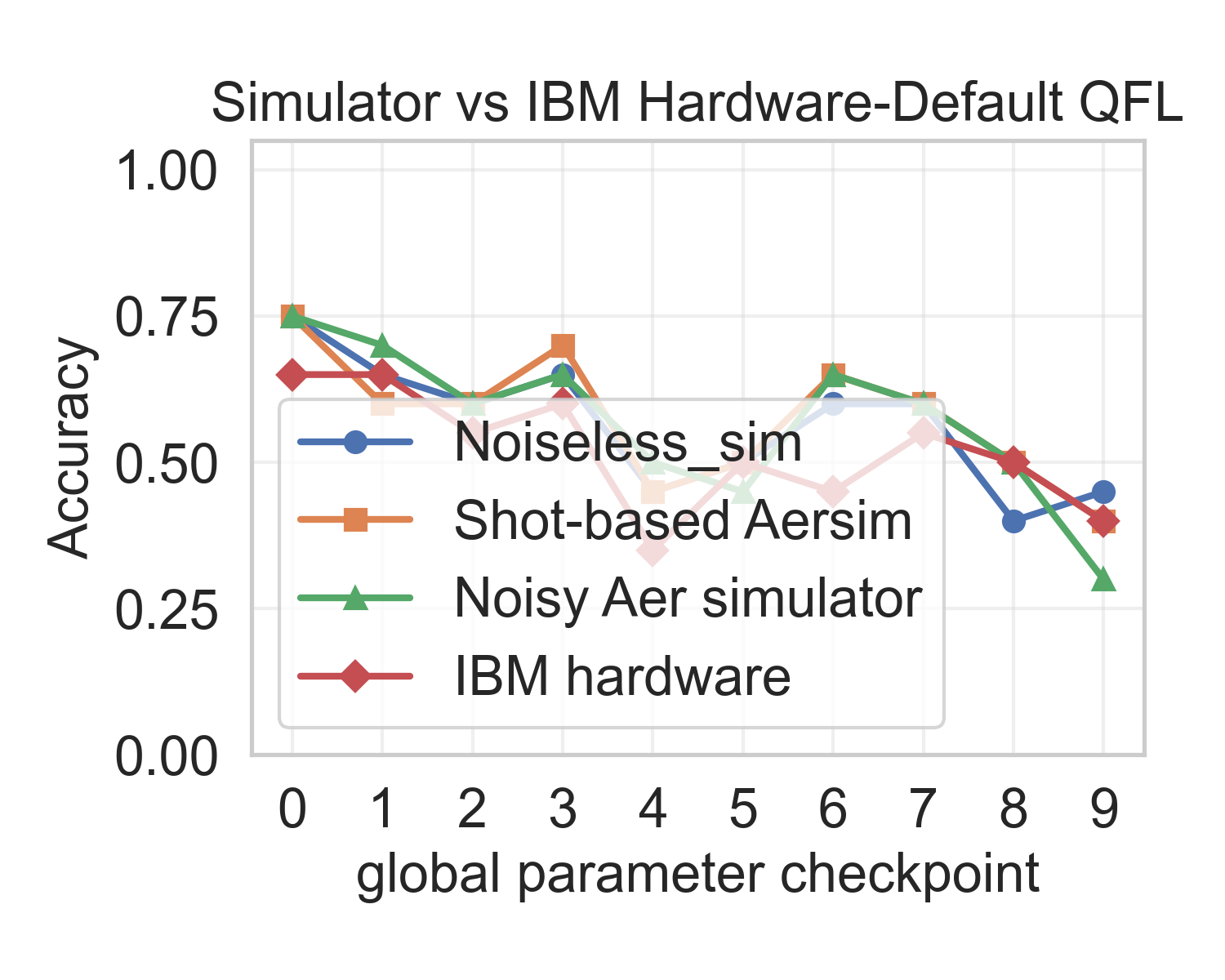}
    \caption{Default-QFL}
    \label{fig:ibm_default_qfl}
\end{subfigure}

\caption{Simulator and real IBM quantum hardware validation of trained global
QFL checkpoints. DUQFL-Prox shows a more stable hardware trajectory than
Default-QFL across saved global parameter checkpoints.}
\label{fig:real_ibm_validation}
\end{figure}

\subsection{Controller Behaviour and Hyperparameter Adaptation}

\begin{figure}
    \centering
    \includegraphics[width=1\linewidth]{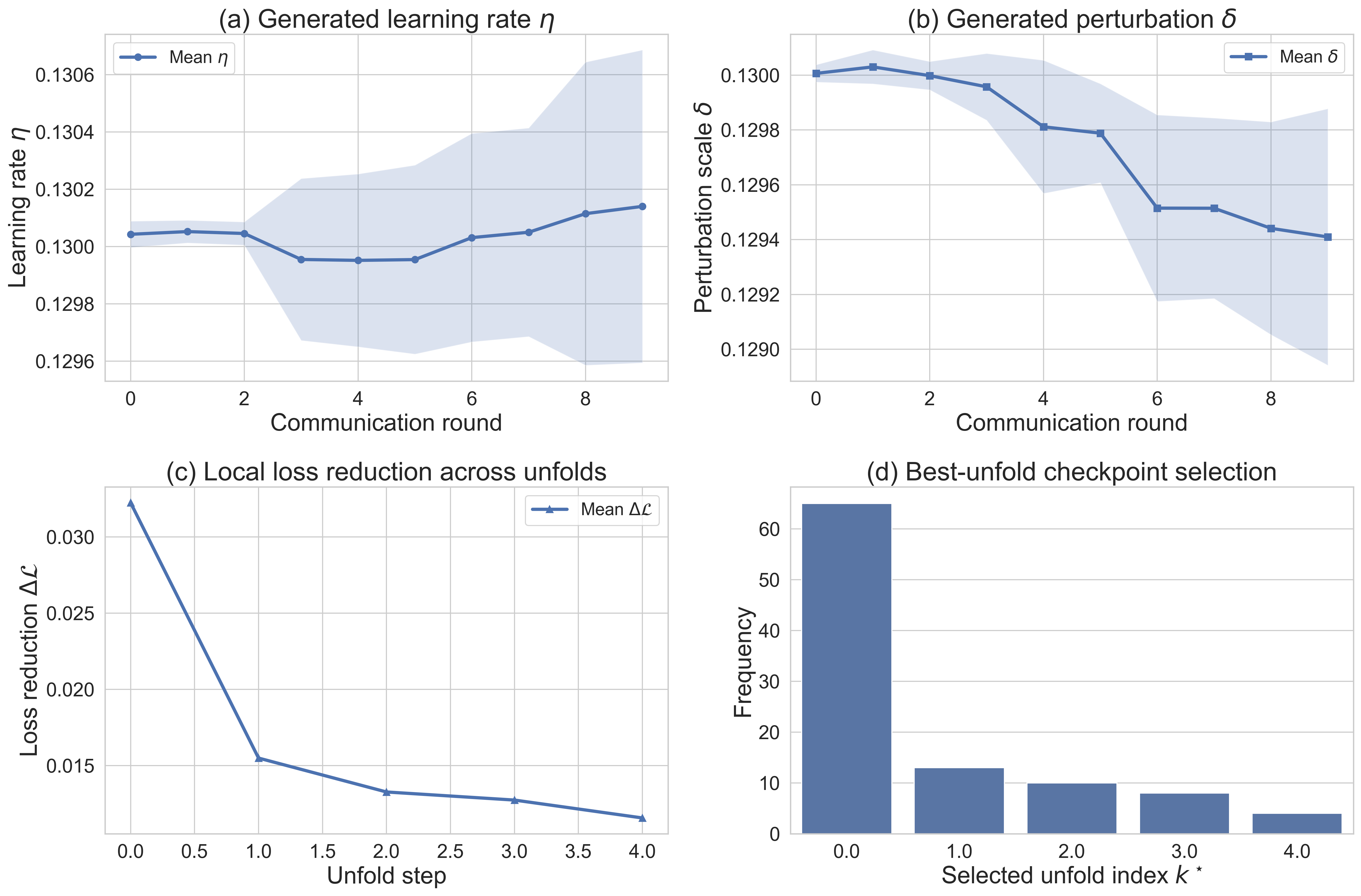}
    \caption{Controller behaviour and hyperparameter adaptation in DUQFL-Prox on the
BAF dataset. The trace logs show the controller-generated SPSA learning rate
$\eta$, perturbation scale $\delta$, local loss reduction
$\Delta\mathcal{L}$, and selected best-unfold checkpoint $k^\star$. These
results verify that DUQFL-Prox performs adaptive local QNN optimization rather
than using a fixed SPSA schedule.}
    \label{fig:controller_behaviour}
\end{figure}
Figure~\ref{fig:controller_behaviour} analyses the controller trace logs of
DUQFL-Prox. The generated learning rate $\eta$ and perturbation scale $\delta$
remain within stable numerical ranges while varying across the unfolded training
process. The local loss-reduction curve shows that the unfolded SPSA blocks
produce measurable improvement in the client objective, particularly in the
early unfold steps. The selected-checkpoint distribution further shows that the
last unfolded state is not always uploaded, supporting the use of
validation-based best-unfold selection. Together, these results provide empirical
evidence that DUQFL-Prox uses controller-guided adaptive optimization rather
than a fixed local SPSA schedule.

\subsection{Real IBM Quantum Hardware Validation}
\label{subsec:ibm_hardware_validation}

To examine practical deployability, we evaluated selected trained global QNN
checkpoints on IBM Quantum hardware. Due to the high queue time and execution
cost of repeatedly training the full federated process on real quantum devices,
hardware execution was used for post-training validation rather than full
hardware-in-the-loop federated training. Specifically, saved global parameter
checkpoints were assigned to the trained QNN circuit, transpiled for the selected
IBM backend, and executed using finite-shot measurement. The resulting hardware
accuracy was compared with noiseless simulator, shot-based Aer simulator, and
backend-inspired noisy Aer simulator results.

Figure~\ref{fig:real_ibm_validation} compares DUQFL-Prox and Default-QFL across
saved global checkpoints. DUQFL-Prox shows a more stable and improving hardware
trajectory than Default-QFL and remains reasonably aligned with the simulator
curves in later checkpoints. The difference between simulator and hardware
results is expected due to finite-shot uncertainty, device noise, calibration
drift, transpilation constraints, and backend-specific routing overhead.

These results should not be interpreted as full federated training on IBM
hardware. Rather, they provide post-training hardware feasibility evidence,
showing that the learned DUQFL-Prox global checkpoints remain executable on real
quantum hardware and retain a more stable trajectory than the Default-QFL
baseline.




\section{Conclusion} \label{sec:conclusion}

The proposed DUQFL-Prox framework is supported by three theoretical observations. First, the optional step-projection mechanism bounds each unfolded local update and therefore bounds cumulative client drift across the unfolded trajectory. Second, the proximal regularization term acts as a drift-control mechanism by penalizing deviation from the broadcast global model, balancing local adaptation with aggregation compatibility. Third, under standard smoothness assumptions on the post-aggregation meta-objective, the outer SPSA update provides a zeroth-order controller-update direction that supports descent-oriented adaptation in expectation. Our results support the use of DUQFL-Prox as a drift-stable QFL`  framework for intelligent services that require privacy preservation, fairness, and reliable generalization across heterogeneous clients. These results do not constitute a full convergence guaranty for arbitrary nonconvex QFL systems; instead, they characterize the stability, drift-control, and meta-optimization behavior induced by DUQFL-Prox. Detailed statements and proofs are provided in Appendix~A.
\bibliographystyle{IEEEtran}
\bibliography{refs}

.\newpage

\appendices

Table~\ref{tab:notation_summary} summarizes the main notation used throughout the paper.

\begin{table*}[htbp]
\centering
\caption{Summary of notation used in DUQFL-Prox.}
\label{tab:notation_summary}
\small
\begin{tabular}{ll}
\hline
\textbf{Notation} & \textbf{Description} \\
\hline
$\mathcal{C}=\{1,\ldots,N\}$ & Set of all clients \\
$N$ & Number of clients \\
$t \in \{0,\ldots,T-1\}$ & Communication round index \\
$T$ & Number of communication rounds \\
$\mathcal{S}^{(t)} \subseteq \mathcal{C}$ & Participating client set at round $t$ \\
$\mathcal{D}_i$ & Local training dataset of client $i$ \\
$n_i = |\mathcal{D}_i|$ & Number of local training samples at client $i$ \\
$\mathcal{V}_i$ & Local validation set of client $i$ \\
$\mathcal{D}_{\mathrm{val}}$ & Global or held-out validation data used for meta-objective evaluation \\
$\boldsymbol{\theta}^{(t)}$ & Global QNN parameter vector at round $t$ \\
$\boldsymbol{\theta}_{i,k}^{(t)}$ & Local QNN parameter vector of client $i$ at unfold step $k$ of round $t$ \\
$\boldsymbol{\theta}_{i,\star}^{(t)}$ & Selected best unfolded local model of client $i$ at round $t$ \\
$P$ & Number of trainable QNN parameters \\
$K$ & Number of unfolded local optimization steps \\
$k$ & Unfold-step index \\
$f(\mathbf{x};\boldsymbol{\theta})$ & QNN prediction function \\
$\mathcal{L}_i(\boldsymbol{\theta})$ & Empirical local loss of client $i$ \\
$\mathcal{L}_i^{\mathrm{prox}}(\boldsymbol{\theta})$ & Proximal local objective of client $i$ \\
$\eta_{i,k}^{(t)}$ & SPSA learning rate generated for client $i$ at unfold step $k$ \\
$\delta_{i,k}^{(t)}$ & SPSA perturbation scale generated for client $i$ at unfold step $k$ \\
$\mathbf{z}_{i,k}^{(t)}$ & Optimization-state feature vector used by the controller \\
$\boldsymbol{\phi}^{(t)}$ & Shared controller parameter vector at round $t$ \\
$\boldsymbol{\phi}_{\eta}^{(t)}$ & Controller parameters used to generate $\eta_{i,k}^{(t)}$ \\
$\boldsymbol{\phi}_{\delta}^{(t)}$ & Controller parameters used to generate $\delta_{i,k}^{(t)}$ \\
$\mu$ & Proximal regularization coefficient \\
$w_i^{(t)}$ & Aggregation weight of client $i$ at round $t$ \\
$\mathcal{L}_{\mathrm{meta}}^{(t)}$ & Outer meta-loss at round $t$ \\
$\alpha_{\mathrm{out}}$ & Outer controller learning rate \\
$c_{\mathrm{out}}$ & Outer SPSA perturbation radius \\
$\boldsymbol{\Delta}_{\phi}$ & Rademacher perturbation vector for outer SPSA \\
\hline
\end{tabular}
\end{table*}

\section{Proximal Objective and Drift Control}
\label{app:proximal_objective}

This appendix clarifies the drift-control role of the proximal component in
DUQFL-Prox. In heterogeneous QFL, each client initializes local training from the
broadcast global model $\boldsymbol{\theta}^{(t)}$. Under non-IID data and
stochastic QNN optimization, repeated local updates may move the client model far
from this global reference point, producing updates that are less compatible
with server aggregation.

DUQFL-Prox therefore replaces the unregularized local objective
$\mathcal{L}_i(\boldsymbol{\theta})$ with
\begin{equation}
\mathcal{L}_i^{\mathrm{prox}}(\boldsymbol{\theta})
=
\mathcal{L}_i(\boldsymbol{\theta})
+
\frac{\mu}{2}
\left\|
\boldsymbol{\theta}
-
\boldsymbol{\theta}^{(t)}
\right\|_2^2,
\label{eq:app_prox_objective}
\end{equation}
where $\mu \geq 0$ controls the strength of proximal regularization. The first
term promotes local empirical improvement, whereas the second term penalizes
excessive deviation from the broadcast global model.

Formally, the gradient of the proximal objective is
\begin{equation}
\nabla \mathcal{L}_i^{\mathrm{prox}}(\boldsymbol{\theta})
=
\nabla \mathcal{L}_i(\boldsymbol{\theta})
+
\mu
\left(
\boldsymbol{\theta}
-
\boldsymbol{\theta}^{(t)}
\right).
\label{eq:app_prox_gradient}
\end{equation}
Although DUQFL-Prox uses SPSA rather than analytic gradients, this expression
shows that the proximal term introduces a drift-correcting component toward the
current global model. Thus, local training is discouraged from moving
arbitrarily far from $\boldsymbol{\theta}^{(t)}$.

At unfold step $k$, SPSA is applied to the proximal objective:
\begin{equation}
\boldsymbol{\theta}_{i,k+1}^{(t)}
=
\mathcal{U}_{\mathrm{SPSA}}
\left(
\boldsymbol{\theta}_{i,k}^{(t)};
\eta_{i,k}^{(t)},
\delta_{i,k}^{(t)},
\mathcal{L}_i^{\mathrm{prox}}
\right).
\label{eq:app_prox_spsa}
\end{equation}
Therefore, the local update is influenced by both client-specific loss reduction
and proximity to the broadcast global parameters.

The unfold-step displacement is recorded as
\begin{equation}
\Delta \boldsymbol{\theta}_{i,k}^{(t)}
=
\boldsymbol{\theta}_{i,k+1}^{(t)}
-
\boldsymbol{\theta}_{i,k}^{(t)},
\label{eq:app_displacement}
\end{equation}
with magnitude
\begin{equation}
\left\|
\Delta \boldsymbol{\theta}_{i,k}^{(t)}
\right\|_2 .
\end{equation}
This quantity provides a diagnostic measure of local movement and can be used as
an optimization-state feature in subsequent unfold steps.

Overall, the proximal term does not prevent local learning. Instead, it balances
local improvement with aggregation compatibility. This is particularly important
in DUQFL-Prox because deep unfolding increases the flexibility of local
optimization, which can otherwise increase the risk of client over-specialization
under heterogeneous data.
\section{Outer SPSA Gradient Estimation}
\label{app:outer_spsa}

The shared controller in DUQFL-Prox is parameterized by
$\boldsymbol{\phi}^{(t)}$ and generates the SPSA learning rate and perturbation
scale used during unfolded local QNN optimization. The purpose of the outer
update is to refine $\boldsymbol{\phi}^{(t)}$ so that the induced local
optimization trajectories improve post-aggregation global performance.

At communication round $t$, the outer meta-loss is
\begin{equation}
\mathcal{L}_{\mathrm{meta}}^{(t)}
=
\mathcal{L}_{\mathrm{val}}^{\mathrm{global}}
\left(
\boldsymbol{\theta}^{(t+1)}
\right)
+
\lambda_{\mathrm{fair}}\Omega_{\mathrm{fair}}^{(t)}
+
\lambda_{\mathrm{comm}}\Omega_{\mathrm{comm}}^{(t)}
+
\lambda_{\mathrm{stab}}\Omega_{\mathrm{stab}}^{(t)}.
\label{eq:app_meta_loss}
\end{equation}
The dependence of $\mathcal{L}_{\mathrm{meta}}^{(t)}$ on
$\boldsymbol{\phi}^{(t)}$ is indirect: the controller determines
$(\eta,\delta)$, which affects local QNN training, server aggregation, and
finally the global validation loss. Since this mapping is expensive and not
easily differentiable, DUQFL-Prox applies an outer SPSA update.

A Rademacher perturbation vector
$\boldsymbol{\Delta}_{\phi}\in\{-1,+1\}^{\dim(\boldsymbol{\phi})}$ is sampled,
and two perturbed controllers are formed:
\begin{equation}
\boldsymbol{\phi}^{+}
=
\boldsymbol{\phi}^{(t)}
+
c_{\mathrm{out}}\boldsymbol{\Delta}_{\phi},
\qquad
\boldsymbol{\phi}^{-}
=
\boldsymbol{\phi}^{(t)}
-
c_{\mathrm{out}}\boldsymbol{\Delta}_{\phi}.
\label{eq:app_outer_perturb}
\end{equation}
Evaluating the federated process under these two controllers gives
$\mathcal{L}_{\mathrm{meta}}^{+}$ and
$\mathcal{L}_{\mathrm{meta}}^{-}$. For the $j$-th controller parameter, the
SPSA finite-difference estimate is
\begin{equation}
\widehat{g}_{\phi,j}^{(t)}
=
\frac{
\mathcal{L}_{\mathrm{meta}}^{+}
-
\mathcal{L}_{\mathrm{meta}}^{-}
}
{
2c_{\mathrm{out}}\Delta_{\phi,j}
}.
\label{eq:app_outer_spsa_component}
\end{equation}
Equivalently, in vector form,
\begin{equation}
\widehat{\mathbf{g}}_{\phi}^{(t)}
=
\frac{
\mathcal{L}_{\mathrm{meta}}^{+}
-
\mathcal{L}_{\mathrm{meta}}^{-}
}
{2c_{\mathrm{out}}}
\boldsymbol{\Delta}_{\phi}^{-1}.
\label{eq:app_outer_spsa_vector}
\end{equation}
Since $\Delta_{\phi,j}\in\{-1,+1\}$, we have
$\Delta_{\phi,j}^{-1}=\Delta_{\phi,j}$.

The controller is then updated as
\begin{equation}
\boldsymbol{\phi}^{(t+1)}
=
\boldsymbol{\phi}^{(t)}
-
\alpha_{\mathrm{out}}
\widehat{\mathbf{g}}_{\phi}^{(t)},
\label{eq:app_controller_update}
\end{equation}
where $\alpha_{\mathrm{out}}$ is the outer learning rate. This update does not
directly overwrite client-local QNN models. Instead, it modifies the controller
mapping
\begin{equation}
\Gamma
\left(
\mathbf{z}_{i,k}^{(t)};
\boldsymbol{\phi}^{(t)}
\right)
\mapsto
\left(
\eta_{i,k}^{(t)},
\delta_{i,k}^{(t)}
\right),
\end{equation}
so that future local SPSA behaviour becomes more aligned with the
post-aggregation global objective.

\section{Stability and Convergence Discussion}
\label{app:stability_convergence}

This appendix provides a stability-oriented interpretation of DUQFL-Prox. We do
not claim a general convergence theorem for arbitrary non-convex QNN objectives,
since variational quantum loss landscapes may be non-convex and affected by
finite-shot measurement noise. Instead, we clarify how the proposed components
are designed to improve optimization stability under heterogeneous federated
data.

Let the global objective be written as
\begin{equation}
\mathcal{F}(\boldsymbol{\theta})
=
\sum_{i=1}^{N}
p_i
\mathcal{L}_i(\boldsymbol{\theta}),
\qquad
p_i
=
\frac{n_i}{\sum_{j=1}^{N} n_j}.
\label{eq:app_global_objective}
\end{equation}
Here, $\mathcal{L}_i(\boldsymbol{\theta})$ denotes the local objective of client
$i$, and $p_i$ is its sample-size proportion. Under non-IID data, the local
objectives may induce different descent directions. Therefore, a client model
that improves its own local objective may still be poorly aligned with the
global objective after aggregation. This mismatch is the source of client drift.

DUQFL-Prox addresses this instability through three complementary mechanisms.
First, the proximal penalty
\begin{equation}
\frac{\mu}{2}
\left\|
\boldsymbol{\theta}
-
\boldsymbol{\theta}^{(t)}
\right\|_2^2
\end{equation}
discourages excessive deviation from the broadcast global model
$\boldsymbol{\theta}^{(t)}$. This encourages local updates to remain within a
controlled neighbourhood of the global reference point, reducing the risk of
over-specialized client updates.

Second, validation-based best-unfold selection prevents the method from always
uploading the final unfolded state. Instead, client $i$ uploads
\begin{equation}
\boldsymbol{\theta}_{i,\star}^{(t)}
=
\boldsymbol{\theta}_{i,k_i^\star}^{(t)},
\qquad
k_i^\star
=
\arg\min_{k\in\{1,\ldots,K\}}
\mathcal{L}_{i,\mathrm{val}}
\left(
\boldsymbol{\theta}_{i,k}^{(t)}
\right).
\end{equation}
This mechanism is important because later unfold steps may reduce training loss
while increasing validation loss or local specialization.

Third, the outer meta-objective evaluates the consequence of local optimization
after aggregation:
\begin{equation}
\mathcal{L}_{\mathrm{meta}}^{(t)}
=
\mathcal{L}_{\mathrm{val}}^{\mathrm{global}}
\left(
\boldsymbol{\theta}^{(t+1)}
\right)
+
\lambda_{\mathrm{fair}}\Omega_{\mathrm{fair}}^{(t)}
+
\lambda_{\mathrm{comm}}\Omega_{\mathrm{comm}}^{(t)}
+
\lambda_{\mathrm{stab}}\Omega_{\mathrm{stab}}^{(t)}.
\end{equation}
Thus, the controller is guided by post-aggregation global behaviour rather than
local training loss alone.

The proximal coefficient $\mu$ controls the trade-off between local adaptation
and aggregation compatibility. A larger $\mu$ more strongly penalizes movement
away from $\boldsymbol{\theta}^{(t)}$, whereas a smaller $\mu$ gives clients
greater freedom to adapt to their local data:
\begin{equation}
\text{local adaptivity}
\quad
\leftrightarrow
\quad
\text{global aggregation compatibility}.
\end{equation}

Therefore, DUQFL-Prox should be interpreted as a stability-aware optimization
framework rather than a method that guarantees global convergence for all QNN
loss landscapes. Its purpose is to reduce harmful client drift, select more
generalizable local unfolded checkpoints, and align the controller with
post-aggregation validation performance. This interpretation is consistent with
our experimental protocol, which reports global accuracy together with mean
client test accuracy, train--test gap, fairness gap, and classification metrics.

\section{Relation to Classical Deep-Unfolded Federated Learning}
\label{app:relation_classical_dufl}

Classical deep-unfolded federated learning and DUQFL-Prox share the same
high-level principle: an iterative federated optimization process can be
unfolded into structured steps, and selected components of this process can be
made learnable. However, the level at which learning is introduced is different.

In classical deep-unfolded weighted averaging for FL, the unfolded process is
typically applied at the server aggregation level \cite{nakai2024deep}. Standard FedAvg computes
\begin{equation}
\mathbf{w}^{(t+1)}
=
\sum_{i=1}^{N}
\frac{n_i}{\sum_{j=1}^{N} n_j}
\mathbf{w}_i^{(t)},
\end{equation}
where $\mathbf{w}_i^{(t)}$ denotes the local model from client $i$ and $n_i$ is
the corresponding local sample size. Deep-unfolded weighted averaging replaces
these fixed aggregation weights with learnable coefficients:
\begin{equation}
\mathbf{w}^{(t+1)}
=
\sum_{i=1}^{N}
\Theta_i^{(t)}
\mathbf{w}_i^{(t)},
\label{eq:app_classical_duw}
\end{equation}
where $\Theta_i^{(t)}$ determines the contribution of client $i$ to the global
model at round $t$. Thus, classical deep-unfolded FL primarily asks how client
updates should be weighted during aggregation.

DUQFL-Prox addresses a different question: how each client should optimize its
local QNN before aggregation. Instead of learning aggregation weights
$\Theta_i^{(t)}$, DUQFL-Prox learns a controller parameterized by
$\boldsymbol{\phi}^{(t)}$. This controller generates unfold-specific SPSA
hyperparameters:
\begin{equation}
\left(
\eta_{i,k}^{(t)},
\delta_{i,k}^{(t)}
\right)
=
\Gamma
\left(
\mathbf{z}_{i,k}^{(t)};
\boldsymbol{\phi}^{(t)}
\right).
\end{equation}
The server aggregation remains sample-size-weighted:
\begin{equation}
\boldsymbol{\theta}^{(t+1)}
=
\sum_{i\in\mathcal{S}^{(t)}}
w_i^{(t)}
\boldsymbol{\theta}_{i,\star}^{(t)}.
\end{equation}
The key difference is that the uploaded model
$\boldsymbol{\theta}_{i,\star}^{(t)}$ is produced through controller-guided
deep-unfolded proximal QNN optimization and validation-based best-unfold
selection.

Therefore, DUQFL-Prox transfers the deep-unfolding principle from server-side
aggregation-weight learning to client-side quantum optimizer-policy learning.
This distinction is important because QFL introduces optimization challenges not
present in ordinary classical FL, including finite-shot measurement noise, SPSA
perturbation sensitivity, and non-convex variational quantum loss landscapes.
DUQFL-Prox is therefore complementary to classical deep-unfolded FL methods: a
future extension could jointly learn client-side QNN optimization policies and
server-side aggregation weights, whereas this work focuses on stabilizing local
QNN optimization before aggregation.

\section{Additional Ablation and Controller Trace Analysis}
\label{sec:Additional Ablation}

\begin{figure*}[htbp]
\centering

\begin{subfigure}[t]{0.245\textwidth}
    \centering
    \includegraphics[width=\linewidth]{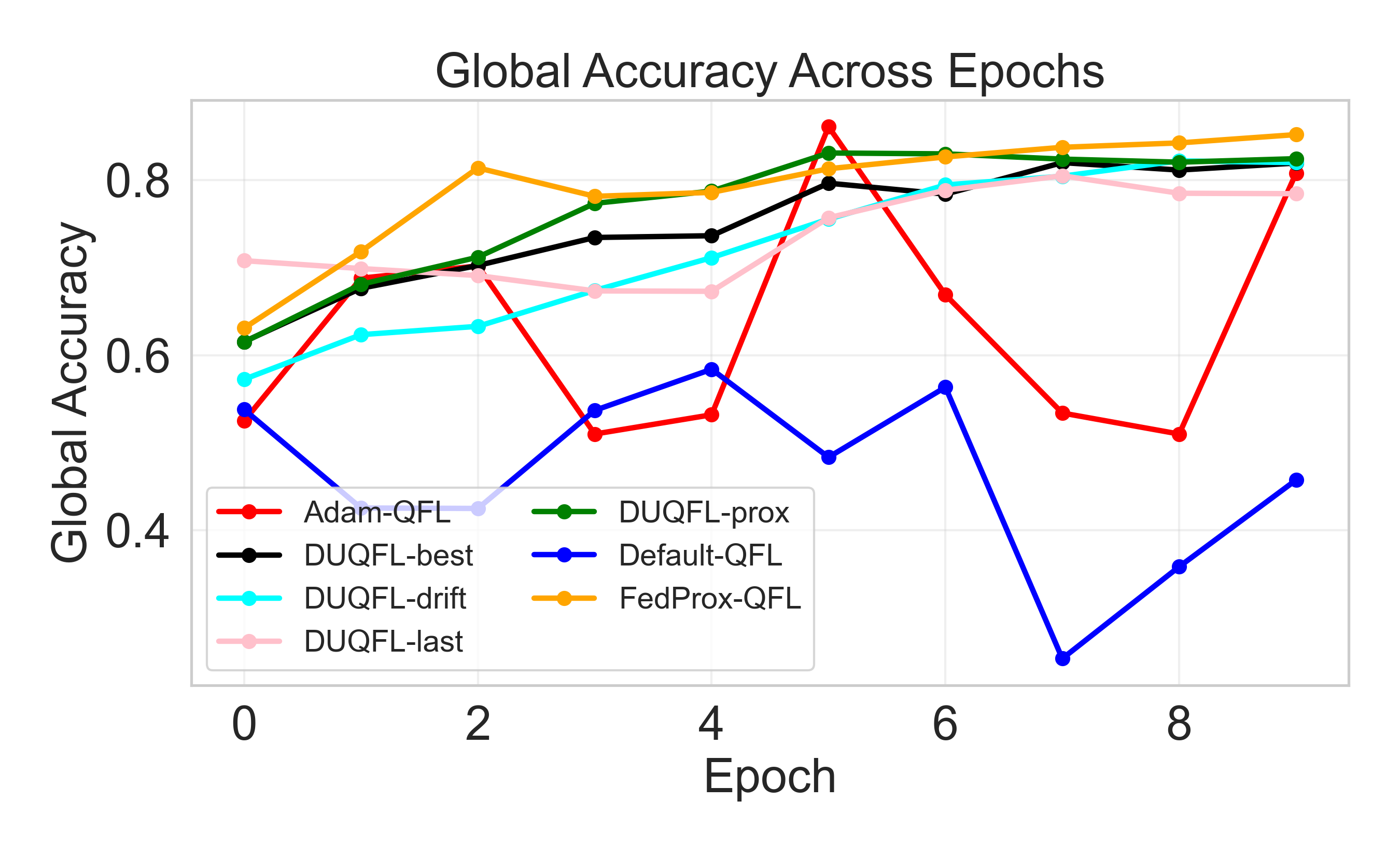}
    \caption{Global accuracy}
    \label{fig:ablation_global_accuracy}
\end{subfigure}\hfill
\begin{subfigure}[t]{0.245\textwidth}
    \centering
    \includegraphics[width=\linewidth]{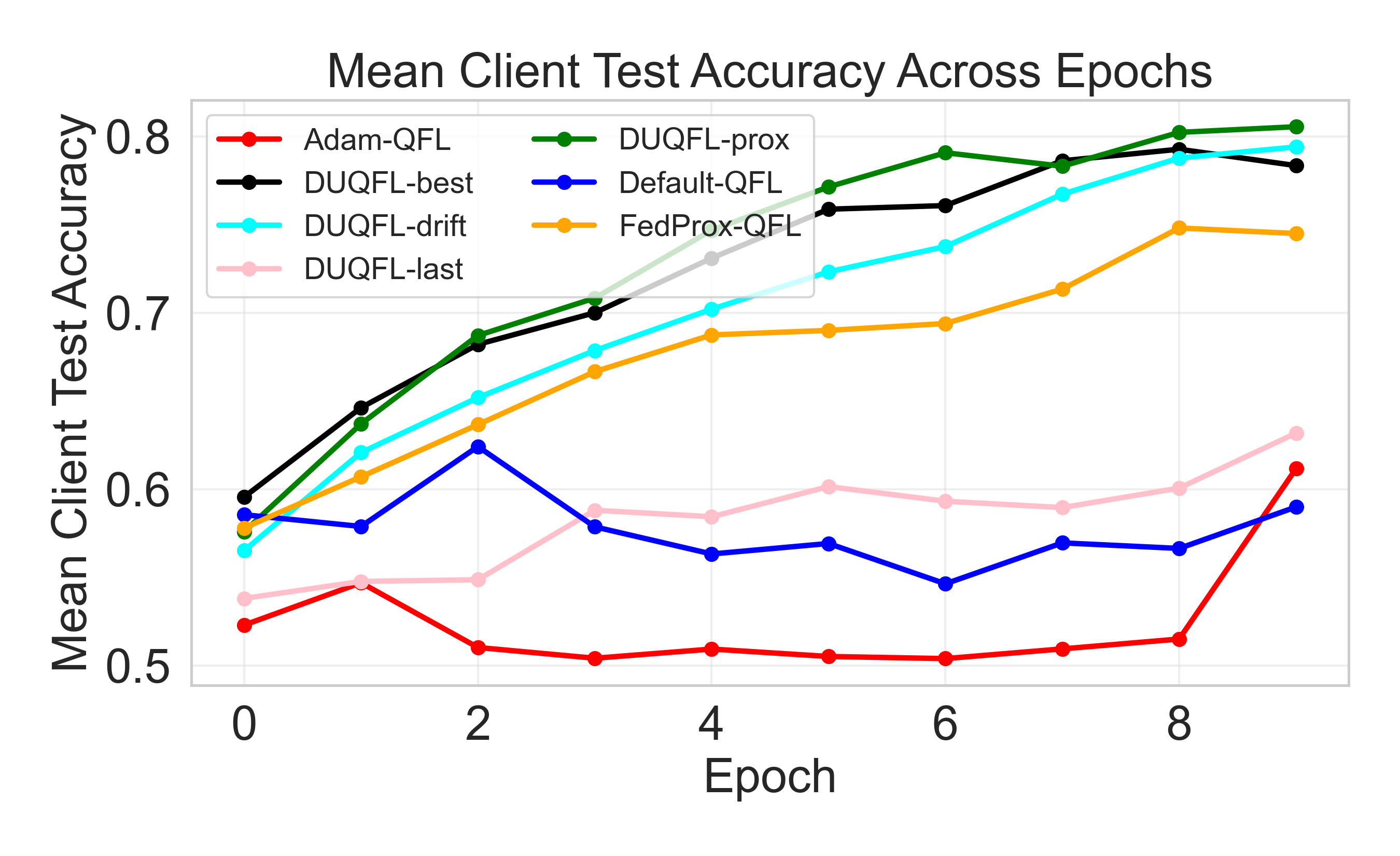}
    \caption{Mean client test accuracy}
    \label{fig:ablation_mean_client_test}
\end{subfigure}\hfill
\begin{subfigure}[t]{0.245\textwidth}
    \centering
    \includegraphics[width=\linewidth]{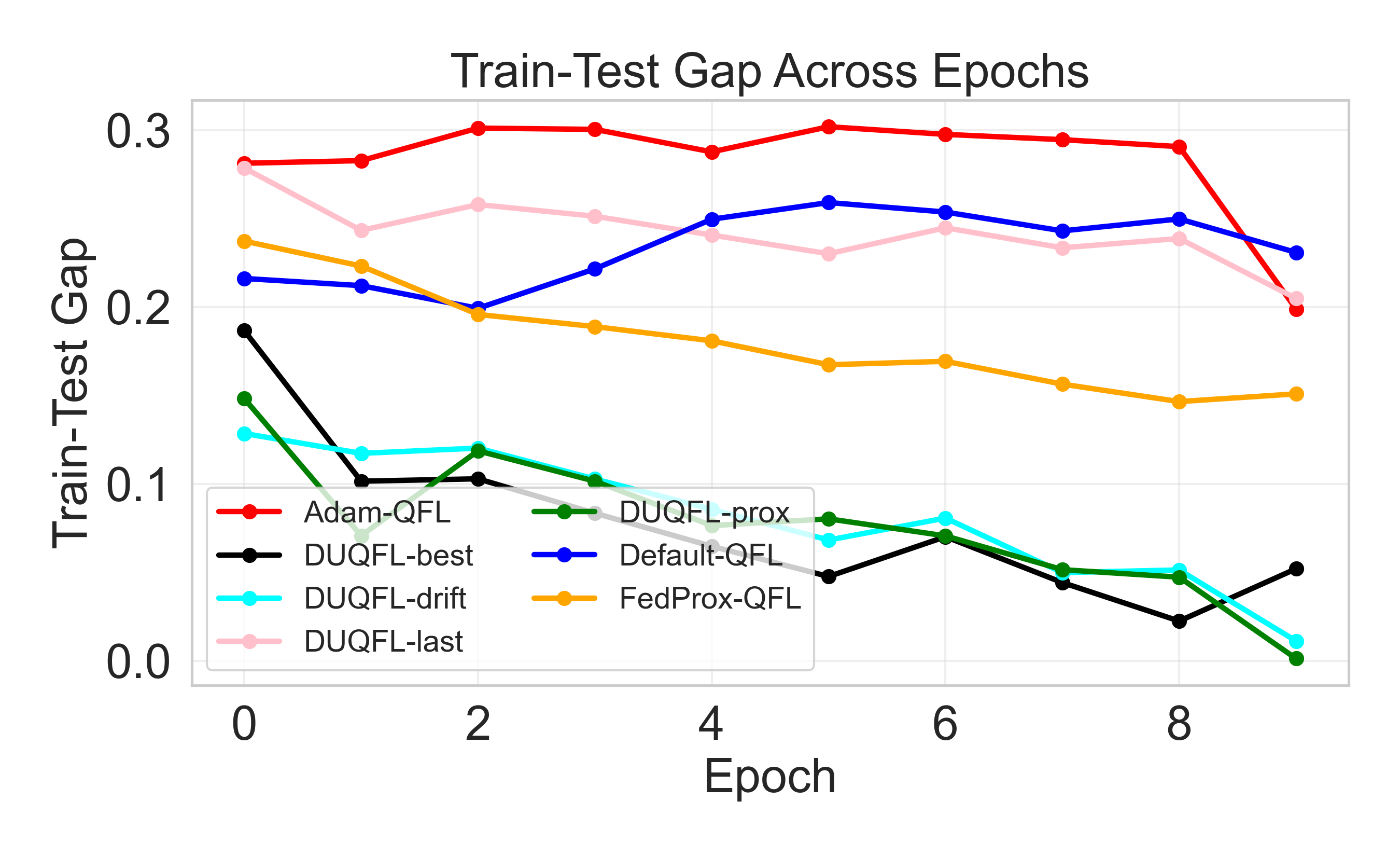}
    \caption{Train--test gap}
    \label{fig:ablation_train_test_gap}
\end{subfigure}\hfill
\begin{subfigure}[t]{0.245\textwidth}
    \centering
    \includegraphics[width=\linewidth]{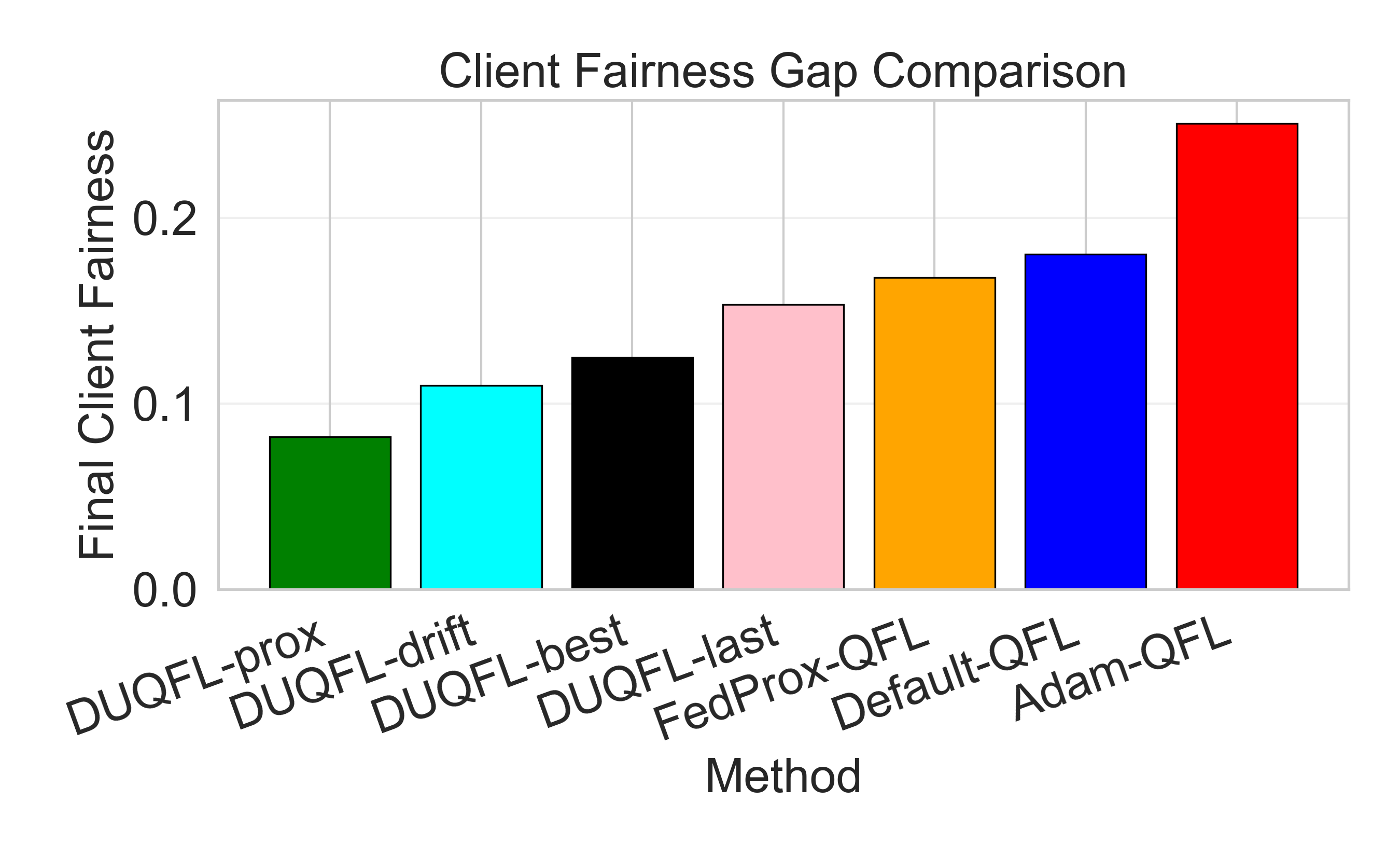}
    \caption{Client fairness gap}
    \label{fig:ablation_fairness_gap}
\end{subfigure}

\caption{Ablation analysis of DUQFL variants on the Genome non-IID setting.
DUQFL-last uploads the final unfolded local state, DUQFL-best uploads the
validation-preferred unfolded checkpoint, DUQFL-drift adds explicit displacement
control, and DUQFL-Prox incorporates proximal drift regularization. DUQFL-Prox
achieves the strongest client-level generalization, lowest train--test gap, and
smallest fairness gap, indicating that best-unfold selection and proximal drift
control improve stability and balance under heterogeneous QFL.}
\label{fig:ablation_duqfl_components}
\end{figure*}
\subsection{Compared Methods}
\label{subsec:compared_methods}

We compare DUQFL-Prox with representative QFL baselines and ablation variants.
All methods use the same QNN architecture, dataset split, client partitioning,
number of communication rounds, and evaluation protocol for each dataset. Thus,
differences in performance are attributed to the local optimization strategy,
checkpoint-selection mechanism, and drift-control design rather than to model
capacity or data partitioning. Table~\ref{tab:duqfl_ablation_roles} summarizes
the role of each method.

\begin{table*}[htbp]
\centering
\scriptsize
\caption{Interpretation of QFL baselines and DUQFL ablation variants.}
\label{tab:duqfl_ablation_roles}
\begin{tabular}{lcccc p{7cm}}
\hline
\textbf{Method}
& \textbf{Deep unfold}
& \textbf{Controller}
& \textbf{Best unfold}
& \textbf{Drift control}
& \textbf{Role} \\
\hline

Default-QFL
& $\times$ & $\times$ & $\times$ & $\times$
& Baseline QFL using Qiskit's default SPSA optimizer and FedAvg aggregation. \\

Adam-QFL
& $\times$ & $\times$ & $\times$ & $\times$
& Generic adaptive optimizer baseline used to assess whether standard local adaptivity is sufficient under QFL heterogeneity. \\

FedProx-QFL
& $\times$ & $\times$ & $\times$ & $\checkmark$
& Proximal QFL baseline that tests drift control without deep unfolding or controller-guided SPSA adaptation. \\

DUQFL-last
& $\checkmark$ & $\checkmark$ & $\times$ & $\times$
& Tests whether uploading the final unfolded local checkpoint is sufficient. \\

DUQFL-best
& $\checkmark$ & $\checkmark$ & $\checkmark$ & $\times$
& Tests validation-based best-unfold checkpoint selection. \\

DUQFL-drift
& $\checkmark$ & $\checkmark$ & $\checkmark$ & $\checkmark$
& Tests explicit displacement-based drift control within unfolded local QNN optimization. \\

DUQFL-Prox
& $\checkmark$ & $\checkmark$ & $\checkmark$ & $\checkmark$
& Full proposed method combining controller-guided unfolded SPSA, best-unfold selection, and proximal drift regularization. \\
\hline
\end{tabular}
\end{table*}

Although DUQFL-drift and DUQFL-Prox both include drift-control mechanisms, they
regularize drift differently. DUQFL-drift uses an explicit displacement
constraint, whereas DUQFL-Prox uses a proximal penalty that continuously
discourages deviation from the broadcast global model. Similarly, FedProx-QFL
and DUQFL-Prox both use proximal regularization, but at different levels:
FedProx-QFL regularizes a conventional local optimization routine, whereas
DUQFL-Prox embeds proximal regularization within a controller-guided unfolded
QNN optimization trajectory.

\subsection{Effect of Best-Unfold Selection and Proximal Drift Control}
\label{subsec:ablation_best_prox}

Figure~\ref{fig:ablation_duqfl_components} presents the ablation study on the
Genome non-IID setting. The comparison isolates the contribution of the main
DUQFL-Prox components: controller-guided unfolding, validation-based best-unfold
selection, and drift control. DUQFL-last uploads the final unfolded local state,
DUQFL-best uploads the validation-preferred unfolded checkpoint, DUQFL-drift adds
explicit displacement control, and DUQFL-Prox incorporates proximal drift
regularization.

The results show that DUQFL-last improves over Default-QFL in some rounds, but
its client-level performance remains less stable than DUQFL-best and
DUQFL-Prox. This indicates that the final unfolded state is not necessarily the
most generalizable checkpoint under heterogeneous client data. Validation-based
best-unfold selection improves this behaviour by preventing over-aggressive or
poorly generalizing later unfold states from being uploaded to the server.

Adding drift control further improves stability. DUQFL-Prox achieves the highest
final mean client test accuracy, the lowest train--test gap, and the smallest
client fairness gap among the DUQFL variants. These trends suggest that proximal
regularization improves aggregation compatibility by limiting excessive client
movement, while best-unfold selection improves the quality of the checkpoint
selected for aggregation.

The comparison with FedProx-QFL is also important. Although both FedProx-QFL and
DUQFL-Prox include proximal regularization, FedProx-QFL applies it to a standard
local optimization process. In contrast, DUQFL-Prox applies proximal
regularization inside an adaptive unfolded QNN optimization trajectory whose
SPSA learning rate and perturbation scale are generated by a controller.
Therefore, the improvement of DUQFL-Prox over FedProx-QFL indicates that
proximal regularization alone is insufficient; stable heterogeneous QFL also
requires adaptive control of the local QNN optimization process before
aggregation.
\begin{figure}[t]
\centering

\begin{subfigure}[t]{0.48\linewidth}
    \centering
    \includegraphics[width=\linewidth]{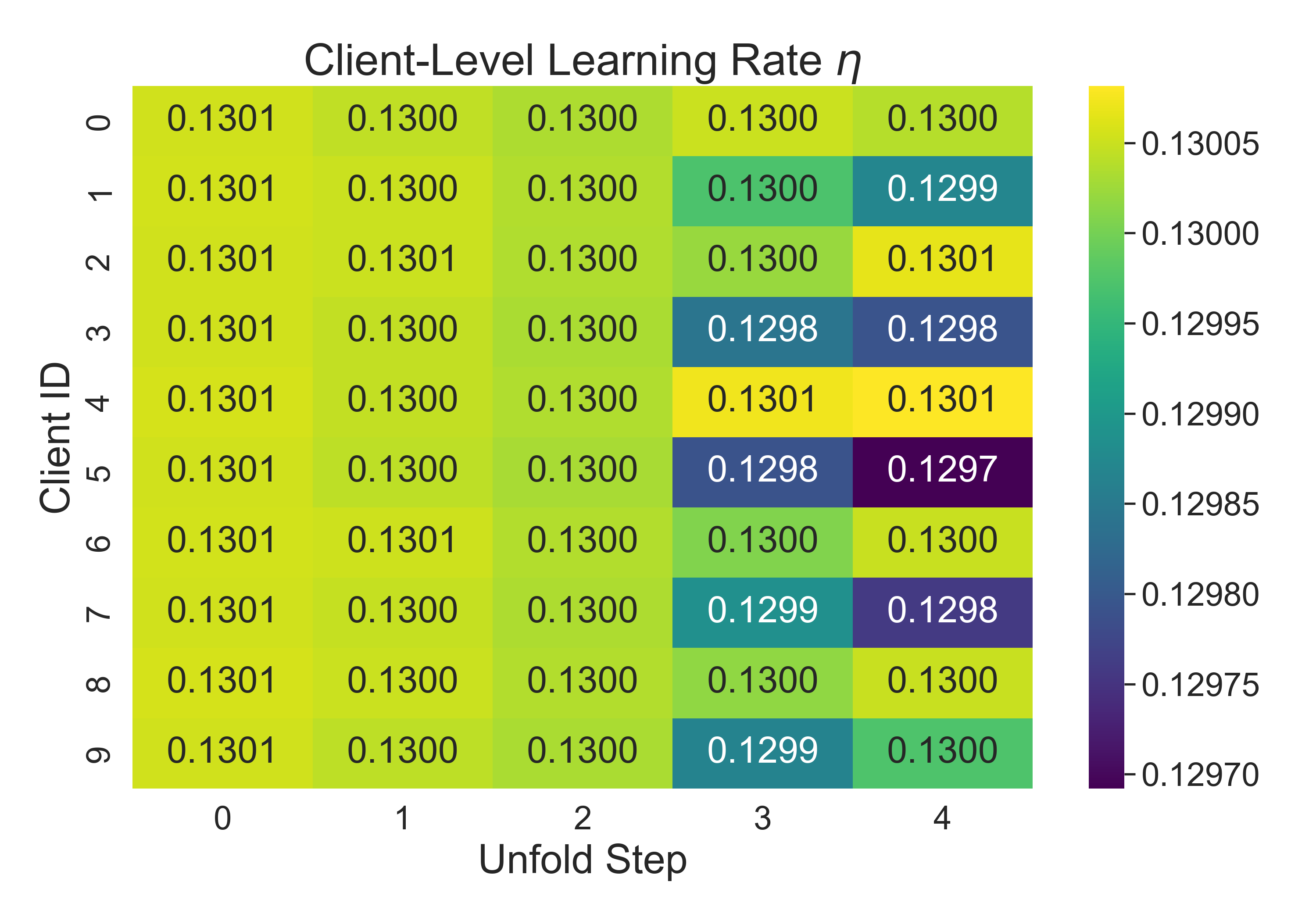}
    \caption{Generated learning rate $\eta$}
    \label{fig:heatmap_lr_client_unfold}
\end{subfigure}\hfill
\begin{subfigure}[t]{0.48\linewidth}
    \centering
    \includegraphics[width=\linewidth]{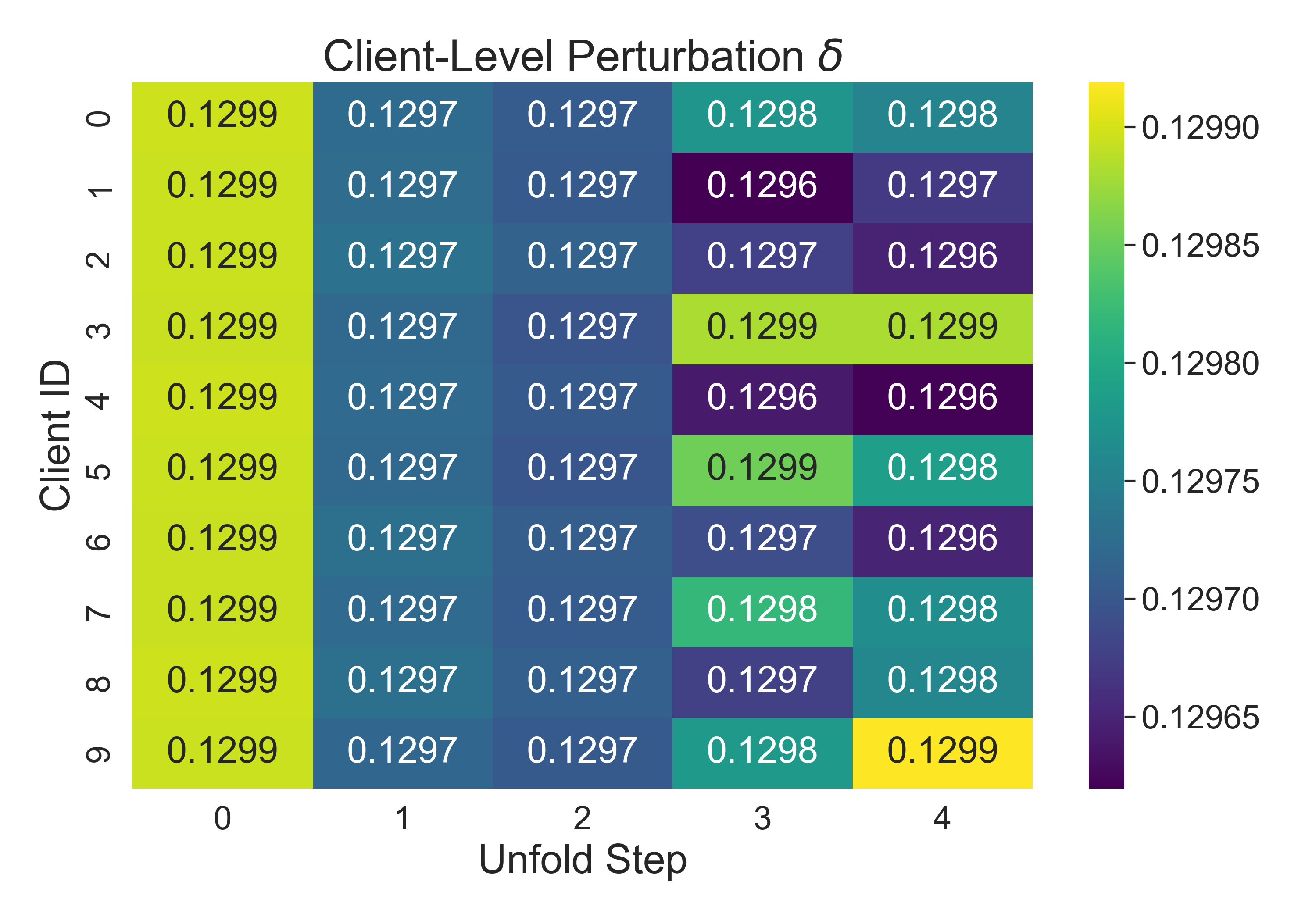}
    \caption{Generated perturbation $\delta$}
    \label{fig:heatmap_pert_client_unfold}
\end{subfigure}
\caption{Client- and unfold-level controller behaviour in DUQFL-Prox. The
heatmaps show that the generated SPSA learning rate and perturbation scale vary
across clients and unfold steps, supporting the claim that DUQFL-Prox performs
controller-guided local optimization rather than using a fixed SPSA schedule.}
\label{fig:controller_heatmaps}
\end{figure}

\begin{figure}[t]
\centering

\begin{subfigure}[t]{0.48\linewidth}
    \centering
    \includegraphics[width=\linewidth]{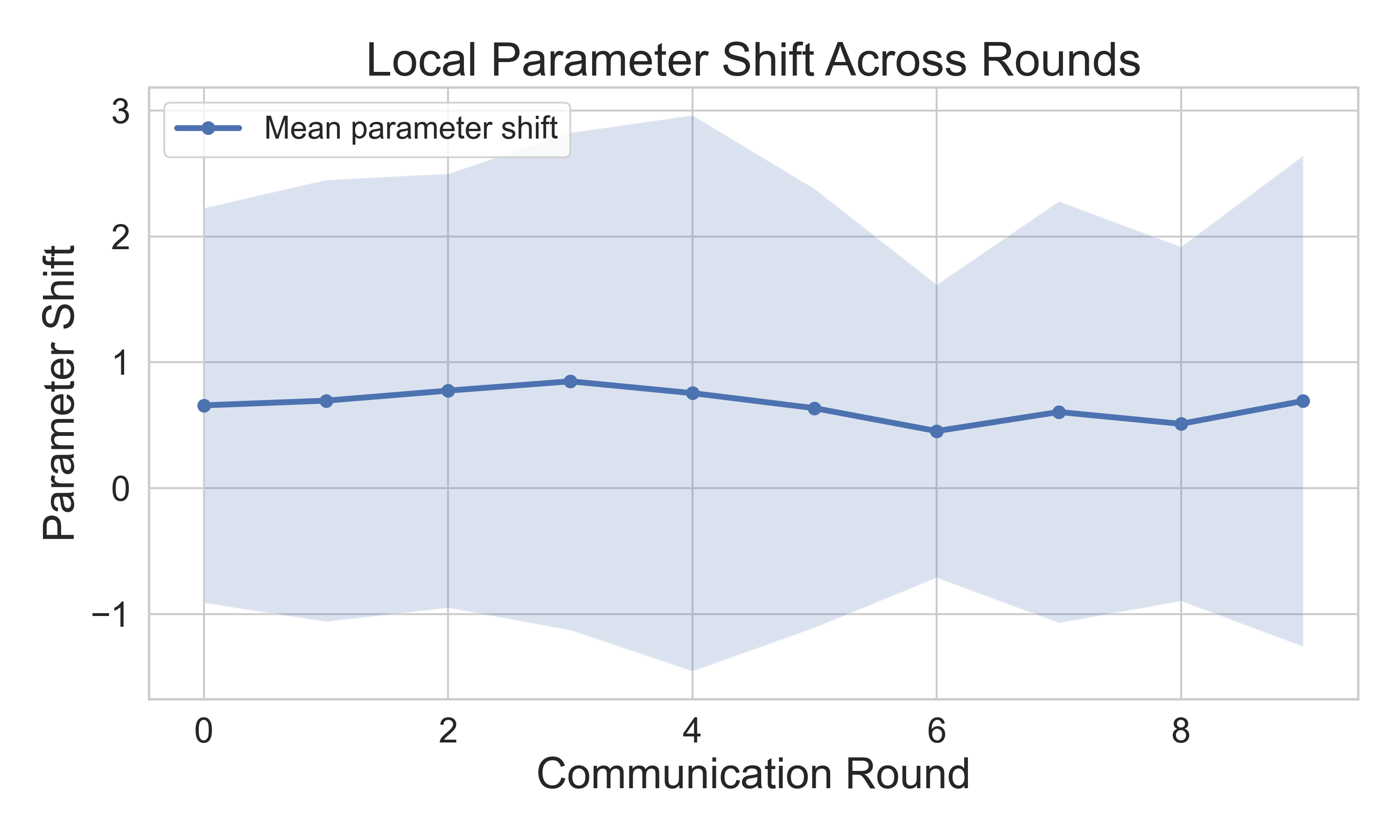}
    \caption{param shift over rounds}
    \label{fig:param_shift_over_rounds}
\end{subfigure}\hfill
\begin{subfigure}[t]{0.48\linewidth}
    \centering
    \includegraphics[width=\linewidth]{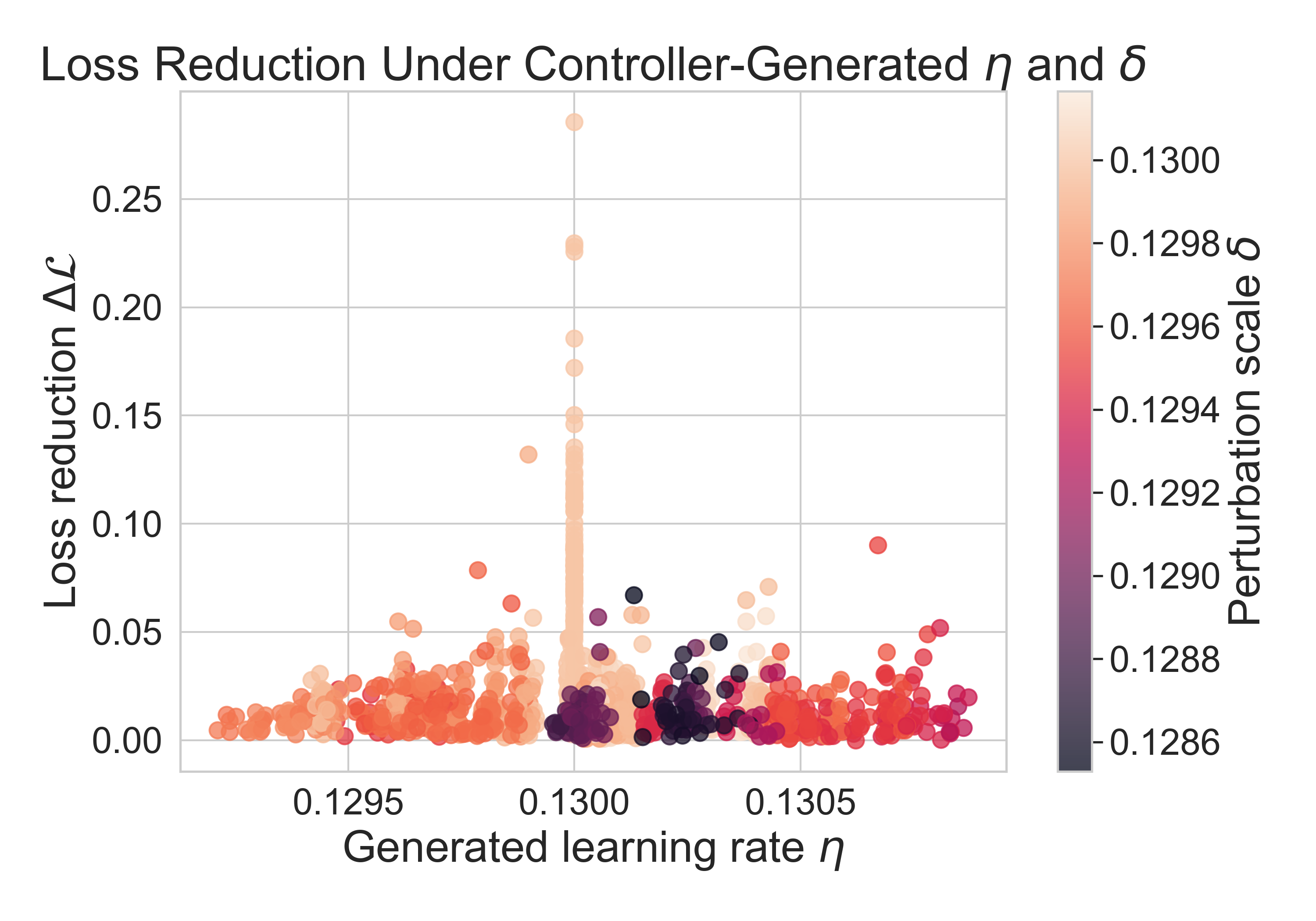}
    \caption{scatter lr pert loss delta}
    \label{fig:heatmap_pert_client_unfold}
\end{subfigure}

\caption{Trace-level diagnostics of local optimization in DUQFL-Prox. The
parameter-shift curve monitors local movement during unfolded optimization,
while the scatter plot shows that controller-generated SPSA hyperparameters
remain within a stable numerical range and produce positive local loss reduction
for most client-unfold updates.}
\label{fig:loss_shift_trace}
\end{figure}

\begin{figure}[t]
\centering

\begin{subfigure}[t]{0.48\linewidth}
    \centering
    \includegraphics[width=\linewidth]{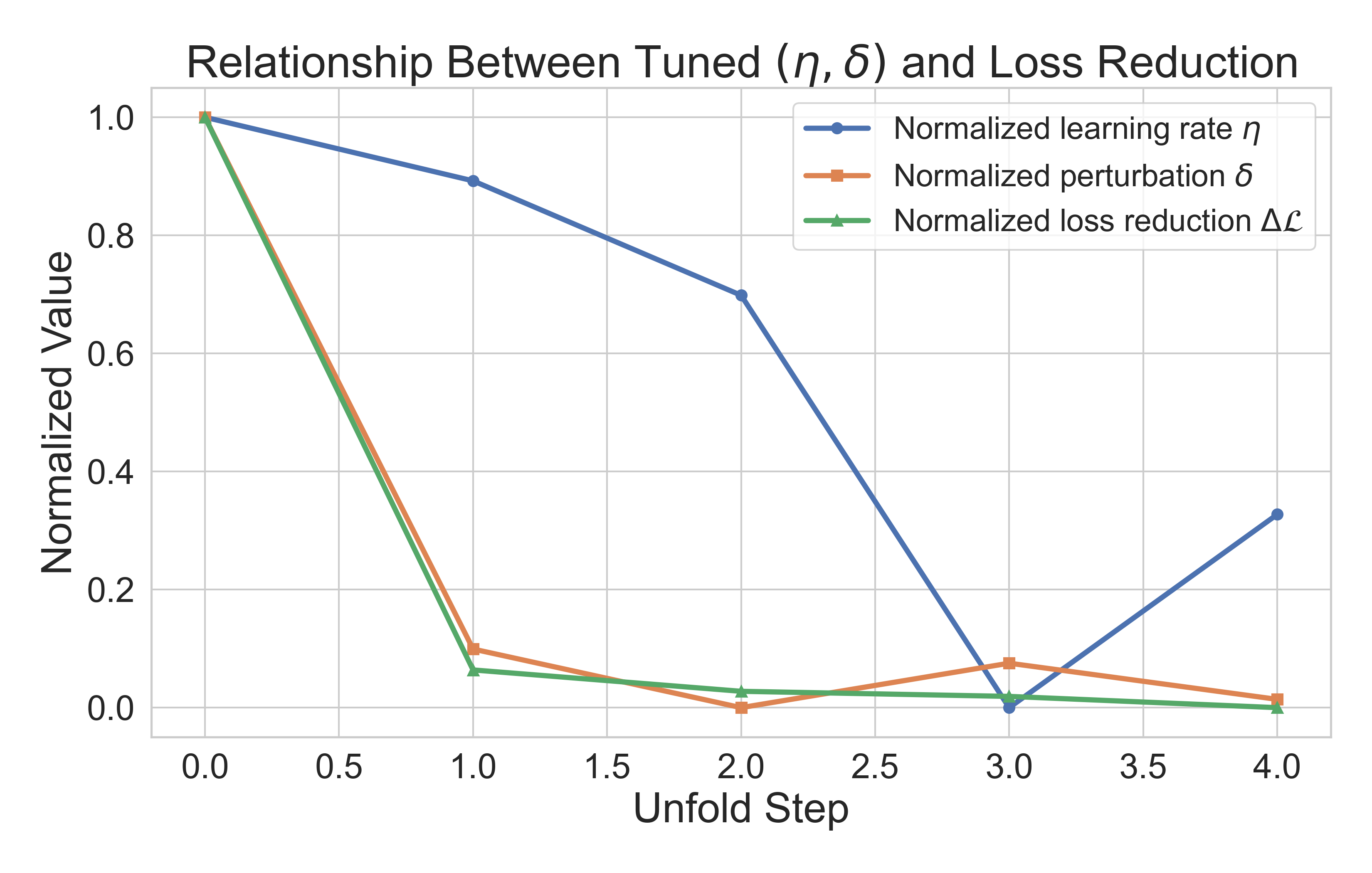}
    \caption{Normalized $\eta$, $\delta$, and loss reduction}
    \label{fig:eta_delta_loss_reduction_by_unfold}
\end{subfigure}\hfill
\begin{subfigure}[t]{0.48\linewidth}
    \centering
    \includegraphics[width=\linewidth]{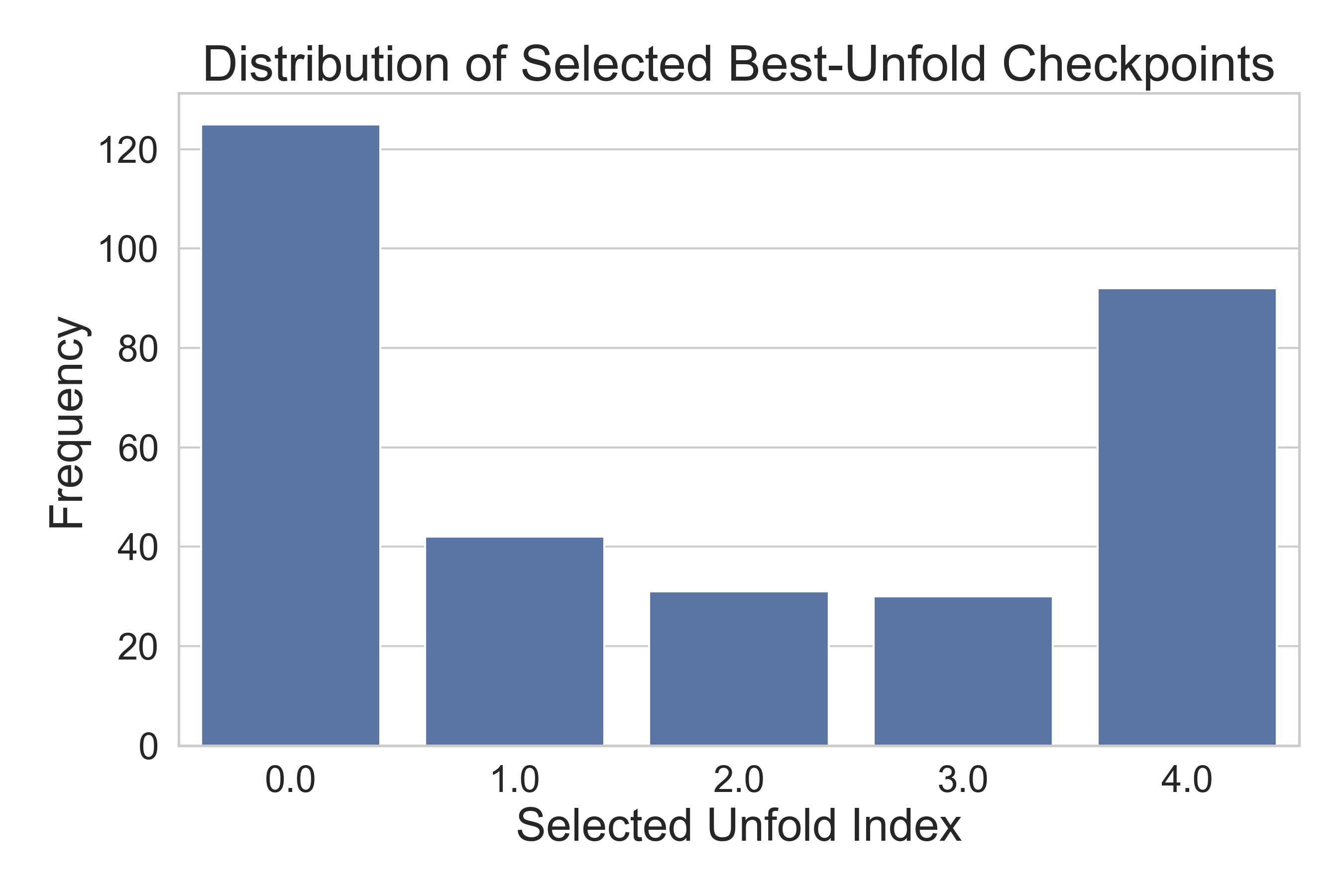}
    \caption{selected unfold distribution}
    \label{fig:selected_unfold_distribution}
\end{subfigure}
\caption{Unfold-level behaviour of DUQFL-Prox. The normalized plot shows that
the largest local loss reduction often occurs during early unfold steps, while
the selected-unfold distribution confirms that the final unfolded state is not
always the checkpoint uploaded for aggregation. These diagnostics support the
use of validation-based best-unfold selection.}
\label{fig:unfold_level_behaviour}
\end{figure}

\section{Additional Controller Trace Analysis}
\label{app:controller_trace}

This appendix provides trace-level diagnostics for DUQFL-Prox. While the main
ablation study evaluates global and client-level performance, the following
figures examine the internal behaviour of the controller during unfolded local
QNN optimization. These diagnostics support the claim that DUQFL-Prox does not
use a fixed SPSA schedule, but instead generates client- and unfold-dependent
optimization behaviour.

Figure~\ref{fig:controller_heatmaps} reports the learning rate $\eta$ and
perturbation scale $\delta$ generated by the controller for each client and
unfold step. The heatmaps show that the generated SPSA hyperparameters vary
across clients and unfold steps, indicating that the controller adapts local
optimizer behaviour according to the optimization state rather than applying a
single fixed schedule.

Figure~\ref{fig:loss_shift_trace} provides two complementary diagnostics. The
parameter-shift curve monitors local movement across communication rounds,
whereas the loss-reduction scatter plot relates controller-generated
hyperparameters to local loss improvement. Most client-unfold updates yield
positive loss reduction while the generated learning rates and perturbation
scales remain within a stable numerical range. This supports the interpretation
that the controller performs fine-grained adaptation without inducing unstable
hyperparameter excursions.

Figure~\ref{fig:unfold_level_behaviour} further supports the
validation-based checkpointing mechanism. The normalized unfold-step analysis
shows that the largest average loss reduction often occurs during early unfold
steps, while the selected-unfold distribution confirms that the final unfolded
state is not always selected for upload. Therefore, best-unfold selection is
necessary to prevent later, less generalizable unfold states from dominating the
server aggregation.

\subsection{Trace Logging and Reproducibility}
\label{subsec:trace_logging_reproducibility}

To support reproducibility and detailed ablation analysis, each experimental run
saved round-level, client-level, and classification-level outputs. Round-level
CSV files stored global accuracy, validation loss, client accuracy statistics,
meta-loss, fairness information, and timing. Client-level trace files stored
the unfold index, generated learning rate, perturbation scale, local losses,
local train/test accuracy, validation loss, parameter displacement, clipping
indicator, selected unfold index, client size, and heterogeneity value. These
trace logs make it possible to verify whether the controller changes the local
SPSA behaviour across clients, unfold steps, and communication rounds.

The main saved outputs were:
\begin{itemize}
    \item \texttt{global\_accuracies.csv}: global and client train/test
    accuracies across communication rounds;
    \item \texttt{validation.csv}: validation loss across rounds;
    \item \texttt{client\_trace.csv}: unfold-level client traces, including
    generated $\eta$, generated $\delta$, validation loss, parameter shift, and
    selected unfold index;
    \item \texttt{outer\_meta.csv}: outer SPSA controller-update information,
    including perturbed meta-loss values and controller-gradient norm;
    \item \texttt{classification\_metrics.csv}: precision, recall, F1-score,
    ROC-AUC, PR-AUC, MCC, specificity, and confusion-matrix values;
    \item \texttt{global\_params.npz}: saved global QNN parameter vectors across
    communication rounds.
\end{itemize}
The result files were organized by method, dataset, split type, random seed, and
timestamp. This structure allows each run to be regenerated from the stored
configuration and output files.

\section{IBM Quantum Hardware Workload Evidence}
\begin{figure}
    \centering
    \includegraphics[width=0.95\linewidth]{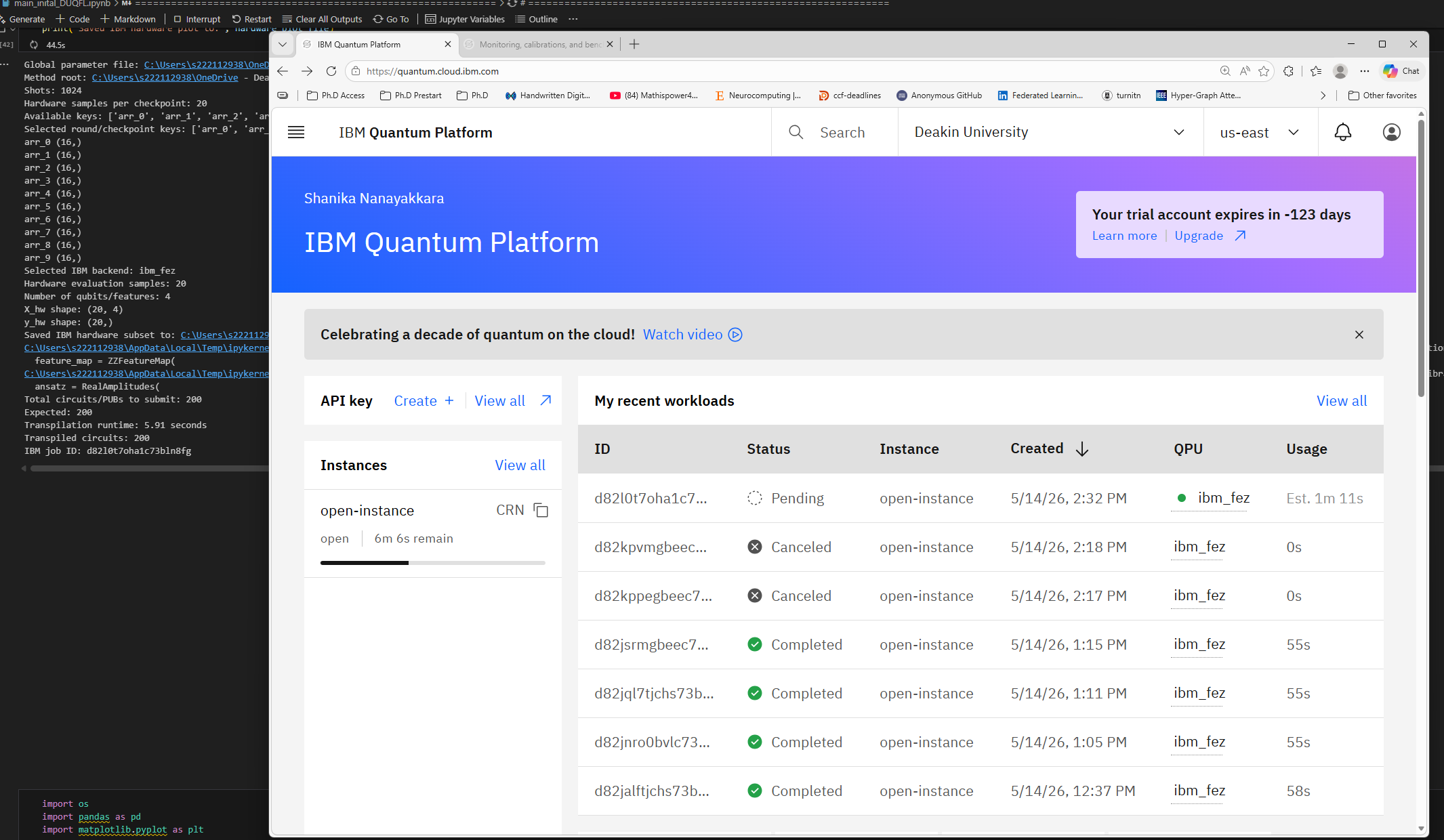}
    \caption{IBM Quantum Platform workload evidence for real-hardware execution of
trained QFL checkpoints. The workload history shows completed executions on the
\texttt{ibm\_fez} backend using finite-shot measurement. This screenshot is
included as execution evidence only; the quantitative simulator--hardware
accuracy comparison is reported separately.}
\label{fig:ibm_workload_evidence}
\end{figure}

To assess practical deployability, selected trained global QNN checkpoints were
executed on real IBM quantum hardware. This experiment was performed as
post-training hardware validation rather than full hardware-in-the-loop
federated training. The saved global checkpoints were assigned to the trained
QNN circuits, transpiled for the selected IBM backend, and executed using
finite-shot measurement.

Figure~\ref{fig:ibm_workload_evidence} provides IBM Quantum Platform workload
evidence for the hardware execution. The workload history shows completed
executions on the \texttt{ibm\_fez} backend using the IBM Quantum open instance.
This screenshot is included only as execution evidence; the quantitative
simulator--hardware accuracy comparison is reported separately in the main
experimental results.

\vfill

\end{document}